\documentclass{article} 
\usepackage{iclr2027_conference,times}

\usepackage{amsmath,amsfonts,bm}

\def\eqref#1{equation~\ref{#1}}
\def\Eqref#1{Equation~\ref{#1}}

\def\1{\bm{1}}

\def\rp{{\textnormal{p}}}

\DeclareMathAlphabet{\mathsfit}{\encodingdefault}{\sfdefault}{m}{sl}
\SetMathAlphabet{\mathsfit}{bold}{\encodingdefault}{\sfdefault}{bx}{n}

\newcommand{\E}{\mathbb{E}}

\newcommand{\R}{\mathbb{R}}

\newcommand{\KL}{D_{\mathrm{KL}}}

\usepackage{url}

\usepackage{amsmath}
\usepackage{amssymb}
\usepackage{mathtools}
\usepackage{amsthm}
\RequirePackage{algorithm}
\RequirePackage{algorithmic}
\usepackage{booktabs}

\usepackage[most]{tcolorbox}
\usepackage{fontawesome5}

\definecolor{cLoop}{HTML}{2A9D8F}
\definecolor{cStep}{HTML}{E07A28}
\definecolor{cInk}{HTML}{263248}
\definecolor{cBase}{HTML}{7A8FB0}
\definecolor{cEnt}{HTML}{7B5EA7}

\colorlet{TakeawayBorder}{cBase}
\colorlet{TakeawayBlue}{cBase!15!white}

\colorlet{ContributionBorder}{cLoop}
\colorlet{ContributionGreen}{cLoop!15!white}

\definecolor{fontcolor}{HTML}{212121}

\newtcolorbox{takeawaybox}[1][]{%
  enhanced,
  breakable,
  colback=TakeawayBlue,
  colframe=TakeawayBorder,
  boxrule=0pt,
  frame hidden,
  left=8pt, right=8pt, top=6pt, bottom=6pt,
  boxsep=2pt,
  arc=0pt,
  before skip=0.75\baselineskip,
  after skip=0.9\baselineskip,
  overlay={%
    \draw[TakeawayBorder, line width=2pt]
      ([yshift=1pt]frame.north west) -- (frame.south west);
  },
  fontupper=\small\itshape\color{fontcolor},
}

\newtcolorbox{contributionbox}[1][]{%
  enhanced,
  breakable,
  colback=ContributionGreen,
  colframe=ContributionBorder,
  boxrule=0pt,
  frame hidden,
  left=8pt, right=8pt, top=6pt, bottom=6pt,
  boxsep=2pt,
  arc=0pt,
  before skip=0.75\baselineskip,
  after skip=0.9\baselineskip,
  overlay={%
    \draw[ContributionBorder, line width=2pt]
      ([yshift=1pt]frame.north west) -- (frame.south west);
  },
  fontupper=\small\itshape\color{fontcolor},
}

\theoremstyle{plain}
\newtheorem{theorem}{Theorem}[section]
\newtheorem{proposition}[theorem]{Proposition}
\newtheorem{lemma}[theorem]{Lemma}
\newtheorem{corollary}[theorem]{Corollary}
\theoremstyle{definition}

\theoremstyle{remark}
\newtheorem{remark}[theorem]{Remark}

\usepackage{xargs}
\usepackage{xcolor}
\usepackage{tikz}
\usepackage{caption}

\usepackage{siunitx} 
\usepackage{color, colortbl}
\definecolor{Gray}{gray}{0.9}
\definecolor{LightGray}{gray}{0.97}

\newcolumntype{g}{>{\columncolor{LightGray}}S}

\usepackage{graphicx}

\usepackage{enumitem}      
\usepackage{xcolor}        
\usepackage{pifont}        

\definecolor{mydarkblue}{rgb}{0,0.08,0.45}
\definecolor{mylightblue}{rgb}{0.41,0.69,0.83}
\usepackage[%
  colorlinks = true,
  citecolor  = mydarkblue,
  linkcolor  = mydarkblue,
  filecolor  = mydarkblue,
  urlcolor   = mydarkblue,
  unicode,     
  pagebackref  
  ]{hyperref}

\usepackage{pifont}

\usepackage{xargs}

\usepackage{float}
\usepackage{bbding}
\usepackage{colortbl}
\definecolor{LightGray}{gray}{0.93}
\usepackage{graphicx}
\usepackage{subcaption}
\usepackage{amsmath, amsfonts, amsthm}
\usepackage{multirow}

\usepackage{graphicx}
\usepackage{wrapfig}
\usepackage{pdfpages}

\usepackage{standalone}
\usepackage{tikz}
\usetikzlibrary{arrows.meta, calc}

\usepackage{xspace}

\newcommand{\MDM}{\textsc{MDM}}

\newcommand{\AR}{\textsc{AR}}

\newcommand{\ie}{\textit{i.e.},\xspace}

\newcommand{\mask}{\texttt{[MASK]}}
\newcommand{\xt}{x_t}
\newcommand{\xo}{x_0}

\usepackage{bm}

\usepackage{minitoc}

\mtcsetfont{parttoc}{section}{\footnotesize\rmfamily\upshape\mdseries}
\mtcsetfont{parttoc}{subsection}{\footnotesize\rmfamily\upshape\mdseries}

\newcommand{\NFE}{\textsc{nfe}\xspace}
\newcommand{\Ddep}{D_{\mathrm{dep}}}
\newcommand{\Dfid}{D_{\mathrm{fid}}}
\newcommand{\TC}{\mathrm{TC}}
\newcommand{\TV}{\mathrm{TV}}
\newcommand{\CE}{\mathrm{CE}}
\newcommand{\pidata}{\pi_{\mathrm{data}}}

\newcommand{\Vocab}{\mathcal{V}}

\newcommand{\Dfloor}{\mathcal{D}_{\mathrm{dep}}}

\newcommand{\meth}[1]{\texttt{#1}}
\newcommand{\entent}{\meth{entropy\_entropy}\xspace}
\newcommand{\entconf}{\meth{entropy\_confidence}\xspace}
\newcommand{\hident}{\meth{hidden\_entropy}\xspace}
\newcommand{\fixunif}{\meth{fixed\_uniform}\xspace}
\newcommand{\schedunif}{\meth{sched\_uniform}\xspace}
\newcommand{\fixent}{\meth{fixed\_entropy}\xspace}
\newcommand{\entunif}{\meth{entropy\_uniform}\xspace}

\usepackage[capitalize,noabbrev]{cleveref}

\title{\mbox{Recursive Scaling in Masked Diffusion Models}}

\author{Alba Carballo-Castro\thanks{Correspondence to: \texttt{alba.carballocastro@epfl.ch}, \texttt{jp2048@cam.ac.uk}} \\ LTS4, EPFL \\ Lausanne, Switzerland
\And Julianna Piskorz \\ University of Cambridge \\ Cambridge, UK
\And Paulius Rauba \\ University of Cambridge \\ Cambridge, UK
\AND Mihaela van der Schaar \\ University of Cambridge \\ Cambridge, UK
\And Pascal Frossard \\ LTS4, EPFL \\ Lausanne, Switzerland}

\iclrfinalcopy 

\begin{document}

\maketitle

\begin{abstract}
Masked diffusion models (MDMs) generate sequences by iteratively refining a partially masked state and committing tokens in parallel. We introduce recursion in MDMs and propose new Recursive Masked Diffusion Models (R-MDMs), which apply a shared denoising transformer $L$ times within each denoising step, adding recursive depth as an additional compute axis without increasing parameter count. Across structured generation tasks, recursive depth improves quality at fixed parameter budget, matches substantially larger non-recursive models at matched FLOPs, and can reduce the number of denoising steps needed to reach a target quality. We interpret these gains with a dependence--fidelity decomposition of parallel decoding error: recursion refines model marginals at a fixed masked state, whereas denoising steps change that state by committing tokens. Building on this analysis, we propose to treat decoding as a two-axis decision (how many loops to run and which tokens to commit) and show that entropy-guided adaptive rules improve the quality--compute frontier over fixed schedules, transferring across various tasks on Sudoku, Countdown, RNA, and executable math generation. Together, these results establish recursive depth as a practical, complementary test-time scaling mechanism for MDMs.
\end{abstract}

\section{Introduction}
\label{sec:introduction}

Diffusion models have emerged as a compelling paradigm for generative modeling of continuous data~\citep{sohl2015diffusion, ho2020ddpm,song2020score}, while discrete diffusion models have more recently been developed for sequence generation, including masked diffusion~\citep{austin2021d3pm,sahoo2024mdlm,nie2025llada}. 
Among them, masked diffusion models (\MDM{}s) generate sequences by repeatedly denoising a partially masked state, attending bidirectionally to the whole sequence, predicting every masked position at once, and committing some subset of them. Generation is therefore iterative, and the number of denoising steps $T$ is a native test-time compute scaling axis: more steps mean fewer tokens are committed at a time, so fewer dependencies between them go unresolved and generation quality usually improves, but at proportionally higher cost.

Another scaling axis is model size, which recent \MDM{} research has mostly pursued by following \AR{} practice: adding more width, attention heads, and layers, so that compute grows with parameter count~\citep{kaplan2020scaling}. 
Following this recipe, \MDM{}s have grown from small models~\citep{austin2021d3pm,sahoo2024mdlm} to billions of parameters~\citep{nie2025llada,ye2025dream7bdiffusionlarge,gemini2026diffugemma}, reaching generation quality competitive with \AR{} models.

Yet, model size and denoising steps are not the only axes of capability. A third, \emph{recursive depth}, has recently been developed in \AR{} models: applying the same transformation repeatedly to add computation without adding parameters~\citep{dehghani2019universal,giannou2023looped}. In particular, a $K$-layer transformer looped $L$ times can match a $KL$-layer model while using $L\times$ fewer parameters~\citep{saunshi2025looped}. We now bring recursion to \MDM{}s, where it plays a different role. In \AR{} models, looping deepens the computation behind a single next-token prediction. In contrast, bidirectional attention in \MDM{}s lets each loop jointly refine all masked positions at a fixed partially masked state, while denoising steps commit tokens and advance generation.

This motivates our central question: how does recursive computation interact with the denoising process in \MDM{}s? Although both provide additional computation, they operate at different levels: denoising steps change the partially masked sequence by committing tokens, whereas recursive loops refine predictions while holding the denoising state fixed. By making each parallel denoising update more accurate, recursion may reduce the number of denoising steps needed to reach a given quality. Thus, we ask whether additional recursive steps can substitute for denoising steps, whether the two axes address different failure modes, and how their computation should be allocated at inference time.

We study this question through \emph{recursive masked diffusion models} (R-\MDM{}s), which apply a denoising block $L$ times within each denoising step, reusing the same weights at every application and emitting a prediction after each. Recursion is therefore a compute axis rather than a size axis: the parameter count does not depend on $L$. We show that increasing the recursive depth allows R-\MDM{}s to match the performance of larger non-recursive models at equal FLOPs, and that it reduces predictive entropy, within-step conflicts, and early decoding failures, revealing how recursion improves the quality of parallel updates and explaining its stronger benefits at small denoising budgets. Building on this, we treat decoding as a two-axis policy and show that uncertainty-guided allocation between loops and denoising steps improves the quality--compute frontier across Sudoku, Countdown, RNA sequence generation, and executable math generation on TinyGSM.

\textbf{Contributions.}
\textbf{(i) Model }(Section 3). We propose R-MDMs, a new, simple method for introducing recursive depth into \MDM{}s, and ablate their main design choices; we empirically show that recursive depth improves generation quality at fixed parameter count and matches substantially larger non-recursive models at matched FLOPs. \textbf{(ii) Analysis} (Section 4). Through a careful decomposition of the parallel-decoding error we separate the roles of recursive steps and denoising steps (token commitment), identifying their complementary roles, and empirically establish why and when recursive refinement helps \MDM{}s. \textbf{(iii) Decoding }(Section 5). We turn this analysis into entropy-guided rules for \textit{when to stop refining} and \textit{which tokens to commit}, and empirically validate their quality--compute benefits across generative tasks on Sudoku, Countdown, RNA infilling, and executable math generation.

\section{Related Work}
\label{sec:related}

We review the lines of work most directly relevant to our study and defer a broader discussion of the related literature to Appendix~\ref{app:related}.

\vspace{-0.2cm}

\paragraph{Masked diffusion language models.} Masked (absorbing-state) discrete diffusion has become a competitive alternative to \AR{} language modeling~\citep{austin2021d3pm, lou2024sedd, sahoo2024mdlm, nie2025llada, gemini2026diffugemma}. Because tokens are denoised in parallel from bidirectional context, the number of denoising steps $T$ trades speed against the dependencies the sampler can resolve before committing. 
As a result, a large body of work asks \emph{which} masked tokens to reveal and \emph{how many}: MaskGIT~\citep{chang2022maskgit} established confidence-based unmasking, \citet{kim2025tokenordering} show that adaptive ordering matters on structured problems, EB-Sampler~\citep{benhamu2025ebsampler} uses an entropy bound to determine how many tokens may be safely unmasked per evaluation, and \citet{cai2026confidence} give iteration-complexity guarantees for entropy-sum unmasking. We take our commit rules from this line rather than proposing new ones.

\vspace{-0.2cm}
\paragraph{Recursive and looped computation.}
Depth-wise weight sharing runs from Universal Transformers and ALBERT~\citep{dehghani2019universal, lan2020albert} to looped transformers, where a $K$-layer block applied $L$ times can approach the effective depth of a $KL$-layer model with far fewer parameters ~\citep{giannou2023looped, yang2024looped,
geiping2025recurrent, yu2025relay, xu2025formal}; \citet{saunshi2025looped} further interpret loops as latent reasoning steps. Among these variants we adopt the plainest construction (a single shared block applied repeatedly, with no architectural search) for simplicity of our exploration.

\vspace{-0.2cm}
\paragraph{Positioning of this work.}
We study recursive computation inside each MDM denoising step as a complementary axis of inference-time scaling. Concurrent work introduces repeated computation into masked generation through different training mechanisms: LoopMDM~\citep{lee2026loopmdm} loops a subset of transformer layers, and Fixed-Point Masked Generative Modeling~\citep{miele2026fpmgm} embeds a fixed-point solver in the denoiser. 
Rather than comparing recursive architectures for \MDM{}s, we seek to determine whether recursion provides an effective scaling axis, when it is most beneficial, and how it should be used at inference time. We therefore keep our recursive architecture deliberately simple, enabling us to isolate the effects of recursion and study its interaction with \MDM{} decoding.

\section{Recursive Masked Diffusion Models}
\label{sec:rmdm}

In this section, we introduce R-\MDM{}s, describe their architecture, optimization procedure, and key design choices, and empirically show that recursive depth provides an effective scaling axis.

\subsection{Definition}
\label{sec:rmdm_def}

\textbf{Preliminaries.}
For $\xo \in \Vocab^n$, the absorbing-state forward process masks each token independently, $q(\xt\mid\xo) = \prod_i \mathrm{Cat}(x_t^{(i)}; \alpha_t x_0^{(i)} + (1-\alpha_t)\mathbf{e}_\mask)$ with $\alpha_t$ monotone from $\alpha_0{=}1$ to $\alpha_1{=}0$; the denoiser $p_\theta(\xo\mid\xt)$ is a bidirectional transformer trained on the masked ELBO~\citep{sahoo2024mdlm}, and the sampler unmasks progressively over $T$ steps. We write $d$ for hidden width, $K$ for the layers in the shared block, $L$ for the loop budget, and $(K\otimes L)$ for a $K$-layer block applied $L$ times, mirroring~\citet{saunshi2025looped}.

\textbf{Architecture.}
A standard \MDM{} applies a $KL$-layer transformer once and decodes from its final hidden state. Our new R-\MDM{} instead applies a $K$-layer block $f_\theta:\R^{n\times d}\!\to\!\R^{n\times d}$ with shared weights $L$ times (Figure~\ref{fig:main}). Starting from $\mathbf{h}^{(0)} = \mathrm{Norm}(E_\theta(\xt) + P_\theta(\xt))$, the process computes
\begin{equation}
  \mathbf{h}^{(\ell)} = f_\theta\big(\mathrm{Norm}(\mathbf{h}^{(\ell-1)} + \mathbf{v}^{(\ell)})\big),
  \qquad
  \hat{\mathbf{y}}^{(\ell)} = W_\theta\,\mathrm{Norm}\big(\mathbf{h}^{(\ell)}\big),
  \qquad \ell = 1,\dots,L,
  \label{eq:hloop}
\end{equation}
with $E_\theta$ the token embedding, $P_\theta$ the positional encoding, $\mathrm{Norm}$ RMS normalization~\citep{zhang2019rmsnorm}, and $W_\theta$ a shared LM head producing logits after \emph{every} loop. 

\begin{wrapfigure}{r}{0.45\textwidth}
    \vspace{-0.5cm}
    \centering
    \resizebox{0.43\textwidth}{!}{\input{figures/tikz/fig_scaling_axis}}
    \vspace{-0.2cm}
    \caption{\textbf{Recursion is a second scaling axis.} Adding layers moves a model diagonally, buying compute with parameters. Looping a $K$-layer block $L$ times moves it straight up: it matches the compute of a $KL$-layer model at the same FLOPs per evaluation, while holding $L\times$ fewer parameters.}
    \label{fig:main}
    \vspace{-0.6cm}
\end{wrapfigure}
Since $f_\theta$ and $W_\theta$ are identical across loops, the model has exactly the number of parameters of a single-pass $(K\otimes1)$ baseline, and the only information carried between loops is $\mathbf{h}^{(\ell)}$; a $(KL\otimes1)$ baseline instead applies $KL$ \emph{distinct} layers once, at identical FLOPs per forward pass and $L\times$ the parameters. Three design choices appear at \emph{training} time when adding recursive depth to standard \MDM{}s (Appendices~\ref{app:rec_steps_ablation}--\ref{app:loss_mode_ablation}): \textbf{(i)} an added \emph{step embedding} $\mathbf{v}^{(\ell)} = W_2\sigma(W_1 s_\ell + b_1)$ of the normalized loop progress $s_\ell = (\ell-1)/(L-1)$; \textbf{(ii)} the supervision scheme over the $L$ logit sets produced; and \textbf{(iii)} how many loops $L_t$ to run during training. 
We ablate these choices in Appendix~\ref{app:algorithms} and, based on the results, use learned step embeddings, masked cross-entropy averaged across all loops (as in deep supervision~\citep{lee2014deeplysupervisednets}), and reverse-curriculum annealing or fixed training depth $L_t$. All other hyperparameters are given in Appendix~\ref{app:algorithms}.

\begin{contributionbox}
\textbf{Contribution} R-\MDM{}s reuse a shared $K$-layer block across $L$ loops, increasing effective depth without increasing the parameter count. This introduces recursive depth as a distinct scaling axis for \MDM{}s.
\end{contributionbox}

\vspace{-0.1cm}
\subsection{Recursive depth is a parameter-efficient scaling axis}
\label{sec:rmdm_results}

We evaluate recursive depth as a scaling axis. 
We focus on two structured generation tasks that allow controlled comparisons across recursion depth and denoising budget: \textbf{Sudoku} ($1.8$M $9{\times}9$ boards and $100$k synthetic $25{\times}25$ boards), evaluated by Valid Puzzle Rate (VPR)~\citep{he2026latent}, and \textbf{Countdown}, with $100$k examples for each operand count $k\in\{3,4,5\}$, evaluated by Reaches Target Rate (RTR)~\citep{yao2023tree,gandhi2024streamsearchsoslearning}. 
Throughout this section, recursion matches the training setup: every denoising step runs $L=L_t=L_s$ loops, where $L_s$ is the inference-time loop depth, $T$ is fixed, and we commit the most confident tokens (alternative decoders are compared in Section~\ref{sec:decoding}). Dataset details, metrics, and hyperparameters are given in Appendices~\ref{app:datasets} and~\ref{app:algorithms}.

\textbf{Iso-parameter.}
At fixed parameter count, increasing $L$ consistently improves quality, particularly when the denoising budget is small. This comparison adds compute but keeps the number of learned parameters fixed. For example, on Countdown-3 at $T{=}10$, $L{=}3$ reaches $92.4\%$ RTR, compared with $59.4\%$ for the single-pass model. The effect is larger on the more difficult Countdown-5 task: at $T{=}30$, $L{=}10$ reaches $56.4\%$ RTR, compared with $15.6\%$ for the baseline. On Sudoku, the same pattern is visible in the full sweep (Appendix~\ref{app:sudoku_full}): at $T{=}5$, VPR increases from $65.2\%$ for the baseline to $93.6\%$ with $L{=}5$, while at $T{=}40$ it goes from $95.8\%$ to $98.0\%$. By refining its hidden state before committing tokens, R-\MDM{} can reach higher performance.

\begin{figure*}[t]
    \centering
    \includegraphics[width=\textwidth]{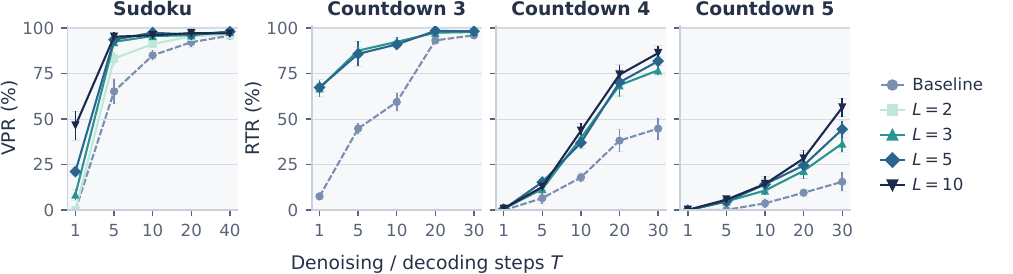}
    \vspace{-0.5cm}
    \caption{Effect of recursive refinement depth ($L$) on task performance for Sudoku $9\times9$ and Countdown across different target lengths (3, 4, and 5 digits). Increasing $L$ consistently improves success rates, with the strongest gains for more difficult problems.}
    \label{fig:countdown_rq1}
    \vspace{-0.25cm}
\end{figure*}
\begin{figure*}[t]
    \centering
    \includegraphics[width=\textwidth]{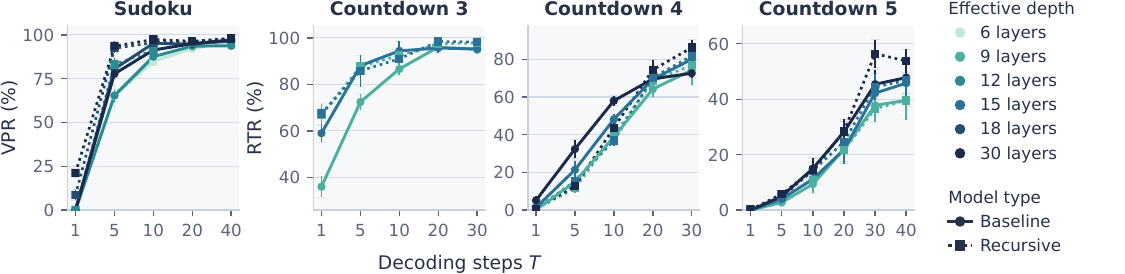}
    \vspace{-0.5cm}
    \caption{Effect of recursion at matched effective model depth (iso-FLOP) for Sudoku $9\times 9$ and Countdown tasks of different target length (3, 4, and 5 digits). Recursion matches or exceeds non-recursive baselines of equal effective depth while using $L\times$ fewer parameters.}
    \label{fig:isoFLOP}
    \vspace{-0.6cm}
\end{figure*}

\textbf{Iso-FLOP.}
To isolate the effect of weight sharing, we compare $(K\otimes L)$ R-\MDM{} with a $(KL\otimes1)$ baseline matched in effective depth and gradient updates; the recursive model uses $L\times$ fewer parameters (Appendix~\ref{app:algorithms}; Figure~\ref{fig:isoFLOP}). Recursion matches or outperforms the deeper baselines: on Sudoku at $T{=}5$, $(6\otimes5)$ achieves $93.6\%$ VPR with $10.8$M parameters versus $77.8\%$ with $53.1$M; on Countdown-4 at $T{=}20$, $(3\otimes5)$ and $(3\otimes10)$ reach $70.2\%$ and $74.4\%$ RTR, versus $69.4\%$ and $69.6\%$, with $5$--$10\times$ fewer parameters. Recursion also benefits beyond depth alone: at $T{=}40$, increasing Sudoku depth from $18$ to $30$ layers raises VPR only $96.6\%\to97.2\%$, whereas increasing $L$ from $1$ to $5$ at depth $6$ gives $95.8\%\to98.0\%$.
This suggests that iterative refinement of the masked state provides a benefit beyond simply increasing depth.

Two further results reinforce this picture. On $25\times25$ Sudoku, recursion is most valuable with high masking: at $90\%$ masking and $T{=}100$, VPR rises from $2.0\%$ to $52.2\%$ with $L{=}3$ (Appendix~\ref{app:sudoku25}). Training and sampling depths can differ: a model trained with $L_t$ loops can be sampled with $L_s\ne L_t$ (Appendix~\ref{app:crossrec_table}). Finally, Figure~\ref{fig:denoising-recursive-steps} fixes the total number of forward passes, and compares allocating compute to recursion or denoising. Recursion can be more effective and raise the performance ceiling; the two costs are equal but their effects are not. In Section~\ref{sec:analysis} we explore why.

\begin{takeawaybox}
 \textbf{Takeaway} These results establish recursion as a parameter-efficient scaling mechanism for \MDM{}s, with the largest gains at small denoising budgets and on structured tasks with explicit multi-step dependencies.
\end{takeawaybox}

\section{Benefits of recursion at test time}
\label{sec:analysis}

The results in Section~\ref{sec:rmdm} validate that additional recursive steps can improve the quality of generations in an R-\MDM{}. They have also suggested that while the cost of an additional denoising step and an additional recursive step are the same (effectively implying one more forward pass through the model), their effects can differ. In this Section, our goal is to better understand the interplay between the denoising and recursive steps and in doing so: \textbf{(i)} identify which types of errors each of these modes of computation can reduce (Section~\ref{sec:analysis_theory}), \textbf{(ii)} investigate why recursion is most beneficial at small \(T\) (Section~\ref{sec:analysis_empirical}).

\vspace{-0.1cm}
\subsection{A dependence--fidelity view of parallel decoding}
\label{sec:analysis_theory}

A decoding run maintains a partially masked state $s_t=(x_{O_t},M_t)$ with committed positions $O_t$ and masked positions $M_t$. At step $t$ it runs $\ell_t\le L$ loops while holding $s_t$ fixed, reads the marginals $q^{(\ell)}_i(v\mid s_t)=\operatorname{softmax}(W_\theta\mathbf{h}^{(\ell)}_i(s_t))_v$ for $i\in M_t$, then chooses a commit set $U_t\subseteq M_t$ and samples those positions \emph{independently}; we count both forms of computation by function evaluations $\NFE=\sum_t\ell_t$. An ideal step would draw $x_U$ from the true conditional joint $p^\star(x_U\mid s)=\pidata(x_U\mid x_O)$; sampling from a product of the model's own marginals instead makes two separable bets: one about each position alone, one about the positions being unrelated. 
As shown by \citet{benhamu2025ebsampler}, the error decomposes along these two axes; in recursive decoding, this decomposition separates what loops and denoising steps can each reduce.

\begin{lemma}[Dependence--fidelity decomposition]
\label{lem:decomp}
For any state $s$, commit set $U$, and strictly positive marginals $q_i$ (as softmax outputs always are),
\begin{equation}
    \KL\!\Big(p^\star(x_U \mid s)\ \Big\Vert \prod_{i\in U} q_i\Big)
    \;=\;
    \underbrace{\TC\big(p^\star(x_U \mid s)\big)}_{\textstyle \Ddep(U \mid s)}
    \;+\;
    \underbrace{\sum_{i \in U} \KL\big(p^\star_i \,\Vert\, q_i\big)}_{\textstyle \Dfid(U \mid s,\, q)} ,
    \label{eq:decomp}
\end{equation}
where $\TC(p) = \sum_{i\in U} H(p_i) - H(p) \ge 0$ is the total correlation of the true joint distribution.
\end{lemma}

\begin{wrapfigure}{r}{0.38\textwidth}
    \vspace{-0.4cm}
    \centering
    \includegraphics[width=0.38\textwidth]{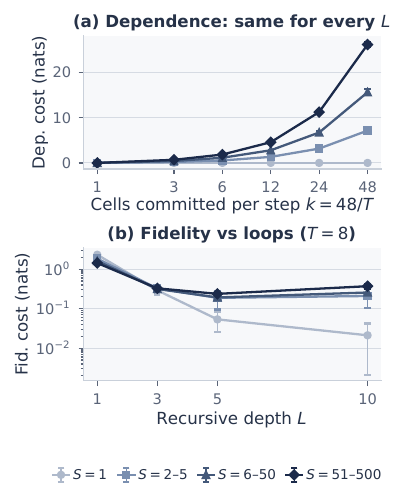}
    \vspace{-0.65cm}
    \caption{\textbf{Each term responds to one lever.} Exact costs along the oracle process. (a) $\sum_t\Ddep$ grows with $S$ and commit size, and is independent of $L$ by construction, since the commit order is fixed. (b) $\sum_t\Dfid$ decreases with $L$.}
    \label{fig:dep_fid}
    \vspace{-0.6cm}
\end{wrapfigure}
In simpler terms, $\Dfid$ measures how wrong the model is about each committed position considered alone, and $\Ddep$ quantifies what it forfeits by treating the positions it commits together as independent. The dependence term is not model-dependent: it is a property of the task and of \emph{which} positions are committed together, and it vanishes whenever those positions are conditionally independent, in particular whenever only a single token is committed in each step (see proofs and details in Appendix~\ref{app:proofs}).

This decomposition isolates the role of denoising and recursive steps. \textbf{Extra denoising steps decrease $\Ddep$.} Splitting one commit $U$ into consecutive commits $A$ and $B$ lowers the dependence cost by exactly $\sum_{i\in B} I(x_i;x_A\mid s)$ (Corollary~\ref{cor:refine}). Thus, reducing $\Ddep$ can be achieved by committing fewer or less dependent positions at a time: a decision of the denoising schedule, not the loop. \textbf{At a fixed commit set, extra loops can only act on $\Dfid$.} A loop leaves $s$ and $U$ untouched and changes only the marginals, so it cannot affect $\Ddep$. However, when $U$ itself is prediction-dependent, loops can also change $U$ and hence affect $\Ddep$; Section~\ref{sec:analysis_empirical} fixes the commit order to isolate this.

Altogether, the decomposition explains the complementary roles of recursive and denoising steps, and why the gains of Section~\ref{sec:rmdm_results} depend so strongly on $T$.

\subsection{Empirical evaluation in controlled settings}
\label{sec:analysis_empirical}

To empirically evaluate whether loops reduce $\Dfid$, and how much error remains once they do, we require access to the true distribution $p^\star$. The Sudoku puzzles of Section~\ref{sec:rmdm_results} cannot answer it: each board has a unique solution, so $p^\star$ is a point mass, $\Ddep\equiv0$, and only the fidelity error remains. We therefore construct boards with multiple valid completions, grouped by their number of solutions $S$ ($1$, $2$--$5$, $6$--$50$, $51$--$500$; 50 boards each). Since all solutions are known, $p^\star$, $\Ddep$, and $\Dfid$ can be computed exactly. 
To stay within the assumptions of Lemma~\ref{lem:decomp}, we sample from the model's untruncated marginals (Remark~\ref{rem:ac_free}) and fix a random commit order before decoding, making the commit sets independent of $L$. Runs terminate at the first rule violation, either a \emph{parallel} clash within a commit step or an \emph{old-cell} clash with a previously committed cell (Appendix~\ref{app:multisol}).

\begin{figure}[t]
    \centering
    \includegraphics[width=\textwidth]{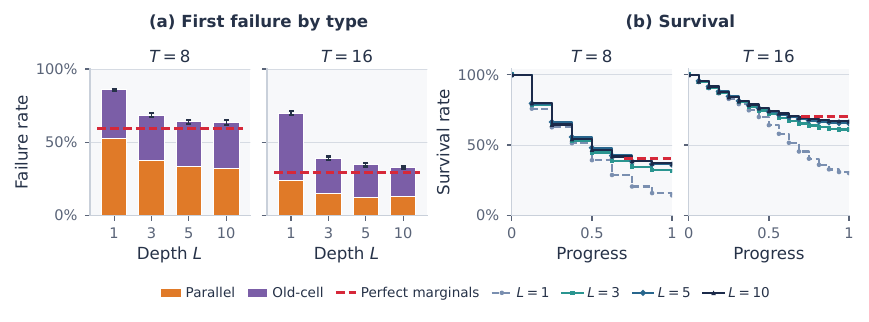}
    \vspace{-0.9cm}
    \caption{\textbf{Recursion removes failures down to the dependence floor.} Multi-solution boards ($S{=}6$--$500$). (a) First rule violation by type. (b) Survival against generation progress. Red: final failure (a) and survival (b) of a sampler with perfect marginals, $q_i{=}p^\star_i$, committing the same cells.}
    \label{fig:errors_survival}
    \vspace{-0.5cm}
\end{figure}

We first verify whether indeed Lemma~\ref{lem:decomp} correctly predicts the dependence of $\Ddep$ on the number of denoising steps, and of $\Dfid$ on the number of recursive steps. We find that indeed, the dependence cost increases as we increase the set of tokens committed at each step, hence lowering the number of denoising steps (Figure~\ref{fig:dep_fid}a). This effect is particularly pronounced for Sudoku boards where multiple solutions exist ($S>1$).  In turn, additional recursive steps reduce the fidelity error $\Dfid$ by $6$--$43\times$ from $L{=}1$ to $L{=}5$ (Figure~\ref{fig:dep_fid}b). Beyond $L{=}5$, the benefits of the extra loops seem to plateau, indicating that additional loops do not always lead to better results, and inspiring the design of our adaptive loop rules in \Cref{sec:decoding}.

This difference between denoising and recursive steps is also reflected in the error rates. On the multi-solution boards in Figure~\ref{fig:errors_survival}a, increasing $L$ eventually brings the failure rate close to that obtained with perfect marginals under the same commit schedule ($64\%$ vs.~$59\%$ at $T{=}8$; $35\%$ vs.~$30\%$ at $T{=}16$). At $L{=}5$, $83$--$88\%$ of the remaining parallel clashes are dependence errors, involving values that are individually valid but jointly impossible. Thus, once this dependence floor is reached, reducing the commit size is more effective than adding loops; parallel-clash rate is consequently a useful proxy for $\Ddep$ when $p^\star$ is spread (Remark~\ref{rem:metrics}). 
The lower error rate also translates into substantial gains in survival, the fraction of runs with no violation (Figure~\ref{fig:errors_survival}b), which more than doubles from $14\%$ to $36\%$ at $T{=}8$ and from $30\%$ to $65\%$ at $T{=}16$, approaching the perfect-marginal levels of $41\%$ and $70\%$.
We note that extra loops help particularly halfway through the denoising process: early on, few cells are committed and the true marginals are spread, leaving little for sharper predictions to fix, while late on predictions are already sharp.

Together, these results explain the behavior in Section~\ref{sec:rmdm_results}. On unique puzzles, errors at small $T$ are primarily fidelity errors amplified by large commits, so recursion substantially improves performance. With multiple valid completions, recursion eventually reaches a dependence-error floor; beyond that point, smaller commit steps are more effective, because they help to reduce the token dependencies. Section~\ref{sec:decoding} turns this tradeoff into a practical decoder by adaptively choosing both how many recursive steps to take, and which tokens to commit once we stop looping.

\begin{takeawaybox}
\textbf{Takeaway} Recursion reduces fidelity error but cannot remove dependence error. Once the dependence floor is reached, smaller commit steps are more effective.
\end{takeawaybox}

\section{Towards a Practical Recursive Decoder}
\label{sec:decoding}

Section~\ref{sec:analysis} showed that additional recursive steps can reduce the fidelity error ($\Dfid$) by improving the model marginals at a fixed masked state, but cannot reduce the dependence error ($\Ddep$) introduced by committing tokens jointly. Their benefit therefore varies throughout the denoising process: additional loops are useful while the marginals continue to improve, but provide little benefit once these improvements saturate or dependence error becomes dominant. Using a fixed number of recursive steps at every masked state may therefore waste computation.

Prior work has studied adaptive \MDM{} decoding by varying which and how many tokens are committed after each model evaluation~\citep{benhamu2025ebsampler,zhang2026generationorder}. \citet{lee2026loopmdm} further show that recursive depth can be chosen adaptively at a fixed denoising budget. To the best of our knowledge, however, no prior method jointly adapts recursive computation and token commitment. To combine these two adaptive axes, we cast an R-\MDM{} decoder as a pair of rules (one determining the number of recursive steps and the other selecting the commit set) and use predictive entropy as a shared observable signal for both.

\subsection{A two-axis decoding framework}
\label{sec:decoding_rules}

\textbf{Decoding space.}
A decoder \meth{[loop]\_[commit]} pairs a loop-stopping rule, which sets $\ell_t\le L$ at state $s_t$, with a commit rule, which selects $U_t\subseteq M_t$. At least one position is always committed per step so that decoding terminates, and compute is quantified as $\NFE=\sum_t\ell_t$. 

\begin{figure*}[t]
    \centering
    \resizebox{\textwidth}{!}{\input{figures/tikz/fig_decoding_lattice}}
    \caption{\textbf{Decoding an R-\MDM{} has two axes of refinement.} Denoising steps run left to right and commit tokens, changing the masked state; recursive loops run top to bottom and sharpen the predictions while the state is held fixed. A decoder therefore makes two decisions per step: how many loops to spend, and which positions to commit.}
    \label{fig:decoding_lattice}
    \vspace{-0.2cm}
\end{figure*}

\textbf{Entropy as a shared signal.}
Designing an adaptive decoder requires an observable signal for two decisions: whether to run another recursive step and which tokens to commit. We use predictive entropy, which summarizes the full marginal distribution over the vocabulary and, in R-\MDM{}, can be computed for every masked position after each loop without additional model evaluations. Although recursive refinement acts on the fidelity term $\Dfid$, this quantity depends on the unknown true marginals and is therefore unavailable at inference time. Instead, we monitor whether the model predictions continue to sharpen through the mean entropy over the masked positions and its reduction between consecutive loops:
\begin{equation}
\label{eq:entgain}
\bar H(q^{(\ell)}) =
\frac{1}{|M_t|}
\sum_{i \in M_t} H\bigl(q_i^{(\ell)}\bigr),
\qquad
\overline{\Delta H}^{(\ell)}_t =
\bar H(q^{(\ell-1)})-\bar H(q^{(\ell)}).
\end{equation}
A large positive entropy gain indicates that an additional loop is still sharpening the predicted distributions, whereas a small gain suggests diminishing returns.

Entropy also provides a principled signal for deciding which tokens to commit, as previously explored by \citet{benhamu2025ebsampler} and \citet{cai2026confidence}. \citet{benhamu2025ebsampler} bound the dependence cost of a commit set by the sum of the true marginal entropies of its positions, and use the model's predictive entropies in their place as an approximation. We make this approximation explicit (Proposition~\ref{prop:entropy_cert}): substituting model entropies adds a correction term controlled by $\Dfid$ and vanishes as the fidelity error decreases. Recursive refinement therefore makes model entropy a more reliable guide, while selecting low-entropy positions limits the dependence cost of each commit.

\begin{figure*}[t]
    \centering
    \includegraphics[width=\textwidth]{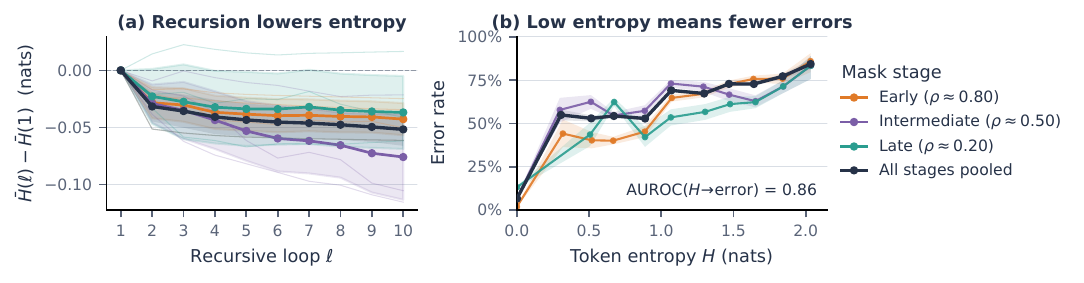}
    \vspace{-0.7cm}
\caption{\textbf{Recursion sharpens predictions and reduces errors.} (a) Change in mean predictive entropy of masked cells relative to the first loop, entropy decreases monotonically. (b) Error rate versus predictive entropy. Low-entropy predictions are more reliable, with error increasing steadily with entropy (AUROC $=0.86$).}
    \label{fig:entropy_study}
    \vspace{-0.6cm}
\end{figure*}

\textbf{Empirical validation.}
Using entropy for these two decisions requires two empirical properties: its reduction across loops should reflect continued refinement, and its value at each position should predict the likelihood of an error. Figure~\ref{fig:entropy_study} tests both on Sudoku. At fixed masked states, mean entropy decreases with recursive depth. Pooled across mask fractions ($\rho_t=|M_t|/n$), the first additional loop accounts for 60\% of the total reduction over ten loops, consistent with analyses of looped language models
\citep{blayney2026mechanisticlooped}. Later loops can still have a significant effect at intermediate states ($\rho_t\approx0.50$), but little effect on early and late states, suggesting a stopping criterion. Because a loop may sharpen some positions while blurring others, $\overline{\Delta H}^{(\ell)}_t$ is a practical stopping signal rather than a convergence certificate. At the token level, entropy also predicts correctness: the pooled error rate increases from 7\% near zero entropy to 85\% at high entropy, with an AUROC of 0.86. These results support predictive entropy as a shared signal for loop stopping and token commitment.

\textbf{Our decoder.}
We combine these two adaptive rules in the \entent{} decoder (Algorithm~\ref{alg:entent}). At state $s_t$, its \emph{loop rule} stops once the entropy gain of \Eqref{eq:entgain} falls below a threshold, $\overline{\Delta H}^{(\ell)}_t<\lambda$ (or when the loops reach the upper limit $L$). In the spirit of EB-Sampler~\citep{benhamu2025ebsampler}, its \emph{commit rule} greedily commits the lowest-entropy positions satisfying $\sum_{i\in U_t}H(q^{(\ell_t)}_i)\leq\delta$, which bounds $\Ddep$ by $\delta$ up to a correction term.
Both rules use the same predictive entropies, and neither requires specifying $T$. 

As baselines, we use the non-adaptive \meth{fixed} loop rule ($\ell_t=L$) and the standard \meth{uniform} commit rule, which commits the $|M_0|/T$ most confident positions. We also consider \meth{sched} (linear decay from $L$ to $1$) and the \meth{hidden} stopping rule, which stops when the relative hidden-state change $\Vert\mathbf{h}^{(\ell)}-\mathbf{h}^{(\ell-1)}\Vert/\Vert\mathbf{h}^{(\ell)}\Vert$ falls below $\varepsilon$~\citep{lee2026loopmdm}. The \meth{hidden} rule achieves similar quality at higher NFE cost (Appendix~\ref{app:decoding_comparison}), motivating our use of entropy. The alternative \meth{confidence} commit rule, which commits all positions with top-1 probability above $c$~\citep{chang2022maskgit}, is also evaluated. See further details in Appendix~\ref{app:decodingzoo} and the complete evaluation in Appendix~\ref{app:decoding_comparison}.

\begin{contributionbox}
    \textbf{Contribution} We introduce a two-axis decoder for R-\MDM{}s that adaptively decides \emph{how long to refine} and \emph{how much to commit}, using entropy as a common signal for both (Algorithm~\ref{alg:entent}).
\end{contributionbox}

\subsection{Empirical verification across structured generation tasks}
\label{sec:decoding_results}

\textbf{Tasks and metrics.} We evaluate the \entent{} decoder across four domains. We retain Sudoku and Countdown from the previous sections, reporting Valid Puzzle Rate and Reaches Target Rate, respectively, and add RNA infilling and executable program generation on TinyGSM.

RNA infilling is a natural test of the mechanisms of Section~\ref{sec:analysis} because base pairing creates long-range dependencies between nucleotide positions, challenging decoders that commit multiple tokens in parallel. We randomly mask \(40\%\) of held-out RNA sequences of length \(n\in\{76,110,254\}\) from RNAcentral~\citep{rnacentral2026}, using the original complete sequence as a reference. Because raw folding energy depends on sequence length and composition, we evaluate each completion using its ViennaRNA~\citep{lorenz2011viennarna} minimum-free-energy $Z$-score, standardized against length- and composition-matched shuffles. We report the paired gap ($\Delta Z=Z(\text{infilled})-Z(\text{native})$), which controls for sequence-specific difficulty; $\Delta Z=0$ matches the reference folding score, and lower is better (see Appendix~\ref{app:rna_datametrics} for further details on metrics).

TinyGSM, our largest setting (161M parameters, 512 tokens, 49k vocabulary), requires executable Python solutions to grade-school math problems~\citep{liu2023tinygsm}, whose correctness is decided only by running the program. We report GSM8K execution accuracy~\citep{cobbe2021gsm8k} and further details in Appendix~\ref{app:tinygsm_datametrics}.

Because \entent{} varies the number of recursive steps at inference time, we use checkpoints trained with variable loop depths: (\emph{Schedule} $10\to1$) for Sudoku and Countdown, and (\emph{Schedule} $5\to1$) for RNA and TinyGSM (see ablation in Appendix~\ref{app:rec_steps_ablation}). The non-recursive baseline ($L{=}1$) is a separate model with the same architecture, trained without recursion.

\begin{figure*}[t]
    \centering
    \includegraphics[width=\textwidth]{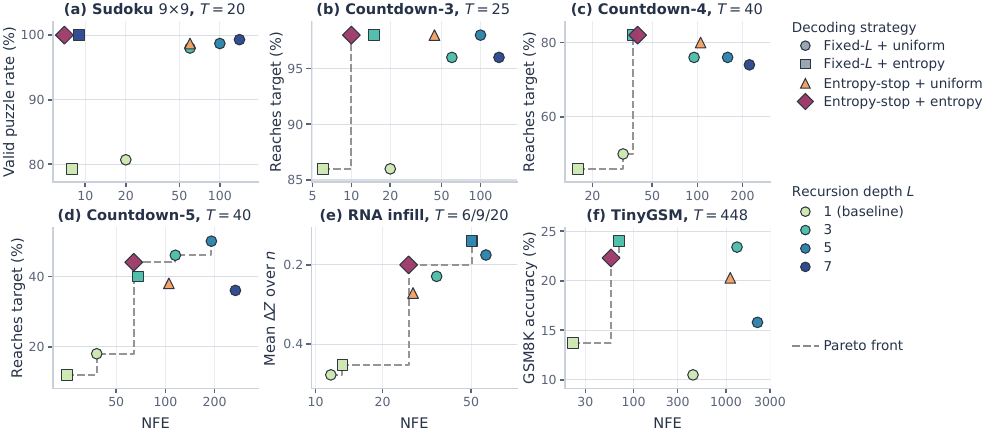}
    \vspace{-0.5cm}
\caption{\textbf{Decoding results across tasks.} Quality vs.\ realized \NFE{} per decoder (marker) and loop budget (color); \entent{} (diamonds) lies on or near the Pareto front on every task. Metrics: VPR (Sudoku), RTR (Countdown), $\Delta Z$ (RNA; lower is better), GSM8K accuracy (TinyGSM).}
    \label{fig:empirical_results}
    \vspace{-0.6cm}
\end{figure*}

\textbf{Results.}
Figure~\ref{fig:empirical_results} compares four decoders on the quality--$\NFE$ plane: \fixunif{}, the default of Section~\ref{sec:rmdm}; \fixent{} and \entunif{}, each of which makes only one axis adaptive; and our \entent{} (we consider alternative decoders in Appendix~\ref{app:decoding_comparison}). The best recursive result exceeds the non-recursive baseline in every domain: Sudoku from $80.7\%$ to $100\%$ VPR, Countdown-4 from $50.0\%$ to $82.0\%$ RTR, RNA from $\Delta Z=0.453$ to $0.140$ (closing $69\%$ of the gap to the native fold), and TinyGSM from $13.7\%$ to $24.0\%$ accuracy.

Under \meth{fixed} stopping, $\NFE$ grows linearly with $L$ even once extra loops no longer change the prediction; schemes which make one part of the decoding process adaptive already help to lower the computational cost, at small costs to the quality. However, our doubly-adaptive method lies on or near the Pareto front across all datasets, enabling even further efficiency gains.

We further observe that on long sequences, making the commit rule adaptive explains most of the efficiency gains: on TinyGSM, \fixent{} at $L{=}3$ matches the best \fixunif{} accuracy ($24.0\%$ vs.\ $24.2\%$) with $69$ instead of $768$ evaluations, while loop stopping trims only half a loop per step (Appendix~\ref{app:tinygsm_full}). However, when allowing for large $L$, loop stopping is necessary for efficiency: on Countdown-4, the cost of \fixent{} grows from $37$ to $98$ evaluations between $L{=}3$ and $L=7$, while \entent{} stays near $40$.

\begin{takeawaybox}
\textbf{Takeaway} Entropy effectively guides both decoding axes: \entent{} pays for loops only while predictions still change, keeps cost flat as $L$ grows, and matches the best quality on Sudoku and Countdown-3/4 at a fraction of the compute. 
\end{takeawaybox}

\section{Conclusions, Limitations, and Future Work}
\label{sec:conclusions}

In this work, we studied recursion as a scaling mechanism for masked diffusion models. Applying the same transformer block $L$ times within each denoising step increases effective depth without increasing parameter count. Across Sudoku and Countdown, recursive depth improves quality at fixed parameter count, matches substantially larger non-recursive models at matched FLOPs, and can reduce the number of denoising steps required to reach a given quality.

Our analysis interprets these gains through the different roles of recursive loops and denoising steps. Recursion refines model marginals at a fixed masked state, while denoising steps change that state by committing tokens; accordingly, the two axes affect different components of parallel-decoding error.  Recursive refinement reduces predictive entropy and conflicts among simultaneously decoded tokens, especially when each denoising step must resolve many dependencies, but it cannot remove the dependence error among tokens committed together. These observations motivate adaptive decoding, where loop stopping and token commitment are selected from state-dependent signals. The resulting methods improve the quality--compute trade-off across all four domains.

This work establishes the viability of recursive MDMs on structured tasks and provides a framework for scaling their effective depth without increasing parameter count. Future work can build on this framework by scaling the paradigm to larger models, and test whether the gains offered by our recursive framework extend to less structured settings such as open-ended language generation.



\subsubsection*{Acknowledgments}
The authors thank Vincent Jung for insightful discussions and assistance with the RNA experiments. ACC was supported by the ELSA Mobility Program for PhD Students and in part by the EPFL AI Center and Swiss AI Initiative PhD Fellowship. JP's PhD studentship is funded by AstraZeneca.

\subsection*{AI use statement}

In this work, we used generative AI tools to help formulate mathematical claims, to provide guidelines for proving those claims and assist in the writing of proofs, to provide feedback on our experiments, and to implement methods in code. We have not used generative AI tools to help develop theoretical models or conceptual frameworks, to propose or refine hypotheses, to design our research methodology, to clean and reformat datasets, to support qualitative and thematic data analysis, to suggest a structure for the paper, to generate synthetic datasets, or to interpret results, and assistance with translation are not applicable to this work. Additionally, we used generative AI tools to create and edit software code, and to edit the paper to improve readability. We have reviewed all AI-assisted work: all mathematical claims and their proofs were checked in full by the authors, AI-assisted code was verified and tested for correctness by two authors, and all AI-edited text was reviewed by all the authors to confirm that it accurately reflects our intended meaning and claims. We take responsibility for the final content of this work, including text, claims or artifacts produced with the aid of generative AI.

\subsection*{Ethics statement}

The primary objective of this paper is to advance discrete generative models with a more parameter-efficient architecture. By reducing the model size and number of denoising steps required to achieve high-quality generation, recursive masked diffusion models could lower the computational cost of deploying sequence generation systems, with potential benefits in resource-constrained settings. While more capable and accessible generative models carry the general dual-use concerns associated with language generation research, we do not foresee any immediate societal concerns specific to the methodology proposed here.

\bibliography{references}
\bibliographystyle{iclr2027_conference}

\clearpage
\appendix

\doparttoc 
\faketableofcontents 

\addcontentsline{toc}{section}{Appendix} 
\part{{\Large Appendix}} 
\parttoc 
\newpage

\section{Proofs}
\label{app:proofs}

\vspace{-0.15cm}

This appendix proves the results of Section~\ref{sec:analysis_theory}: Lemma~\ref{lem:decomp}, Proposition~\ref{prop:entropy_cert} and Theorem~\ref{thm:floor}, and states their scope (Appendix~\ref{app:proofs_scope}). Lemma~\ref{lem:decomp} and Theorem~\ref{thm:floor} follow the decomposition of \citet{benhamu2025ebsampler}, extended to policies that also choose the loop count.

\vspace{-0.15cm}

\subsection{Notation and setting}
\label{app:proofs_notation}

Fix a partially masked state $s = (x_O, M)$ with committed positions $O$ and masked positions $M$, as in Section~\ref{sec:rmdm_def}. Throughout, states satisfy $\pidata(x_O)>0$, so that the conditional below is defined; this holds at every state visited by the oracle process $P$ below, but not after a wrong commit by the sampler. Let $\pidata$ be the data distribution and let
\begin{equation}
    p^\star(x_U \mid s) \;=\; \pidata\big(x_U \mid x_O\big), \qquad U \subseteq M,
\end{equation}
denote the true conditional joint over the commit set $U$ given the committed context, with position-wise marginals $p^\star_i(\cdot \mid s)$. All entropies are Shannon entropies in nats over the finite vocabulary $\Vocab$, $m = |\Vocab|$, and $\TC$ denotes total correlation (multi-information),
\begin{equation}
    \TC(p) \;=\; \sum_{i \in U} H(p_i) - H(p) \;=\; \KL\Big(p \,\Big\Vert\, \prod_{i\in U} p_i\Big) \;\ge\; 0 ,
    \label{eq:tc_def}
\end{equation}
which is zero if and only if $p$ factorizes across $U$.

\vspace{-0.15cm}

\paragraph{The two processes.}
A \emph{decoding policy} is a pair of maps $s \mapsto \ell(s) \in \{1,\dots,L\}$ and $s \mapsto U(s) \subseteq M(s)$ with $U(s) \neq \emptyset$. Both may depend on the model's predictions at $s$, since $\theta$ is fixed and those predictions are themselves functions of $s$. More generally, both may depend on the step index, on the history $(s_0,\dots,s_t)$, and on randomness $\xi$ drawn before decoding, such as a random commit order; reading $s_t$ as that history and conditioning on $\xi$ keeps both chains below Markov, and every result holds unchanged. This covers every decoder of Section~\ref{sec:decoding}, including the step-indexed \meth{uniform} and \meth{sched} rules, as well as the split schedule of Corollary~\ref{cor:refine} and the fixed commit order of Section~\ref{sec:analysis_empirical}. Given $x_O$ and a policy we compare two chains started at $s_0 = (x_O, M)$:
\begin{itemize}[leftmargin=1.4em, itemsep=1pt, topsep=2pt]
    \item the \textbf{oracle process} $P$, which at state $s_t$ sets $U_t = U(s_t)$ and draws $x_{U_t} \sim p^\star(\cdot \mid s_t)$;
    \item the \textbf{sampler} $Q$, which at $s_t$ sets $U_t = U(s_t)$, $\ell_t = \ell(s_t)$ and draws $x_{U_t} \sim \prod_{i \in U_t} q^{(\ell_t)}_i(\cdot \mid s_t)$.
\end{itemize}
Both chains halt when $M_t = \emptyset$, after a number of steps $\tau \le |M|$ determined by the state sequence; terminal states contribute nothing to any sum below.

\begin{remark}[$\KL(P\Vert Q)$ is a statement about outputs]
\label{rem:bijection}
Under $P$, $\prod_t p^\star(x_{U_t}\mid s_t)=\pidata(x_M\mid x_O)$ for any adapted commit order, and the trajectory is a deterministic function of $x_M$ and vice versa, so $\KL(P\Vert Q)$ equals the KL between output laws~\citep{benhamu2025ebsampler}. It measures coverage of $\pidata(\cdot\mid x_O)$ and is less sensitive to off-support samples~\citep{zhang2026generationorder}, which Section~\ref{sec:analysis_empirical} measures via violation rates.
\end{remark}

\vspace{-0.15cm}

\subsection{Proof of Lemma~\ref{lem:decomp}}
\label{app:proof_decomp}

\begin{proof}
Write $p = p^\star(\cdot \mid s)$ and $q_U = \prod_{i\in U} q_i$. The hypothesis is that each $p^\star_i$ is absolutely continuous with respect to $q_i$, \ie $q_i(v) = 0 \Rightarrow p^\star_i(v) = 0$; this is what makes $\KL(p^\star_i \Vert q_i)$, and hence every term below, finite. Expanding,
\begin{align}
    \KL(p \Vert q_U)
    &= \sum_{x_U} p(x_U) \log \frac{p(x_U)}{\prod_{i\in U} q_i(x_i)} \notag\\
    &= -H(p) - \sum_{i \in U} \sum_{x_i} p_i(x_i) \log q_i(x_i)
    && \text{(marginalizing all $j \ne i$)} \notag\\
    &= -H(p) + \sum_{i \in U} \Big[ H(p_i) + \KL(p_i \Vert q_i) \Big]
    && \text{($\CE = H + \KL$)} \notag\\
    &= \underbrace{\Big[\textstyle\sum_{i \in U} H(p_i) - H(p)\Big]}_{= \TC(p) \,=\, \Ddep(U\mid s)} + \underbrace{\sum_{i\in U}\KL(p_i \Vert q_i)}_{= \Dfid(U \mid s, q)} ,
\end{align}
which is \Eqref{eq:decomp}. Both terms are non-negative: the first by \Eqref{eq:tc_def} and the second as a sum of KL divergences. \qedhere
\end{proof}

\begin{remark}[Which distribution the analysis refers to]
\label{rem:ac_free}
Throughout, $q_i$ denotes the \emph{denoiser marginal}, the softmax over the vocabulary in \Eqref{eq:hloop}. Since softmax outputs are strictly positive, $p^\star_i \ll q_i$ holds for any $p^\star$ and the hypothesis of Lemma~\ref{lem:decomp} is never binding for $q_i$. The predictive entropies used by the commit and stopping rules are also computed from $q_i$.

This is not the distribution the experiments sample from. At sampling time we apply temperature scaling and top-$k$ truncation (Table~\ref{tab:hp-choices}), giving a deployed law $\tilde q_i \neq q_i$ whose support is a strict subset of $\Vocab$ when $k < m$; for Countdown and TinyGSM, $k=1$. Whenever $p^\star_i$ places mass outside that support, $\KL(p^\star_i \Vert \tilde q_i) = \infty$, so the finite-KL statements below are results about the untruncated sampler that draws from $q_i$, not certificates for every decoding configuration we run. We keep the truncated samplers in the experiments because they are what one deploys, and read the analysis as describing the denoiser marginals they are built from.
\end{remark}

Two immediate consequences are used in the main text. First, $\Ddep(U \mid s)$ depends on $p^\star$, $s$ and $U$ only, so any operation that holds $s$ and $U$ fixed and modifies only the deployed marginals (in particular running additional recursive loops, which by construction leave the state and the commit set unchanged) changes the step error only through $\Dfid$. Second, if $|U| = 1$ then $\TC = 0$, so a single-position commit incurs no dependence cost at all; this is the formal version of the standard observation that the speed--quality trade-off of parallel \MDM{} decoding is governed by inter-token dependence~\citep{benhamu2025ebsampler, zhang2026generationorder, li2025convergence, chen2026optimalmdm}.

\subsection{The entropy certificate}
\label{app:proof_commit}

Section~\ref{sec:decoding_rules} uses the following bound to justify committing low-entropy
positions; we state it here in full, with the slack term that the main text describes
only qualitatively.

\begin{proposition}[Entropy bounds the dependence cost]
\label{prop:entropy_cert}
Fix a state $s$, a non-empty commit set $U$, a vocabulary of size $m \ge 2$, and denoiser marginals $q = q^{(\ell)}$; write $H_i = H(q_i)$ for the decoder's own predictive entropies. Then
\begin{equation}
    \Ddep(U \mid s)
    \;\le\;
    \sum_{i \in U} H_i \;-\; \max_{i \in U} H_i \;+\; \varepsilon ,
    \qquad
    \varepsilon \;\coloneqq\; \sum_{i \in U} \big| H(p^\star_i) - H_i \big| ,
    \label{eq:entropy_cert}
\end{equation}
where $\varepsilon$ is the total error in those entropies. If in addition $\KL(p^\star_i \Vert q_i) \le 2(1-1/m)^2$ for every $i \in U$, with $m = |\Vocab|$, then $\varepsilon \le |U|\,\omega\big(\Dfid(U \mid s, q)/|U|\big)$ for an increasing function $\omega$ with $\omega(0) = 0$, given explicitly below.
\end{proposition}

The function referred to in the statement is
\begin{equation}
    \omega(x) \;=\; f\big(\sqrt{x/2}\big), \qquad f(\eta) \;=\; \eta\log(m-1) + h(\eta),
    \label{eq:omega_def}
\end{equation}
with $h$ the binary entropy. We show below that $\omega$ is concave and increasing on $[0, 2(1-1/m)^2]$ with $\omega(0)=0$.

\begin{proof}
\textbf{Step 1: the bound in the true marginals.}
Joint entropy dominates any marginal entropy: for $j \in U$, $H(x_U) = H(x_j) + H(x_{U\setminus j}\mid x_j) \ge H(x_j)$, so $H(p^\star(x_U\mid s)) \ge \max_{i\in U} H(p^\star_i)$. Substituting into \Eqref{eq:tc_def},
\begin{equation}
    \Ddep(U\mid s) \;\le\; \sum_{i\in U} H(p^\star_i) - \max_{i \in U} H(p^\star_i).
    \label{eq:cert_true}
\end{equation}
This is the bound of \citet{benhamu2025ebsampler}.

\textbf{Step 2: transfer to the model's entropies.}
Write $a_i = H(p^\star_i)$ and $b_i = H_i = H(q_i)$, so that $\varepsilon = \sum_{i \in U}|a_i - b_i|$. The functional $F(z)=\sum_{i\in U}z_i-\max_{i\in U}z_i$ is $1$-Lipschitz with respect to $\|\cdot\|_1$: each subgradient has entries in $[0,1]$ and at least one zero entry. Therefore $F(a)\le F(b)+\|a-b\|_1=F(b)+\varepsilon$, i.e.
\begin{equation}
    \sum_{i\in U} a_i - \max_{i\in U} a_i
    \;\le\;
    \sum_{i\in U} b_i - \max_{i\in U} b_i + \varepsilon
    \;=\; \sum_{i\in U} H_i - \max_{i\in U} H_i + \varepsilon ,
\end{equation}
which combined with \Eqref{eq:cert_true} is \Eqref{eq:entropy_cert}. Note that Steps 1 and 2 are unconditional.

\textbf{Step 3: $\varepsilon$ is controlled by $\Dfid$.}
We assume the \emph{coordinate-wise} condition $D_i \le 2(1-1/m)^2$ for each $i \in U$, rather than the cumulative condition $\Dfid \le 2(1-1/m)^2$. The latter would force the average per-position error below $2/|U|$ nats and so would void the certificate precisely when commits are large, which is the regime of interest. Let $D_i = \KL(p^\star_i \Vert q_i)$ and $\eta_i = \TV(p^\star_i, q_i)$, so $\Dfid = \sum_{i\in U} D_i$. The sharpened Fannes inequality~\citep{Audenaert2007fannes} states that for distributions on an alphabet of size $m$ with total variation $\eta \le 1-1/m$, $|H(p)-H(q)| \le f(\eta)$. Pinsker's inequality gives $\eta_i \le \sqrt{D_i/2}$, and the hypothesis $D_i \le 2(1-1/m)^2$ gives $\sqrt{D_i/2} \le 1-1/m$ directly, so the inequality applies at each position. It applies in the useful direction because $f$ is non-decreasing on $[0,1-1/m]$: $f'(\eta) = \log(m-1) + \log\frac{1-\eta}{\eta}$ vanishes exactly at $\eta = 1-1/m$ and is positive to its left. (Note that $h$ alone is not monotone on this interval when $m>2$; it is the sum that is.) Hence $|H(p^\star_i) - H_i| \le f(\sqrt{D_i/2}) = \omega(D_i)$, and
\begin{equation}
    \varepsilon \;=\; \sum_{i\in U}\big|H(p^\star_i)-H_i\big|
    \;\le\; \sum_{i \in U} \omega(D_i)
    \;\le\; |U|\,\omega\Big(\tfrac{1}{|U|}\textstyle\sum_{i\in U} D_i\Big)
    \;=\; |U|\,\omega\Big(\tfrac{\Dfid}{|U|}\Big),
\end{equation}
the second inequality by Jensen. This is legitimate because $\omega$ is concave, being an increasing concave function ($f$) of a concave one ($x \mapsto \sqrt{x/2}$); it is increasing for the same reason, and $\omega(0) = f(0) = 0$. The argument $\Dfid/|U| = \tfrac{1}{|U|}\sum_i D_i \le \max_i D_i$ lies in $[0, 2(1-1/m)^2]$ by hypothesis, so Jensen applies on the interval where $\omega$ is concave. \qedhere
\end{proof}

The bound $\sum_i H_i - \max_i H_i$ is the criterion of EB-Sampler~\citep{benhamu2025ebsampler}, who use model entropies in place of true ones as an approximation valid when model error is negligible. The budget $\sum_{i\in U} H_i \le \delta$, which our \meth{entropy} commit rule uses and \citet{cai2026confidence} analyse, is a weaker bound on $\Ddep$ and hence a stricter commit criterion; it follows since $\max_i H_i \ge 0$. What Steps~2 and~3 add is to make that approximation explicit: substituting model entropies costs at most $\varepsilon$, which vanishes with the fidelity error at a rate set by the \emph{average} fidelity error per committed position, $\Dfid/|U|$.

\begin{remark}[Where the certificate is weakest]
\label{rem:cert_range}
The slack $|U|\,\omega(\Dfid/|U|)$ increases in both arguments, so it is largest where commits are large and the model is poor, a regime that often corresponds to small $T$. The per-position hypothesis $\KL(p^\star_i\Vert q_i)\le 2(1-1/m)^2$ is likewise most likely to fail there, since it must hold at every committed position. 
This is a limitation of the certificate rather than of the decomposition: \Eqref{eq:decomp} holds unconditionally, and so do Steps 1 and 2 above. A forced commit, the single position revealed when no candidate meets the budget (Appendix~\ref{app:decodingzoo_commit}), has $|U| = 1$ and therefore $\Ddep = 0$ identically, so it never violates the certificate.
\end{remark}

\subsection{Trajectory decomposition and the price of a finer schedule}
\label{app:proof_floor}

\begin{theorem}[Trajectory decomposition and schedule-dependent lower bound]
\label{thm:floor}
\begin{equation}
    \KL(P \Vert Q)
    \;=\; \Dfloor(\mathcal{U}) \;+\; \sum_{t} \E_{s_t \sim P}\big[\Dfid\big(U_t \mid s_t, q^{(\ell_t)}\big)\big]
    \;\geq\; \Dfloor(\mathcal{U}),
    \label{eq:traj}
\end{equation}
where $\Dfloor(\mathcal{U})=\sum_t\E_{s_t\sim P}[\Ddep(U_t\mid s_t)]$. For a fixed schedule $\mathcal{U}$, this term is determined by $p^\star$ and $\mathcal{U}$ and contains no learned marginals $q$. If the schedule is selected adaptively from model predictions, $q$ can affect it indirectly through $\mathcal{U}$.
\end{theorem}

\begin{proof}[Trajectory decomposition and the floor]
Both $P$ and $Q$ are Markov chains on the finite space of partial assignments, started at the same $s_0$, and at every state they fill the \emph{same} set of positions $U(s_t)$, differing only in the conditional law used to fill it. The map $(s_t, x_{U_t}) \mapsto s_{t+1}$ is injective for fixed $s_t$, so the divergence between the one-step transition kernels is the divergence between the laws of $x_{U_t}$. The chain rule for KL divergence between two Markov chains with a common initial state gives
\begin{equation}
\begin{aligned}
    \KL(P \Vert Q)
    &= \sum_{t} \E_{s_t \sim P}\Big[ \KL\big(P(\cdot \mid s_t) \,\big\Vert\, Q(\cdot \mid s_t)\big) \Big] \\
    &= \sum_{t} \E_{s_t \sim P}\Big[ \KL\Big(p^\star(x_{U_t}\mid s_t) \,\Big\Vert\, \prod_{i \in U_t} q^{(\ell_t)}_i \Big) \Big].
\end{aligned}
\end{equation}
This is the trajectory form of \citet{benhamu2025ebsampler}, whose reference process $q_\phi$ is our $P$; with $q_i^{(\ell_t)}=p^\star_i$ it reduces to the dependence-only decomposition of \citet{zhang2026generationorder}.
Adaptivity is not an obstacle to the identity: $U(\cdot)$ and $\ell(\cdot)$ are deterministic functions of the state, so conditionally on $s_t$ both chains commit at the same locations with the same loop count, and no policy-action distribution enters. Applying Lemma~\ref{lem:decomp} to each summand (legitimate by Remark~\ref{rem:ac_free}) splits it into a dependence and a fidelity part; collecting the former into $\Dfloor(\mathcal{U})$ gives \Eqref{eq:traj}. Dropping the non-negative fidelity terms gives the lower bound. The qualification in the theorem matters when comparing different loop budgets: a prediction-dependent commit policy may produce different schedules at different $L$, even if $|U_t|$ is fixed.
\end{proof}

\begin{lemma}[Splitting a commit]
\label{lem:split}
For any state $s$ and any partition $U = A \sqcup B$ into non-empty blocks,
\begin{equation}
    \Ddep(U \mid s)
    \;=\;
    \Ddep(A \mid s) \;+\; \E_{x_A}\big[\Ddep(B \mid s_A)\big] \;+\; \sum_{i \in B} I\big(x_i ; x_A \mid s\big),
    \label{eq:split}
\end{equation}
where $s_A$ is the state after committing $x_A \sim p^\star(\cdot \mid s)$ and all information quantities are taken under $p^\star(\cdot\mid s)$. Every term on the right is non-negative, so splitting a commit never increases the dependence cost.
\end{lemma}

\begin{proof}
Fix $s = (x_O, M)$ and $U = A \sqcup B$ with $A, B \neq \emptyset$. All entropies below are under $p^\star(\cdot\mid s) = \pidata(\cdot \mid x_O)$, and $s_A$ denotes the state after committing $x_A$, so that $p^\star(\cdot \mid s_A) = \pidata(\cdot \mid x_O, x_A)$. Averaging the definition of $\Ddep$ at $s_A$ over $x_A \sim p^\star(\cdot\mid s)$ turns the entropies into conditional entropies,
\begin{equation}
    \E_{x_A}\big[\Ddep(B \mid s_A)\big]
    = \sum_{i\in B} H(x_i \mid x_A) - H(x_B \mid x_A).
\end{equation}
Adding $\Ddep(A \mid s) = \sum_{i\in A} H(x_i) - H(x_A)$ and using the chain rule $H(x_A) + H(x_B \mid x_A) = H(x_U)$,
\begin{equation}
    \Ddep(A \mid s) + \E_{x_A}\big[\Ddep(B \mid s_A)\big]
    = \sum_{i\in A} H(x_i) + \sum_{i \in B} H(x_i \mid x_A) - H(x_U).
\end{equation}
Subtracting this from $\Ddep(U \mid s) = \sum_{i\in A} H(x_i) + \sum_{i\in B} H(x_i) - H(x_U)$ leaves
$\sum_{i\in B}\big[H(x_i) - H(x_i \mid x_A)\big] = \sum_{i \in B} I(x_i ; x_A)$, which is \Eqref{eq:split}. Non-negativity of the three terms is that of $\TC$ and of mutual information. \qedhere
\end{proof}

\begin{corollary}[Refining a schedule lowers the floor]
\label{cor:refine}
Let $\mathcal{U}$ be a commit schedule and let $\mathcal{U}'$ be obtained from it by replacing $U_t$ with the two consecutive commits $A$ and $B$, $A \sqcup B = U_t$, leaving the rest unchanged. Then
\begin{equation}
    \Dfloor(\mathcal{U}) - \Dfloor(\mathcal{U}')
    \;=\;
    \E_{s_t \sim P}\Big[\,\sum_{i \in B} I\big(x_i ; x_A \mid s_t\big)\Big] \;\ge\; 0 .
\end{equation}
Consequently $\Dfloor$ is non-increasing under refinement; it equals $\TC\big(\pidata(x_M \mid x_O)\big)$ for one-shot decoding ($T{=}1$, $U_1 = M$) and $0$ when every commit is a singleton.
\end{corollary}

\begin{proof}
Under $P$, drawing $x_U \sim p^\star(\cdot\mid s_t)$ and drawing $x_A \sim p^\star(\cdot \mid s_t)$ followed by $x_B \sim p^\star(\cdot \mid s_A)$ induce the same law on $x_U$, hence the same law on $s_{t+1}$. The steps before $t$ and, since the remaining rules are functions of the state, the steps after $t$ therefore contribute identically to $\Dfloor$ under $\mathcal{U}$ and $\mathcal{U}'$, and the difference is the step-$t$ difference, which is $\E_{s_t\sim P}[\sum_{i\in B} I(x_i;x_A \mid s_t)]$ by \Eqref{eq:split}. Non-negativity is that of mutual information. The endpoint values follow from $\TC$ evaluated at $U_1 = M$ and, for singletons, from $\TC = 0$ when $|U| = 1$. \qedhere
\end{proof}

\subsection{Scope of the results}
\label{app:proofs_scope}

\begin{remark}[What is proved and what is not]
\label{rem:premises}
Theorem~\ref{thm:floor} is an identity together with the bound obtained by discarding one non-negative term; it is not an optimality result. It does not prove that additional recursive loops monotonically reduce $\Dfid$ (that is an empirical question about the trained model, tested in Section~\ref{sec:analysis_empirical}, where $\Dfid$ falls up to $L{=}5$ but plateaus beyond it on multi-solution boards) and it does not identify an optimal allocation of a fixed $\NFE$ budget between $T$ and $L$, which would require modeling how $\Dfid$ varies with $\ell$. What it does establish is the conditional asymmetry the paper uses: at a fixed state and commit set, recursion cannot change the dependence term, while a refinement of the commit schedule lowers that term by the amount made explicit in \Eqref{eq:split}.
\end{remark}

\begin{remark}[KL versus task metrics]
\label{rem:metrics}
By Remark~\ref{rem:bijection}, $\KL(P \Vert Q)$ is the divergence between the sampler's output law and $\pidata(\cdot \mid x_O)$. The metrics we report (VPR, RTR, execution accuracy, structural gap $\Delta Z$) are task functionals and are not monotone in that divergence (in particular, forward KL can remain moderate while factorized sampling produces frequent jointly impossible samples~\citep{zhang2026generationorder}). A decoder can raise validity while losing diversity, and the entropy budget of Section~\ref{sec:decoding_rules} trades exactly along that axis. We therefore use task metrics to test the qualitative predictions of Section~\ref{sec:analysis_theory} rather than as estimators of either error term. The parallel-clash rate of Section~\ref{sec:analysis_empirical} is the closest task-level proxy for $\Ddep$, since it counts inconsistencies attributable specifically to simultaneous commitment. When $p^\star$ is spread and loops have reduced $\Dfid$, it is a useful proxy ($83$--$88\%$ of parallel clashes at $L{=}5$ are dependence errors), but it is not an estimator: fidelity error also produces parallel clashes, and produces all of them on unique boards, where $\Ddep\equiv0$.
\end{remark}

\section{Extended Related Work}
\label{app:related}

Section~\ref{sec:related} covers the threads most directly relevant to our contribution. This appendix broadens the context in three parts: the wider discrete-diffusion landscape (Appendix~\ref{app:discrete_diffusion}); decoding strategies for masked diffusion, including token ordering, adaptive denoising schedules, and remasking and iterative refinement (Appendix~\ref{app:decoding}); and looped or fixed-point computation (Appendix~\ref{app:recursive_decoding}).

\subsection{The Landscape of Discrete Diffusion for Language}
\label{app:discrete_diffusion}

The discrete diffusion family has grown rapidly. The D3PM framework~\citep{austin2021d3pm} provided the absorbing-state (masking), uniform, and nearest-neighbor transition variants inspired from continuous settings~\citep{sohl2015diffusion, ho2020ddpm, song2020score}, while SEDD~\citep{lou2024sedd} proposed score-based training for discrete spaces. Within this broader class, masked (absorbing-state) discrete diffusion (MDMs) have received particular attention. In contrast to \AR{} models, which generate tokens in a fixed left-to-right order, MDMs iteratively denoise a partially masked sequence using bidirectional context. This flexible generation order may benefit tasks requiring global or non-sequential reasoning \citep{ye2024beyond, he2026latent}, while the ability to update multiple tokens per step creates opportunities for faster inference. It also introduces an inference-time scaling axis absent from standard autoregressive decoding: increasing the number of denoising steps gives the model additional opportunities to resolve dependencies among tokens, at the cost of greater computation. In this work, we investigate recursive computation within each MDM denoising step as a complementary axis of inference-time scaling. 

Among masked approaches, MDLM~\citep{sahoo2024mdlm} demonstrated strong language modeling performance with a simplified objective, and LLaDA~\citep{nie2025llada} scaled masked diffusion to billions of parameters. Finally, more recent work including DiffusionGemma~\citep{gemini2026diffugemma} further demonstrates the practical appeal of block-parallel masked diffusion for high-throughput generation. Subsequent scaling law studies~\citep{sahoo2026scaling} challenge the assumption that masked diffusion is the uniquely best discrete noise type, finding that uniform diffusion may be more parameter-efficient at scale. Continuous--discrete hybrids have also been proposed \citep{cadd2025,onestepcont2026,coevo2025}, but tend to suffer from trainability issues in the continuous space.  
This supports our decision to remain in the masked discrete setting, where training is stable and the parallel denoising structure makes bidirectional recursion most natural.

\subsection{Decoding Strategies for Masked Diffusion}
\label{app:decoding}

A central characteristic of MDMs is that training does not prescribe a unique inference order. Consequently, a substantial literature studies how to choose which masked tokens to reveal and how many to reveal at each iteration.

\paragraph{Confidence- and entropy-based token ordering.}
MaskGIT~\citep{chang2022maskgit} is an early and influential example of iterative masked generation in which token confidence determines which positions are progressively revealed. More recent work makes this idea explicit for MDMs. \citet{kim2025tokenordering} theoretically and empirically study token ordering in masked diffusion and show that adaptive ordering can substantially improve performance on structured tasks. EB-Sampler~\citep{benhamu2025ebsampler} uses an entropy-bounded rule to dynamically determine how many tokens can be unmasked in a single model evaluation while controlling an approximate error tolerance. Confidence-based decoding has also received a formal efficiency analysis: \citet{cai2026confidence} analyze an entropy-sum unmasking strategy and provide iteration-complexity guarantees in terms of the entropy of the target distribution.

These methods are directly relevant to our use of entropy. Their primary role for entropy is to determine \emph{which tokens are sufficiently certain to commit} or \emph{how many tokens can safely be committed}. Our work extends this perspective to a second decision: whether the current state is ready for commitment or whether another recursive computation should first be performed.

\paragraph{Planning and alternative token orders.}
Several works frame decoding as an explicit planning problem. Path Planning~\citep{peng2025pathplanning} introduces a planner for choosing which tokens to unmask and shows that learned or model-based planning can improve over standard unmasking rules. Another approach separates planning from denoising and allows the planner to identify positions requiring further refinement~\citep{liu2024planned}. These approaches emphasize that the order of commitments can have a substantial effect on the final sample, complementary to our focus on whether additional computation should occur before a commitment.

\paragraph{Position and interaction-aware decoding.}
Recent work also points out limitations of purely token-wise confidence scores. UNCODE~\citep{huang2025pcsampler} studies position-dependent biases in uncertainty-based decoding and reports a tendency for trivial positions to be selected early. ADAS~\citep{sahin2026adas} explicitly accounts for interactions among candidate tokens by discounting candidates that attend strongly to uncertain positions already selected for decoding. This is closely related to our interest in inter-token dependencies, but ADAS addresses the problem through attention-aware reranking of the commit set, whereas we study whether recursive computation can resolve those dependencies before commitment. 

\paragraph{Inference Schedules and Information-Theoretic Decoding}
The number and placement of denoising steps constitute another major axis of MDM inference. \citet{li2025convergence} provide information-theoretic convergence guarantees in which sampling error depends on token mutual information. \citet{chen2026optimalmdm} give an exact characterization of sampling error for arbitrary unmasking schedules and derive schedules based on total correlation and dual total correlation. These results provide a formal perspective on the amount of parallelism that can be safely exploited when tokens are dependent.

Other work optimizes the discretization schedule directly by optimizing discrete diffusion timesteps using a bound on compounding decoding error~\citep{park2025jys}. \citet{zhang2025cosine} provide an information-geometric analysis showing that the cosine schedule is Fisher--Rao optimal for masked discrete diffusion. These approaches primarily optimize the \emph{outer diffusion schedule}; our setting introduces an additional inner computation axis and asks how compute should be allocated between another recursive pass and an additional denoising transition.

\paragraph{Remasking, Predictor--Corrector Methods, and Refinement}

A complementary family of approaches addresses the irreversibility of standard MDM decoding by allowing previously generated tokens to be revised. ReMDM~\citep{wang2025remasking} introduces inference-time remasking so that tokens can be unmasked again when later context indicates that an earlier commitment was unreliable. WINO~\citep{hong2025wino} similarly enables revokable decoding through parallel drafting, verification, and re-masking of suspicious tokens. These approaches provide a form of \emph{post-commitment correction}.

Our recursive mechanism is conceptually different: recursion operates \emph{before} the current set of tokens is committed, repeatedly refining the same partially masked state. Thus, remasking asks whether an existing commitment should be revoked, whereas our decoding policy asks whether commitment should be postponed in favor of additional computation. Predictor--corrector and planned-denoising approaches provide related perspectives on separating prediction, planning, and refinement~\citep{zheng2023rdm, liu2024planned, peng2025pathplanning}.

In other lines of work, \citet{chao2025partialmasking} explore partial masking strategies that interpolate between absorbing and uniform noise, while~\citet{genorder2026} learn generation orders via variational inference. Finally, \citet{horvitz2025med} also study inference-time compute allocation in MDMs and investigate entropy-based decoding choices alongside reasoning-oriented infilling. It is particularly relevant as an empirical study of the interaction between decoding strategy and reasoning performance, including structured reasoning tasks. Our focus is narrower: we study entropy and hidden-state change as signals for allocating compute between recursive refinement and denoising.

\paragraph{Efficient and Adaptive MDM Samplers}

A growing set of training-free samplers targets the speed--quality trade-off created by parallel MDM decoding. Fast-dLLM~\citep{wu2026fastdllm} combines KV-cache reuse with parallel decoding and identifies disrupted token dependencies as a key source of quality degradation. TABES~\citep{saini2026tabes} uses a backward entropy signal to estimate how revealing a candidate token will affect the future entropy of the remaining masked positions, thereby introducing a form of lookahead into token selection. These methods are closely related to our use of uncertainty and future effects, but operate primarily at the token-selection level. Fast solvers for discrete diffusion models~\citep{ren2025fastsolvers} instead reduce the number of numerical evaluations through higher-order integration schemes. These methods are complementary to our work: they accelerate the underlying diffusion trajectory or reduce the cost of individual evaluations, whereas our objective is to understand how to spend a fixed evaluation budget across recursive and denoising computation.

\subsection{Recursive and Fixed-Point Computation}
\label{app:recursive_decoding}

Weight sharing across depth has a long history in Transformer architectures, from Universal Transformers and ALBERT~\citep{dehghani2019universal, lan2020albert} to recent looped Transformers that trade parameter count for repeated computation~\citep{giannou2023looped, yang2024looped, saunshi2025looped, geiping2025recurrent, yu2025relay, xu2025formal}. Complementary analyses study what such recurrence actually buys: LT2 interprets looped attention as iterative memory refinement and receptive-field expansion~\citep{deng2026lt2}, DeepLoop examines the optimization and stability consequences of repeatedly revisiting shared parameters~\citep{li2026deeploop}, and~\citet{zhang2026recurrencealgorithm} ask whether weight-tied recurrence implements an iterative algorithm with identifiable convergence and propagation behavior. Closely related fixed-point reasoners adapt computation by monitoring hidden-state convergence on structured tasks~\citep{movahedi2026fprm}, while self-conditioned flow maps~\citep{yoo2026fixedpointflows} and Recursive Flow Matching~\citep{huang2026recfm} cast self-consistency as fixed-point iteration in continuous generative settings. Collectively, these works support treating recursion as a computational dimension distinct from parameter scaling. In \AR{} models, however, that extra computation is typically spent before emitting the next token; masked diffusion raises a qualitatively different allocation problem, because each unit of compute can either refine the current partially masked state or advance the outer denoising trajectory by committing tokens.

Within diffusion and masked generation, two concurrent works are closest to ours. LoopMDM selectively loops Transformer layers inside an MDM and shows that additional loops improve performance, including by allocating loops adaptively across denoising timesteps~\citep{lee2026loopmdm}; its analysis also reports increased interactions among masked positions. Fixed-Point Masked Generative Modeling instead embeds a fixed-point solver into masked generation and uses adaptive depth together with cross-step consistency~\citep{miele2026fpmgm}. Both establish that repeated computation is useful for masked generation. Our contribution differs in what is treated as the primary object of study. Rather than asking only whether more loops help, we ask how recursive refinement interacts with token commitment: loops update model marginals while the masked state is held fixed, whereas denoising steps change that state by revealing tokens. This distinction motivates the dependence--fidelity decomposition in Section~\ref{sec:analysis} and the two-axis decoder in Section~\ref{sec:decoding}, which jointly choose \emph{how many} recursive evaluations to run and \emph{which} tokens to commit. Architecturally, we also make different design choices: we loop the entire shared transformer block rather than selected layers, which enables deep supervision across loops, and we inject an explicit recursive step embedding so the shared block is conditioned on loop progress. In short, concurrent work shows that recursion is a viable compute axis inside MDMs; we characterize when that axis is interchangeable with denoising, when it is not, and how to allocate both jointly at inference time.

\section{Dataset and Metrics}
\label{app:datasets}

We evaluate our models on a diverse set of datasets spanning structured combinatorial reasoning (Sudoku), symbolic arithmetic planning (Countdown), biological sequence modeling (RNA), and natural-language arithmetic reasoning (TinyGSM). This selection allows us to test different aspects of sequence modeling: constraint satisfaction, multi-step symbolic reasoning with intermediate state tracking, biological sequence generation, and programmatic mathematical reasoning. For each dataset, we construct train and validation splits and report both training objectives and task-specific evaluation metrics. Further information on split sizes, vocabulary sizes, and sequence lengths are detailed in Table~\ref{tab:datasetsummary}.

For all datasets during training, we monitor the average masked cross-entropy loss over mini-batches, computed for both the train and validation splits by averaging over \texttt{eval\_iters} randomly sampled batches at the end of each evaluation interval.  For a batch $\{(\xt^{(i)}, \xo^{(i)}, m^{(i)})\}_{i=1}^{B}$ of size $B$, the estimated validation loss is
\begin{equation}
  \hat{\mathcal{L}}_{\mathrm{val}} = \frac{1}{B}
    \sum_{i=1}^{B} \mathcal{L}(y^{(i)};\, \xo^{(i)},\, m^{(i)}),
\end{equation}
where $\mathcal{L}$ is whichever training objective is active (depending on the choice of supervision, see Appendix~\ref{app:loss_mode_ablation}).

\begin{table}[h]
\centering
\caption{Summary statistics for all datasets used in our experiments. Sequence length refers to the fixed context length used during training. Vocabulary size includes special tokens such as masks where applicable.}
\label{tab:datasetsummary}
\begin{tabular}{lcccc}
\hline
Dataset & Train Size & Val Size & Vocab Size & Seq Length \\
\hline
Sudoku $9\times 9$ & 1.8M & 100k & 10 & 81 \\
Sudoku $25\times 25$ & 100k & 20k & 26 & 625 \\
Countdown ($k=3$) & 100k & 20k & 18 & 48 \\
Countdown ($k=4,5$) & 100k & 20k & 18 & 64 \\
RNA & 38.8M & 125k & 25 & 256 \\
TinyGSM & 11.7M & 118k & 49,154 & 512 \\
\hline
\end{tabular}
\end{table}

\subsection{Sudoku}
 
\paragraph{Sudoku $9 \times 9$} For the standard Sudoku, we adapt the dataset from~\citet{shah2024causallanguagemodelingelicit}, which consists of pre-existing puzzle files containing solved $9 \times 9$ puzzles encoded as move sequences. Each sample in the raw data is a sequence of 81 moves, where each move records a quadruple $(r, c, v, s)$ that indicates the row index, column index, digit value, and a strategy tag respectively. Since we are interested in training on the raw Sudokus, we preprocess the data and store the boards row-wise as flattened sequences of length 81 with digits in $\{1, \ldots, 9\}$.
The vocabulary consists of digits $\{0, 1, \ldots, 9\}$, where 0 serves as the mask token. 

\paragraph{Extended sudoku} For experiments at larger scales, we construct a synthetic dataset of $n \times n$ Sudoku boards, where $n$ is chosen such that a valid rectangular block decomposition exists. Specifically, we require $n = b_r \times b_c$ for integers $b_r, b_c \geq 2$ chosen as the factor pair closest to $\sqrt{n}$. Each board is a valid Sudoku with digits in $\{1, \ldots, n\}$, stored as a flattened sequence of length $n^2$.

Boards are generated by first constructing a single deterministic base solution using the closed-form pattern
\begin{equation*}
    G_{r,c} = \left(b_c \cdot (r \bmod b_r) + \lfloor r / b_r \rfloor + c\right) \bmod n + 1,
\end{equation*}
which yields a valid completed $n \times n$ grid for any compatible block shape. Each subsequent sample is derived from this base by applying a random symmetry-preserving permutation: row-bands and rows within each band are independently shuffled, column-stacks and columns within each stack are independently shuffled, and digit labels are remapped via a random permutation of $\{1, \ldots, n\}$. All three operations preserve Sudoku validity. The vocabulary consists of digits $\{0, \ldots, n\}$, with 0 again serving as the mask token, giving a vocabulary size of $n+1$.

We generate 100,000 boards for training and 20,000 for validation with $n=25$. A fixed set of binary blank masks is pre-generated for the validation split, where each mask is drawn i.i.d. with a per-cell blank probability (0.7, 0.8 or 0.9), subject to the constraint that at least one cell is blank and at least one is given.

\paragraph{Evaluation protocol}
Sudoku evaluation is conditional infilling on held-out boards: the given cells (clues) remain visible, while blank cells are replaced by the mask token and filled by the decoder. For $9\times9$, blanks follow the dataset's strategy annotations (non-clue cells); for $25\times25$, validation blanks are sampled independently.

\paragraph{Metrics and evaluation}

Here $n$ denotes the grid side length and the board has $n^2$ cells; $b_r$ and $b_c$ denote the row and column dimensions of each subgrid block, with $n=b_r b_c$.

\begin{itemize}
    \item \textbf{Valid Puzzle Rate (VPR)}: A generated $n \times n$ board $y \in \{1,\ldots,n\}^{n^2}$ is valid if and only if every row, every column, and every $b_r \times b_c$ block contains each digit in $\{1,\ldots,n\}$ exactly once.  
Any out-of-range digit, repeated digit in a unit, or board of incorrect length counts as invalid.
\item \textbf{Soft Constraint Loss (SCL)}: This metric is adapted from~\citet{he2026latent} and allows a more fine-grained evaluation of the quality of the generated sudokus. For a board $y \in \mathbb{Z}^{n^2}$ with vocabulary $\mathcal{V} = \{1,\ldots,n\}$, it measures the fractional coverage of the required digit set for each constraint unit:
\begin{equation}
  \ell(u;\, y) = 1 - \frac{|\,\{y_j : j \in u\} \cap \mathcal{V}\,|}{n},
  \label{eq:unit_loss}
\end{equation}
where $u$ ranges over all $3n$ units (rows, columns, blocks).  A perfectly valid board has $\ell(u;\,y) = 0$ for all $u$; a board where every unit contains only one distinct valid digit has $\ell(u;\,y) = (n-1)/n$ for all $u$.  
The maximum value $3(n-1)$ corresponds to a board where every unit contains a single repeated in-vocabulary digit. 
\noindent The board-level metric is $\mathrm{SCL}(y)=\sum_{u}\ell(u;y)$ over the $3n$ rows, columns, and blocks. Mask and out-of-vocabulary tokens are absent from the covered set, and therefore $0\leq \mathrm{SCL}(y)\leq 3(n-1)$.
\end{itemize}

\subsection{Countdown}

\paragraph{Dataset} The Countdown dataset~\citep{yao2023tree, gandhi2024streamsearchsoslearning} is a synthetic arithmetic reasoning task designed to evaluate a model's ability to perform multi-step planning, maintain a dynamic state (the ``pool''), and execute precise arithmetic operations. The task requires generating a sequence of operations that transform a set of initial numbers into a specific target value.

Each sample is generated procedurally to ensure the existence of at least one valid solution. The generation follows the logic:
\begin{enumerate}
    \item \textbf{Initialization:} A set of $k$ initial operands $\mathcal{N} = \{n_1, n_2, \dots, n_k\}$ is sampled uniformly from the range $[1, 100]$. In our experiments, $k=3, 4, 5$.
    \item \textbf{Chain Construction:} The algorithm iteratively reduces the pool of numbers until a single value remains. In each step, two numbers $a$ and $b$ are sampled from the current pool and combined using an operator $\odot \in \{+, -, *, /\}$. 
    \item \textbf{Validation:} Operations must result in integers. For division, it is required that $b \neq 0$ and $a \equiv 0 \pmod b$. 
    \item \textbf{Target Assignment:} The final remaining number in the pool after $k-1$ steps is designated as the target $\tau$.
\end{enumerate}

Each example is represented as a fixed-length character sequence of length $N_{seq}=48$ for $k=3$; and $N_{seq}=64$ for $k=4, 5$. The dataset uses a restricted vocabulary of 18 tokens: digits $0$--$9$, operators $+$, $-$, $*$, $/$, and delimiters $=$, $,$, and $\backslash n$. The mask token is \_.

\paragraph{Evaluation protocol}
Countdown is prefix-conditioned infilling: the operand list and target stay visible, and only the arithmetic completion region is masked.

\paragraph{Metrics and evaluation}

\begin{itemize}
    \item \textbf{Reaches Target Rate (RTR):}
We consider a solution \emph{valid} if and only if: (i) each step is arithmetically correct ($a \odot b = c$ evaluates correctly); (ii) at each step both operands $a$ and $b$ are available in the current pool, which is initialized to $\mathcal{N}$ and updated by removing $a$ and $b$ and inserting $c$ after each step; and (iii) after all steps the pool contains exactly one element equal to $\tau$. The Reaches Target Rate is then:
\begin{equation}
  \mathrm{RTR} = \frac{1}{N} \sum_{i=1}^{N}
    \mathbf{1}\!\left[y^{(i)} \text{ is a valid Countdown solution}\right].
  \label{eq:rtr}
\end{equation}
\item \textbf{Local Arithmetic Fraction (LAF):}
LAF measures the fraction of generated steps whose arithmetic is
locally correct, regardless of whether the operands were available in
the pool at that point in the solution:
\begin{equation}
  \mathrm{LAF}(y) = \frac{1}{|S|}
    \sum_{s \in S}
    \mathbf{1}\!\left[
      \text{parse}(s) \neq \varnothing \;\text{ and }\;
      a \odot b = c \text{ is numerically correct}
    \right],
  \label{eq:laf}
\end{equation}
where $S$ is the set of step strings in $y$ that match the expected
format $a \odot b = c$.  LAF distinguishes between arithmetic errors
(low LAF, low RTR) and correct arithmetic with pool-management errors
(high LAF, low RTR).  
\item \textbf{Pool Prefix Fraction (PPF):}
PPF measures how far through a \emph{pool-consistent} simulation the
model gets before its first failure, normalized by the expected number
of steps $k - 1$ where $k = |\mathcal{N}|$:
\begin{equation}
  \mathrm{PPF}(y) = \frac{
    \max\bigl\{j : \text{steps } s_1,\ldots,s_j \text{ are all valid and pool-consistent}\bigr\}
  }{k - 1}
  \;\in\; [0, 1].
  \label{eq:ppf}
\end{equation}
A model that always fails on the first step has PPF $= 0$; RTR $= 1$
implies PPF $= 1$.  PPF tracks progress toward the solution and is particularly informative at intermediate loop counts. 
\item \textbf{Target Residual Norm (TRN):}
After following the longest valid pool-consistent prefix of the generated solution, TRN measures how close the remaining pool values are to the target $\tau$:
\begin{equation}
  \mathrm{TRN}(y) = \frac{
    \min_{x \in \mathcal{P}^*}\, |x - \tau|
  }{
    \max(1,\, |\tau|)
  }
  \;\in\; [0,\, +\infty),
  \label{eq:tr}
\end{equation}
where $\mathcal{P}^*$ is the pool state after the longest pool-consistent prefix.  TRN $= 0$ means the correct answer appears somewhere in the pool (a necessary condition for RTR $= 1$); TRN $> 0$ quantifies how far the best available value is from the target, scaled by the magnitude of the target to be comparable across instances.  An unparseable or entirely invalid chain gives TRN $= 1.0$ by convention.
\end{itemize}

\subsection{RNA}
\label{app:rna_datametrics}

\paragraph{Dataset} The dataset corpus is built from RNAcentral~\citep{rnacentral2026} (39,048,508 sequences). Removing exact duplicates leaves 39,035,267 sequences,
each carrying the RNA type assigned by RNAcentral. We retain the nineteen most
frequent types as families and collapse the remainder into a single \emph{other}
family, giving 20 families in total. Held-out sets are drawn per family with a fixed quota rather than proportionally:
10,000 sequences for each of rRNA, lncRNA, tRNA, misc\_RNA and \emph{other}, and
5,000 for the remaining fifteen. This yields 125,000 sequences per held-out split
and leaves 38,785,267 for training. The quota makes the held-out sets
quasi-uniform over families, so validation loss is not dominated by ribosomal RNA, which accounts for 72.5\% of the training pool.

The RNA dataset therefore consists of non-coding RNA sequences stored as per-family HuggingFace Arrow datasets, with separate train and validation directories. The training split contains 38,785,267 sequences and the validation split contains 125,000 sequences, distributed across 20 RNA families. During training, family streams are interleaved with temperature-scaled sampling probabilities \( p_f \propto n_f^{\tau}\), where $n_f$ is the number of sequences in family $f$ and we use $\tau=0.7$. This damped sampling rule interpolates between uniform family sampling ($\tau=0$) and size-proportional sampling ($\tau=1$).

Sequences are uppercased and tokenized with a RiNALMo-style~\citep{penic2025rinalmo} RNA vocabulary of size 25: the special tokens \texttt{<pad>}, \texttt{<unk>}, \texttt{<mask>}, \texttt{<cls>}, \texttt{<eos>}, and the 20 sequence symbols \texttt{A C G U N R Y S W K M B D H V . X * - I}.
Unless otherwise stated, each sequence is wrapped with \texttt{<cls>} and \texttt{<eos>}, truncated if necessary, and padded to a fixed length of 256 tokens. Masking is applied only to non-special tokens. For each sequence, the masking probability is sampled uniformly from $[0,1]$, selected positions are replaced by the \texttt{<mask>} token.

\paragraph{Metrics and evaluation}
All metrics compare an infilled sequence $s^{\mathrm{inf}}$ to its held-out native counterpart
$s^{\mathrm{nat}}$. Both sequences have the same length and agree on every unmasked position. We write
$\mathcal{M}$ for the set of masked positions.

\begin{itemize}
    \item \textbf{Structural Z-score:}
    Secondary-structure stability is evaluated with ViennaRNA folding. Let $\mathrm{MFE}(s)$ be the predicted
    minimum free energy of $s$ in kcal/mol. To control for nucleotide composition, we compare it to $20$
    mononucleotide-shuffled copies of $s$:
    \begin{equation}
      Z(s) =
      \frac{
        \mathrm{MFE}(s) - \mu_{\mathrm{shuffle}}(s)
      }{
        \sigma_{\mathrm{shuffle}}(s)
      }.
    \end{equation}
    More negative values mean the sequence folds to a lower energy than its composition alone would predict.

    \item \textbf{Paired Structural Gap ($\Delta Z$):}
    The primary RNA infilling metric:
    \begin{equation}
      \Delta Z(s) \;=\; Z(s^{\mathrm{inf}}) \;-\; Z(s^{\mathrm{nat}}) ,
    \end{equation}
    averaged over the evaluation set. $Z$ is already standardized against length- and
    composition-matched shuffles, and the difference is taken against each sequence's \emph{own}
    native fold. As a result, $\Delta Z$ is comparable across lengths and can be averaged over length strata.
    Lower is better. $\Delta Z = 0$ means the completion is as structured as the native molecule, and
    $\Delta Z > 0$ means it folds less tightly than the original.

    \item \textbf{Masked Recovery:}
    The fraction of masked positions whose nucleotide matches the native sequence,
    \begin{equation}
      \mathrm{Rec}(s) = \frac{1}{|\mathcal{M}|} \sum_{t \in \mathcal{M}} \mathbf{1}\!\left[s^{\mathrm{inf}}_t = s^{\mathrm{nat}}_t\right].
    \end{equation}
    Uniform guessing over $\{\mathrm{A},\mathrm{C},\mathrm{G},\mathrm{U}\}$ gives $25\%$. We report this
    metric as a diagnostic, not as a target: values well above chance show that the completion is consistent
    with its context, while values close to $100\%$ would mean the model copies the native molecule
    instead of designing a new one.

    \item \textbf{Exact Recovery:}
    The fraction of evaluated sequences in which \emph{every} masked position is restored to the native
    nucleotide, i.e.\ $\mathbf{1}[\mathrm{Rec}(s) = 1]$ averaged over the evaluation set. Lower is better:
    a low exact-recovery rate shows that good $\Delta Z$ does not come from reproducing the validation set.
\end{itemize}

\paragraph{Infilling evaluation protocol}
The RNA results of Section~\ref{sec:decoding_results} are conditional infilling on the validation split,
not unconditional design. For each evaluated sequence of length $n$, exactly $\mathrm{round}(\rho n)$
nucleotide positions are masked uniformly at random and decoded. The remaining positions, together with
the special and padding tokens, stay fixed, so the model restores the blanks in the context of the native
molecule. The main text uses a mask fraction of $\rho = 40\%$, and Appendix~\ref{app:rna_full} also
reports $\rho \in \{50\%, 60\%\}$. Sampling is restricted to the main nucleotides
$\{\mathrm{A},\mathrm{C},\mathrm{G},\mathrm{U}\}$, under the decoders and hyperparameters of
Appendix~\ref{app:decodingzoo}. We stratify by length and draw $500$ held-out sequences at each of
$n \in \{76, 110, 254\}$ nucleotides. These are decoded at $T = 6$, $9$ and $20$ respectively, so that
at $\rho = 40\%$ every stratum commits roughly five positions per denoising step. The sequences and
masked positions are fixed by a seed and shared across all decoders and models. We fold both the native
and the infilled sequence with ViennaRNA to obtain $\Delta Z$. Because $\Delta Z$ is length-normalized
by construction, the three strata are averaged into the single RNA panel of
Figure~\ref{fig:empirical_results}. Per-decoder values for all mask fractions are in
Table~\ref{tab:rna-extended}.

\subsection{TinyGSM}
\label{app:tinygsm_datametrics}

\paragraph{Dataset} TinyGSM~\citep{liu2023tinygsm} is a large synthetic grade-school mathematics dataset consisting of natural-language questions paired with executable Python solutions. We use the HuggingFace \texttt{TinyGSM/TinyGSM} training split. After filtering examples that exceed the model context length, the resulting split contains 11,703,733 training examples and 118,220 validation examples.

Each example is tokenized with the \texttt{HuggingFaceTB/SmolLM-135M} tokenizer and represented as
\[
  \texttt{BOS} \;+\; \texttt{question} \;+\; \texttt{separator} \;+\; \texttt{Python solution} \;+\; \texttt{EOS},
\]
then padded to a fixed sequence length of 512. Since the tokenizer does not provide a dedicated mask token, we append an explicit mask token, yielding a vocabulary size of 49,154. Training follows an answer-only masked infilling setup: prompt tokens corresponding to the question are kept fixed, while the solution region is eligible for masking and supervision.

\paragraph{Metrics and evaluation}

\begin{itemize}
    \item \textbf{GSM8K Accuracy:}
    We evaluate TinyGSM models on the \texttt{openai/gsm8k} test split. For each problem, the tokenized question and separator are kept fixed and all post-prefix positions are masked, up to the 512-token canvas. The model generates a Python completion, from which we extract and execute a function named \texttt{simple\_math\_problem} in a restricted sandbox. A sample is correct if the returned value, or a numeric answer extracted from its output, matches the GSM8K gold answer:
    \begin{equation}
      \mathrm{GSM8KAcc}
      =
      \frac{1}{N}
      \sum_{i=1}^{N}
      \mathbf{1}\!\left[
        \mathrm{exec}(y^{(i)}) = a^{(i)}
      \right].
    \end{equation}
    Execution failures, missing functions, timeouts, unparseable outputs, or numerically incorrect answers are counted as incorrect.
\end{itemize}

\section{Training Algorithm and Design Choices}
\label{app:algorithms}

\subsection{Training procedure}
Training uses a masked denoising objective~\citep{sahoo2024mdlm}. We randomly mask positions within the completion region, and the model is trained to minimize the cross-entropy loss at those positions. Further details are described in Algorithm~\ref{alg:training}.

\begin{algorithm}[htb]
  \caption{Training Iteration with Recursive Refinement}
  \label{alg:training}
  \begin{algorithmic}[1]
    \STATE {\bfseries Input:} Dataset $\mathcal{D}$, Model $f_\theta$, Optimizer $\mathcal{O}$, Accumulation Steps $A$
    \FOR{iteration $it=1$ {\bfseries to} $I_{max}$}
        \STATE $\mathcal{O}.zero\_grad()$
        \STATE $\mathcal{L}_{accum} = 0$
        \FOR{micro-step $a=1$ {\bfseries to} $A$}
            \STATE Sample batch $(\mathbf{x}, \mathbf{y}, \mathbf{mask})$ from $\mathcal{D}$
            \STATE Resolve training loop count $L_t$ (Fixed, Sample, or Schedule)
            \STATE $\text{logits}, \mathcal{L}, \{\mathcal{L}_k\} \gets f_\theta(\mathbf{x}, \mathbf{y}, \mathbf{mask}, L_t, \text{step\_idx}=it)$
            \STATE Calculate $\mathcal{L}_{scaled} = \mathcal{L} / A$
            \STATE Compute gradients: $\mathcal{L}_{scaled}.backward()$
            \STATE $\mathcal{L}_{accum} \gets \mathcal{L}_{accum} + \mathcal{L}.item()$
        \ENDFOR
        \STATE $\mathcal{O}.step()$
        \STATE Update Learning Rate Scheduler
    \ENDFOR
  \end{algorithmic}
\end{algorithm}

Models for Sudoku and Countdown were trained on a single NVIDIA A100-SXM4-80GB GPU. The model for RNA was trained on a single NVIDIA H200-SXM5-141GB GPU, and the model for TinyGSM on 8 NVIDIA H200-SXM5-141GB GPUs.

Across all runs, the base optimizer settings are shared: AdamW with learning rate $\mathrm{LR}=3\times 10^{-4}$ and a cosine scheduler with warmup (\texttt{warmup\_steps}=2000, \texttt{min\_lr\_ratio}=0.1). Further details on the models used are detailed in Table~\ref{tab:appendix-tech-details}.

\begin{table*}[ht]
\centering
\small
\caption{Architecture and optimization settings, sequence length, masking, batch size, and effective training epochs for each dataset. Parameter counts include the step-embedding MLP ($\approx d^2$, i.e., $0.15$M for $d{=}384$) that is not a part of the non-recursive baselines.}
    \resizebox{0.97\linewidth}{!}{
\begin{tabular}{lcccccccc}
\hline
\textbf{Dataset} & $n_{\text{layer}}$ & $n_{\text{head}}$ & $n_{\text{embd}}$ & Params & Seq.\ len & Masking & Batch size & Epochs \\
\hline
\textbf{Sudoku $9 \times 9$} & 6 & 6 & 384 & 10.8M & 81 & Uniform & 64 & 10 \\
\textbf{Sudoku $25 \times 25$} & 6 & 6 & 384 & 10.8M & 625 & Uniform & 32 & 75 \\
\textbf{Countdown} & 3 & 12 & 384 & 5.5M & 48/64 & Uniform & 512 & 300 \\
\textbf{RNA} & 12 & 20 & 480 & 33.7M & 256 & Uniform MLM & 128 & 1 \\
\textbf{TinyGSM} & 12 & 12 & 768 & 161M & 512 & Log-linear MDLM & 64 & 1.4 \\
\hline
\end{tabular}}
\label{tab:appendix-tech-details}
\end{table*}

\paragraph{Positional encodings} The model uses two positional-encoding regimes. For non-Sudoku datasets, we employ standard 1D rotary positional embeddings (RoPE, \citep{su2023rope}). For Sudoku datasets, we explore a 2D variant: each flattened token index is mapped back to grid coordinates $(r,c)$, rotary frequencies are computed separately for row and column components, and then concatenated so the attention mechanism is aware of both horizontal and vertical structure. In addition, we add a learned \emph{block embedding} identifying each subgrid (e.g., $3\times3$ blocks in $9\times9$ Sudoku), which explicitly encodes Sudoku block constraints beyond pure sequence order.

\paragraph{Masking strategies}
All datasets use a masked denoising objective~\citep{sahoo2024mdlm}: a subset of positions is replaced by a mask token and the model is trained with cross-entropy on those positions.
The datasets differ in \emph{which} positions are eligible and in \emph{how} the per-sequence mask rate is sampled.
For Sudoku, every grid cell is eligible; each sequence draws a mask probability $p\sim\mathrm{Unif}[0,1)$ and masks cells independently with probability $p$.
Countdown uses the same uniform rate, but only within the completion region after the fixed prompt prefix (padding newlines are never masked).
TinyGSM likewise freezes the question/prompt tokens and applies masking only to the answer span (including padding under our default settings), but samples the mask rate from the MDLM \emph{log-linear} noise schedule rather than a uniform draw:
$t\sim\mathrm{Unif}(\varepsilon_{\mathrm{samp}},1)$ with $\varepsilon_{\mathrm{samp}}=10^{-3}$, total noise $\sigma(t)=-\log\bigl(1-(1-\varepsilon)t\bigr)$ with $\varepsilon=10^{-3}$, and move chance $1-e^{-\sigma(t)}$.
RNA is closest to classical MLM: special tokens (\texttt{<pad>}, \texttt{<unk>}, \texttt{<mask>}, and optionally \texttt{<cls>}/\texttt{<eos>}) are excluded, and remaining nucleotide positions are masked with a uniform per-sequence rate $p\sim\mathrm{Unif}[0,1)$.

\paragraph{Further RNA training details}
RNA uses the RiNALMo-micro encoder architecture~\citep{penic2025rinalmo} 
rather than the GPT-style backbone used on the other datasets. We adopt only the architecture: the model is trained from scratch.
Sequences are fixed to length 256 (including optional \texttt{<cls>}/\texttt{<eos>}), with truncation/padding as needed. The recursive RNA model was trained with Schedule $5 \to 1$ loop count.

\paragraph{Further TinyGSM training details}
TinyGSM uses a larger GPT-style transformer 
over length-512 sequences tokenized with the SmolLM tokenizer.
Examples are question--code pairs from TinyGSM; sequences longer than 512 tokens are filtered out.
Training freezes the prompt and applies log-linear MDLM masking on the answer region. 
TinyGSM additionally uses bf16 precision, global gradient clipping at norm $1.0$, and EMA of the weights with decay $0.9999$. The recursive TinyGSM model model was trained with Schedule $5 \to 1$ loop count.

\subsection{Choosing the training loop count $L_t$}
\label{app:rec_steps_ablation}

We study several strategies for selecting $L_t$, the number of recursive loops run per forward pass during training. This is distinct from the loop budget $L_s$ used at sampling time and from the denoising budget $T$. These strategies control the amount of iterative refinement applied by the model and can be broadly categorized into \textit{fixed}, \textit{stochastic}, and \textit{scheduled} schemes.

\paragraph{Fixed-depth recursion}

In the fixed-depth setting, the number of recursive iterations is held constant, $L_t = L$, where $L$ is a predefined hyperparameter. This approach provides stable optimization and predictable computational cost, and is commonly used in looped or recurrent transformer architectures. Notably, \citet{yang2024looped} demonstrate that Transformers trained with a fixed number of recurrent loops can improve algorithmic generalization when unrolled for multiple iterations. Similarly, RecursiveVLM \citep{xu2026loopingforwardrecursivetransformers} trains with a fixed recursion depth and evaluates the model at different sampling recursive steps.

\paragraph{Random-depth sampling}

In the stochastic depth setting, the recursion depth is sampled per example or batch, $L_t \sim \Lambda$, where $\Lambda$ denotes a predefined probability distribution over positive integers. This formulation exposes the model to a distribution over computational budgets during training, improving robustness to variable inference-time compute. This follows the line of work from \citet{geiping2025recurrent}, which samples recurrent iteration counts using log-normal or Poisson-like distributions combined with truncated backpropagation through time. This enables the model to allocate computation adaptively while maintaining efficiency on average. In our implementation, this category includes:
\begin{itemize}
    \item Uniform distribution: $L_t \sim \mathcal{U}(L_{\min}, L_{\max})$
    \item Poisson distribution: $L_t \sim \text{Poisson}(\lambda) + 1$, where $\lambda$ is drawn from a log-normal prior with parameters chosen to match a desired mean depth
\end{itemize}

\paragraph{Curriculum-based scheduling}

In curriculum-based approaches, the recursion depth is deterministically varied over the course of training, $L_t = f(i)$ where $i$ denotes the training iteration and $f(\cdot)$ is a monotonic schedule mapping training progress to computation depth. This idea has been widely used in recurrent and looped architectures~\citep{dehghani2019universal, fan2024looped}. Typically, $L_t$ is gradually increased or decreased from an initial value $L_{\text{start}}$ to a final value $L_{\text{end}}$, enabling a curriculum over looping depth:
\[
L_t = \mathrm{round}\left(L_{\text{start}} + \alpha(i)(L_{\text{end}} - L_{\text{start}})\right),
\]
with $\alpha(i) \in [0,1]$.

In our implementation, we evaluate both \textbf{increasing} and \textbf{decreasing linear schedules}, corresponding respectively to $L_{\text{start}} < L_{\text{end}}$ and $L_{\text{start}} > L_{\text{end}}$. This allows us to study whether progressively allocating more or less recursive steps affects downstream performance.

\begin{figure*}[ht]
    \centering
    \includegraphics[width=\textwidth]{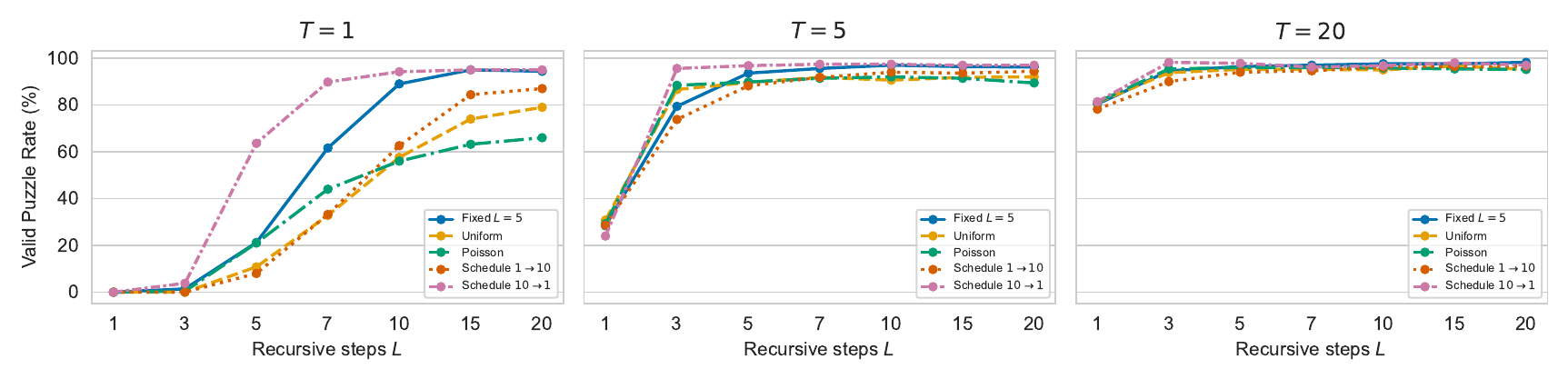}
    \caption{Performance comparison between models trained with different recursive steps choices on $9 \times 9$ Sudoku puzzles. We report the Valid Puzzle Rate across varying loop counts at sampling $L_s$ and denoising steps $T$.}
    \label{fig:recursive_scheme_ablation}
    \vspace{-0.4cm}
\end{figure*}

Figure~\ref{fig:recursive_scheme_ablation} evaluates the impact of different training schemes for the number of recursive steps on the final performance across evaluation depths $L_s \in \{1,3,5,7,10,15,20\}$ and decoding budgets $T \in \{1,5,20\}$. We compare a constant baseline (\emph{Fixed $L{=}5$}), stochastic sampling methods (\emph{Uniform} and \emph{Poisson}), a standard linear curriculum that increases recursive steps during training (\emph{Schedule $1 \to 10$}), and a reverse linear curriculum (\emph{Schedule $10 \to 1$}). The reverse curriculum (\emph{Schedule $10 \to 1$}) emerges as the superior training strategy by a substantial margin, particularly at small-to-moderate evaluation depths and under tight decoding budgets.

For instance, at $T{=}1$ and $L_s{=}5$, \emph{Schedule $10 \to 1$} achieves a $63.6\%$ valid puzzle rate, outperforming the constant baseline \emph{Fixed $L{=}5$} ($21.2\%$), stochastic \emph{Poisson} ($21.0\%$), and the forward curriculum \emph{Schedule $1 \to 10$} ($8.0\%$).
This performance gap is even more pronounced at $L_s{=}7$ under $T{=}1$, where the reverse curriculum reaches $89.8\%$ compared to just $61.6\%$ for \emph{Fixed $L{=}5$} and $33.2\%$ for \emph{Schedule $1 \to 10$}. Even when more decoding steps are allowed, the reverse curriculum retains its advantage: at $T{=}5$ and $L_s{=}3$, it reaches $95.6\%$ validity, whereas \emph{Fixed $L{=}5$} and \emph{Schedule $1 \to 10$} lag behind at $79.4\%$ and $73.8\%$, respectively.
These results suggest that introducing deep recursion early in training forces the model to learn high-capacity, multi-step coordination, while progressively reducing the depth down to $1$ fine-tunes its ability to execute highly efficient, single-step updates.
Conversely, starting with shallow steps (\emph{Schedule $1 \to 10$}) or sampling them randomly (\emph{Uniform}, \emph{Poisson}) fails to establish the same level of systematic multi-step alignment, leading to sub-optimal performance across all evaluation depths.

\subsection{Step embedding ablation}
\label{app:embed_ablation}

We investigate the role of the recursive step embedding introduced in Section~\ref{sec:rmdm}. In its basic form, Equation~\ref{eq:hloop} applies $f_\theta$ repeatedly with shared parameters, where each loop operates on the evolving hidden state but otherwise receives the same conditioning inputs. The model has no explicit signal indicating which loop is currently being executed; it must infer this from the evolving statistics of $\mathbf{h}^{(\ell)}$ alone.  To give the model direct access to its position in the recursive computation, we introduce a \emph{step embedding}.
 
Let $s_\ell = (\ell - 1) / (L - 1) \in [0, 1]$ be the normalized loop progress (with $s_1 = 0$ for $L = 1$). At inference, $L$ is the sampling loop budget $L_s$, which under adaptive stopping is the upper bound on the number of loops.  We map this scalar to a $d$-dimensional vector via a two-layer MLP:
\begin{equation}
  \mathbf{v}^{(\ell)} = W_2\,\sigma\!\left(W_1\, s_\ell + b_1\right),
  \qquad
  W_1 \in \R^{d \times 1},\;
  W_2 \in \R^{d \times d},
  \label{eq:step_embed}
\end{equation}
where $\sigma$ denotes the SiLU activation.  The step vector $\mathbf{v}^{(\ell)}$ is broadcast over the sequence and added to the hidden state before each loop application.
The step embedding parameters $(W_1, b_1, W_2)$ are \emph{shared} across loops (they depend on $\ell$ only through $s_\ell$) and add $\mathcal{O}(d^2)$ parameters to the model, which is a negligible overhead relative to the transformer block.

The step embedding allows the model to specialize computation by loop index while using normalized progress $s_\ell$ to remain well-conditioned across different loop counts. To evaluate its impact, we compare three variants: (i) a model with a \emph{learned} step embedding, where the normalized recursion progress is projected into the hidden space through a small trainable network; (ii) a model with a \emph{fixed} step embedding, constructed directly from the normalized recursion progress (ranging from 0 to 1) using a deterministic encoding; and (iii) a model with \emph{no step embedding}, in which all recursive iterations share exactly the same computation and receive no explicit indication of the current recursion depth. 

\begin{figure*}[h!]
    \centering
    \includegraphics[width=\textwidth]{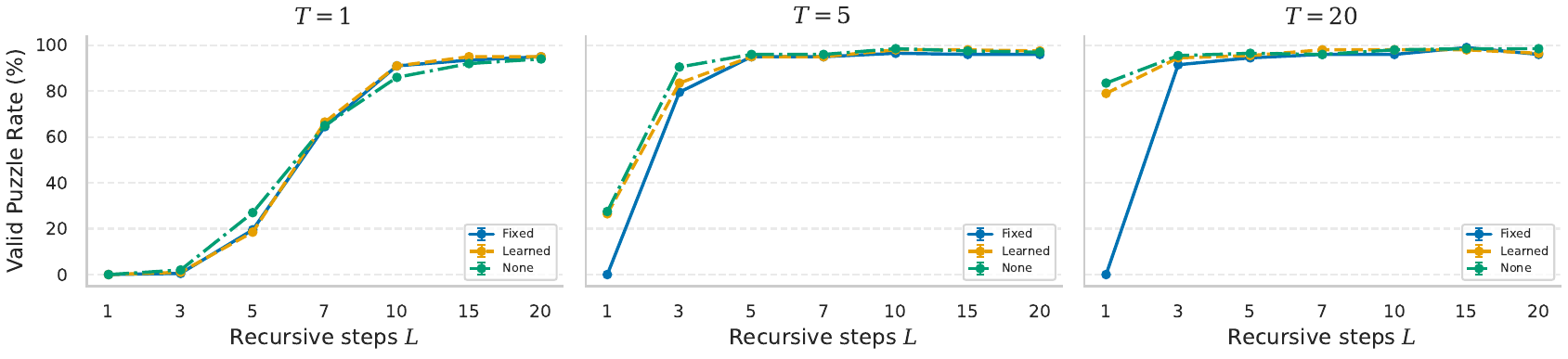}
    \caption{Performance comparison between models trained with learned, fixed and without step-embedding on $9 \times 9$ Sudoku puzzles. We report the Valid Puzzle Rate across varying sampling loop counts $L_s$ and denoising steps $T$.}
    \label{fig:ablation_embed}
    \vspace{-0.6cm}
\end{figure*}

Figure~\ref{fig:ablation_embed} shows a relatively close comparison between learned embeddings and the no-embedding baseline, with both outperforming fixed embeddings in the most important low-depth regimes. Learned embeddings perform strongly across the sweep and give the model an explicit, trainable signal for the current recursion step, which makes them a natural general-purpose choice. However, the no-embedding baseline is also very competitive. The advantage of learned embeddings is therefore not absolute; rather, they offer a balanced tradeoff: more flexible and robust than fixed embeddings, while remaining close to the best observed performance across the full grid.

\subsection{Training objectives}
\label{app:loss_mode_ablation}

Here we extend the discussion on the different supervision methods, and report the results from the ablation study.
 
\paragraph{Final-step loss (\textsc{Final}).}
Only the logits produced at the last loop, $\hat{\mathbf{y}}^{(L)}$, are supervised:
\begin{equation}
  \mathcal{L}_{\mathrm{final}}(\theta)
  = \mathbb{E}_{t,\,\xo,\,\xt}\!\left[
      \frac{1}{|\mathcal{M}|}
      \sum_{i \in \mathcal{M}}
        \mathrm{CE}\!\left(
          \hat{\mathbf{y}}^{(L)}_i,\; x_0^{(i)}
        \right)
    \right],
  \label{eq:loss_final_app}
\end{equation}
where $\mathcal{M} = \{i : x_t^{(i)} = \mask\}$ is the set of masked positions, $|\mathcal{M}|$ is its cardinality, and $\mathrm{CE}$ denotes the cross-entropy loss.  This is the natural extension of the standard single-pass \MDM{} objective to the looped setting: the intermediate loop outputs are used only as hidden-state refinements, not directly supervised.  The gradient signal reaches early loops only through backpropagation through the full chain of shared blocks.
 
\paragraph{All-steps loss (\textsc{All}).}
All $L$ sets of logits are supervised, with the loss averaged across loops:
\begin{equation}
  \mathcal{L}_{\mathrm{all}}(\theta)
  = \mathbb{E}_{t,\,\xo,\,\xt}\!\left[
      \frac{1}{L}
      \sum_{\ell=1}^{L}
        \frac{1}{|\mathcal{M}|}
        \sum_{i \in \mathcal{M}}
          \mathrm{CE}\!\left(
            \hat{\mathbf{y}}^{(\ell)}_i,\; x_0^{(i)}
          \right)
    \right].
  \label{eq:loss_all_app}
\end{equation}
This provides a denser gradient signal: every loop receives direct supervision rather than relying entirely on backpropagation through subsequent loops. This idea is closely related to deeply supervised learning, where intermediate representations are explicitly trained to be predictive~\citep{lee2014deeplysupervisednets}. It also regularizes the model to produce semantically meaningful intermediate predictions, not only a high-quality final output, the \MDM{} analogue of the per-iteration supervision in RELAY for \AR{} looped models~\citep{yu2025relay}. The all-steps loss slightly increases training memory due to storing intermediate activations for backpropagation through all loops, but it does not increase forward-pass compute.

\begin{figure*}[t]
    \centering
    \includegraphics[width=\textwidth]{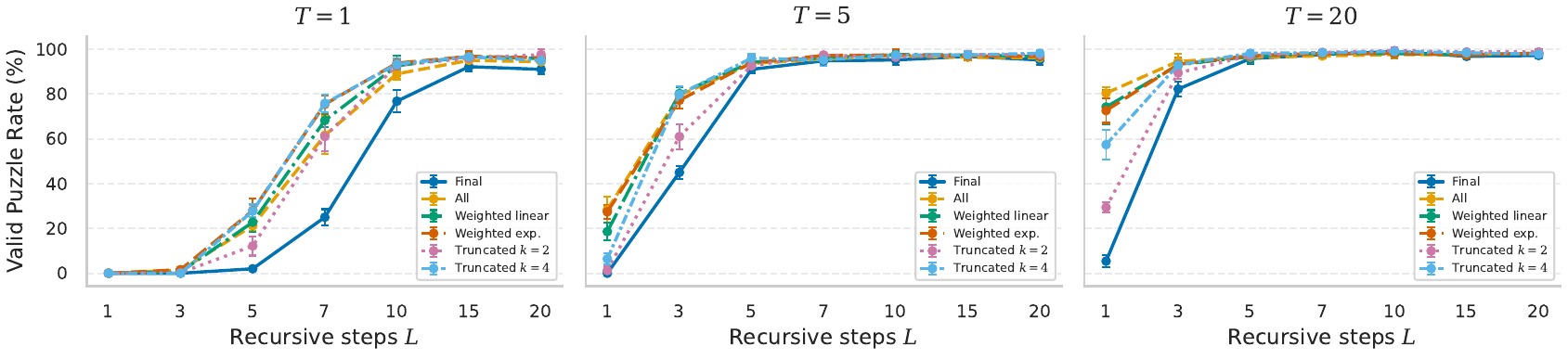}
    \caption{Performance comparison between supervision losses on $9 \times 9$ Sudoku puzzles for the model trained with $L_t = 5$ recursive steps. We report the Valid Puzzle Rate across varying loop counts at sampling $L_s$ and denoising steps $T$.}
    \label{fig:loss_vs_all}
\end{figure*}

Figure~\ref{fig:loss_vs_all} reports Sudoku 9$\times$9 Valid Puzzle Rate performance for models trained with the different losses and 5 loops at training. We show the results sampling recursive steps $L_s \in \{1,3,5,7,10,15,20\}$ and denoising steps $T \in \{1,5,20\}$.

\paragraph{Weighted loss (\textsc{Weighted}).}
Rather than supervising all loops equally, the weighted objective assigns a non-uniform weight $w_\ell$ to the loss at each loop:
\begin{equation}
  \mathcal{L}_{\mathrm{weighted}}(\theta)
  = \mathbb{E}_{t,\,\xo,\,\xt}\!\left[
      \sum_{\ell=1}^{L}
        w_\ell
        \frac{1}{|\mathcal{M}|}
        \sum_{i \in \mathcal{M}}
          \mathrm{CE}\!\left(
            \hat{\mathbf{y}}^{(\ell)}_i,\; x_0^{(i)}
          \right)
    \right]
  ,
  \label{eq:loss_weighted_app}
\end{equation}
where $w_\ell \ge 0$ and $\sum_{\ell=1}^{L} w_\ell = 1$. In our implementation, we consider two weighting schemes. For \emph{linear} weighting,
\begin{equation}
  w_\ell
  =
  \frac{\ell^{\alpha}}
       {\sum_{j=1}^{L} j^{\alpha}},
\end{equation}
while for \emph{exponential} weighting,
\begin{equation}
  w_\ell
  =
  \frac{\exp\!\bigl(\alpha(\ell-L)\bigr)}
       {\sum_{j=1}^{L}\exp\!\bigl(\alpha(j-L)\bigr)},
\end{equation}
where $\alpha>0$ controls how strongly supervision is concentrated toward later loops. Larger values of $\alpha$ place increasing emphasis on the final recursive iterations while still providing direct supervision to earlier loops. This objective is related to weighted auxiliary-loss training in anytime and deeply supervised neural networks, where intermediate predictions are optimized with unequal importance to balance early predictive capability against final performance~\citep{hu2018learninganytimepredictionsneural}.

\paragraph{Truncated loss (\textsc{Truncated}).}
The truncated objective supervises only the final $k$ recursive loops. Let $k_{\mathrm{eff}}=\min(k,L)$ and let \(\mathcal{S}_k
  =
  \{L-k_{\mathrm{eff}}+1,\ldots,L\} \)
denote the set of the last $k_{\mathrm{eff}}$ loops. The loss is then
\begin{equation}
  \mathcal{L}_{\mathrm{truncated}}(\theta)
  = \mathbb{E}_{t,\,\xo,\,\xt}\!\left[
      \frac{1}{k_{\mathrm{eff}}}
      \sum_{\ell \in \mathcal{S}_k}
        \frac{1}{|\mathcal{M}|}
        \sum_{i \in \mathcal{M}}
          \mathrm{CE}\!\left(
            \hat{\mathbf{y}}^{(\ell)}_i,\; x_0^{(i)}
          \right)
    \right]
  ,
  \label{eq:loss_truncated}
\end{equation}
This objective provides direct supervision only to the later stages of the recursive computation, which are expected to be closer to the final prediction. \textsc{Final} corresponds to truncated with $k=1$. In the more general setting, multiple late-loop predictions receive explicit supervision, improving gradient flow, while compared to \textsc{All}, early loops are not constrained to produce accurate predictions and can instead focus on building useful intermediate representations. This strategy is similar in spirit to the truncated step-range averaging used in looped transformer architectures, where training emphasizes a subset of later recurrent iterations rather than supervising the entire trajectory equally~\citep{yang2024looped}.

We observe that across panels, \textsc{All} yields the strongest and most robust performance, particularly in the regime with less recursion ($L_s$ small to moderate) and where inference is budget-constrained ($T$ small).
At $L_s{=}1$, the gap between \textsc{All} and \textsc{Final} is notable. This supports the view that dense, per-step supervision is necessary to align every recursive layer with the denoising objective, rather than relying on error signals that propagate only through the terminal state.
As $L_s$ increases, all methods improve and the curves compress, yet \textsc{All} remains at or near the top through moderate depth. We therefore adopt \textsc{All} as the default supervision strategy in subsequent experiments.

\section{Exact Dependence--Fidelity Measurement}
\label{app:multisol}

This appendix details the experiment of Section~\ref{sec:analysis_empirical}. On the held-out $9{\times}9$ boards of Section~\ref{sec:rmdm_results}, every puzzle has a unique solution so given the clues, $\pidata(x_M \mid x_O)$ is therefore a point mass. Every marginal $p^\star_i(\cdot\mid s)$ along the oracle process is a point mass as well, so $\Ddep(U\mid s)=0$ for every commit set, including one-shot decoding (Corollary~\ref{cor:refine}), and $\KL(P\Vert Q)$ reduces to the fidelity term $\sum_t \sum_{i\in U_t} -\log q_i(x^\star_i)$. Any conflict among cells committed in the same step is then a fidelity error, which is why the within-step conflict rate is only a proxy for $\Ddep$ (Remark~\ref{rem:metrics}). Separating the terms requires states at which several completions remain valid.

\paragraph{Board construction.} We start from solved held-out grids, blank a fixed number of cells uniformly at random, and enumerate every completion with an exact solver. Most random masks do not have a unique solution (Table~\ref{tab:multisol_masks}): the fraction of unique boards drops from $43\%$ at $40$ blanks to $6\%$ at $48$, and beyond $52$ blanks most boards have hundreds of completions. We fix the number of blanks at $48$, where all groups occur, so that the commit size $k=48/T$ is identical across groups and the solution count is not confounded with the number of masked cells. Boards are drawn by rejection sampling until each group $S\in\{1,\ 2\text{--}5,\ 6\text{--}50,\ 51\text{--}500\}$ holds $50$ boards; boards with more than $500$ completions are discarded, so every board has an exactly enumerated solution set. The unique group is the control, in which $\Ddep\equiv0$.

\begin{table}[h]
    \centering
    \small
    \caption{Share of random masks by number of solutions $S$, as a function of the number of blanked cells ($300$ masks on held-out grids per row). We use $48$ blanks.}
    \label{tab:multisol_masks}
    \begin{tabular}{lccccc}
        \toprule
        Blanks & $S{=}1$ & $2$--$5$ & $6$--$50$ & $51$--$500$ & $>500$ \\
        \midrule
        40 & 43\% & 44\% & 13\% & 0\% & 0\% \\
        44 & 25\% & 49\% & 26\% & 0\% & 0\% \\
        \textbf{48} & \textbf{6\%} & \textbf{28\%} & \textbf{56\%} & \textbf{11\%} & \textbf{0\%} \\
        52 & 0\% & 4\% & 38\% & 46\% & 12\% \\
        56 & 0\% & 0\% & 2\% & 19\% & 79\% \\
        \bottomrule
    \end{tabular}
\end{table}

\paragraph{Target distribution.}
We take $\pidata(\cdot\mid x_O)$ uniform over the valid completions of a board. Strictly, $\pidata$ is the empirical distribution of the $1.8$M training grids, but with about $6.7\times10^{21}$ valid $9{\times}9$ grids a board's completions are almost never in the training set, and uniformity over valid completions is the natural reading.

\paragraph{Computing both terms exactly.}
Let $\mathcal{S}$ be the set of completions of a board and, at a state $s=(x_O,M)$, let $\mathcal{C}(s)=\{\sigma\in\mathcal{S}:\sigma_O=x_O\}$ be the completions that agree with every committed cell. Under the uniform target,
\begin{equation}
    p^\star(x_U\mid s) \;=\; \frac{\big|\{\sigma\in\mathcal{C}(s) : \sigma_U = x_U\}\big|}{|\mathcal{C}(s)|},
    \qquad
    p^\star_i(v\mid s) \;=\; \frac{\big|\{\sigma\in\mathcal{C}(s) : \sigma_i = v\}\big|}{|\mathcal{C}(s)|}.
\end{equation}
The joint entropy $H(p^\star(x_U\mid s))$ is obtained by grouping $\mathcal{C}(s)$ by the projection $\sigma_U$, and $\Ddep(U\mid s)$ and $\Dfid(U\mid s,q)$ then follow from Lemma~\ref{lem:decomp}, with $q_i$ the untruncated softmax. Along the oracle process of Theorem~\ref{thm:floor}, the trajectory costs $\Dfloor(\mathcal{U})$ and $\sum_t\E_P[\Dfid]$ are averages of these per-step quantities.

The same enumeration gives the exact probability that a step fails. For the model's product of marginals,
\begin{align}
    P_q(\text{some value infeasible alone})
    &= 1-\prod_{i\in U} q_i\big(\operatorname{supp}p^\star_i\big), \\
    P_q(\text{jointly infeasible only})
    &= \prod_{i\in U} q_i\big(\operatorname{supp}p^\star_i\big)
       \;-\; \sum_{y\in\operatorname{supp}p^\star_U}\ \prod_{i\in U} q_i(y_i).
\end{align}
Replacing $q_i$ by $p^\star_i$ gives the failure probability of a model with \emph{perfect marginals}, which predicts every cell exactly but still samples the cells of a step independently. Its failures are pure dependence errors, and it is the reference shown in red in Figure~\ref{fig:errors_survival}. Its final survival is the product over steps of its per-step success probability along the oracle process, averaged over boards. This is a close approximation rather than an identity, since a perfect-marginal sampler that survives a step does not visit states in exactly the proportions of $P$.

\paragraph{Model and sampler.}
We use the recursive Sudoku checkpoint of Section~\ref{sec:analysis_empirical} (trained with a fixed $L_t{=}5$), with $L\in\{1,3,5,10\}$ and $T\in\{1,2,4,8,16,48\}$. Each step commits the next $k_t$ cells of a random permutation of the blanks drawn per board and sample before decoding. The permutation is shared across $L$, $T$ and trajectory modes, so the commit sets never depend on $q$ and $\Dfloor$ is identical for every $L$ by construction. 

In the \emph{model} mode, committed tokens are sampled from $q$ and the run stops at its first failure. In the \emph{oracle} mode, a completion is drawn uniformly from $\mathcal{S}$ and revealed along the same commit order; this is the process $P$ of Theorem~\ref{thm:floor}, along which $\Ddep$, $\Dfid$ and the failure probabilities above are evaluated (Figure~\ref{fig:dep_fid}).

\paragraph{Failure labels.}
Figure~\ref{fig:errors_survival} uses the rule-based labels of Section~\ref{sec:analysis_empirical}: a run fails at its first rule violation, either a \emph{parallel} clash between two cells committed in the same step or an \emph{old-cell} clash between a new cell and one already on the board (no invalid tokens occurred). Because every completion is known, we can also label each run by the step at which the board \emph{became uncompletable}. That step is a \emph{marginal} error if some committed value appears in no completion consistent with the board, so that it would fail even if committed alone, and a \emph{joint} error if every committed value appears in some consistent completion but no single completion contains all of them. The two labellings answer different questions:
\begin{itemize}[leftmargin=1.4em, itemsep=1pt, topsep=2pt]
    \item A board can become uncompletable without any clash (a \emph{silent dead end}), for example when $A{=}3$ forces $B{=}7$ through other cells and $B{=}3$ is committed alongside $A{=}3$ although they share no unit. The clash then surfaces at a later step.
    \item Old-cell clashes are almost never the original mistake: in $97$--$100\%$ of them, across all groups, $L$ and $T$, the board had already become uncompletable at an earlier step.
    \item Parallel clashes are mostly the original mistake. On multi-solution boards, $83$--$88\%$ of them coincide with a joint error at $L{=}5$ ($83$--$92\%$ across $L\ge5$), against $0\%$ on unique boards, where joint errors are impossible.

\end{itemize}
Final completion rates are the same under both labelings: an uncompletable board always ends in a rule violation, and a run without one ends on a valid board, which is one of the enumerated completions. Only the timing and label of failures differ.

\paragraph{Completion rates.}
Table~\ref{tab:multisol_completion} reports completion rates for all groups. On unique boards, all step budgets converge to about $98\%$ as $L$ grows: at $L{=}10$, one-shot decoding completes $97.9\%$ of boards against $0.4\%$ at $L{=}1$. On multi-solution boards, completion plateaus from $L{=}5$ on, $2$--$6$ points below the perfect-marginal ceiling set by $T$. Summed along the oracle process, the model's probability of a joint failure reaches $93$--$98\%$ of that of the perfect-marginal model at $L{=}5$ ($T\in\{2,4,8,16\}$) and does not move at $L{=}10$.

\begin{table}[h]
    \centering
    \small
    \setlength{\tabcolsep}{4pt}
    \caption{Completion rate (\%) by solution-count group, denoising steps $T$ and recursive depth $L$ ($48$ blanks, $50$ boards $\times$ $64$ samples per cell). Last column: perfect-marginal ceiling.}
    \label{tab:multisol_completion}
    \begin{tabular}{llccccc}
        \toprule
        Group & $T$ & $L{=}1$ & $L{=}3$ & $L{=}5$ & $L{=}10$ & Perfect marginals \\
        \midrule
        \multirow{4}{*}{$S{=}1$}
          & 1  & 0.4  & 50.3 & 89.9 & 97.9 & 100 \\
          & 4  & 9.7  & 73.0 & 94.1 & 98.5 & 100 \\
          & 16 & 27.5 & 85.9 & 97.8 & 99.0 & 100 \\
          & 48 & 33.7 & 90.8 & 98.7 & 99.1 & 100 \\
        \midrule
        \multirow{4}{*}{$S{=}2$--$5$}
          & 1  & 0.0  & 1.7  & 2.9  & 2.8  & 3.0 \\
          & 4  & 7.6  & 33.4 & 42.4 & 43.4 & 47.0 \\
          & 16 & 31.8 & 78.1 & 84.5 & 84.2 & 87.2 \\
          & 48 & 43.4 & 90.7 & 97.3 & 98.1 & 100 \\
        \midrule
        \multirow{4}{*}{$S{=}6$--$50$}
          & 1  & 0.0  & 0.0  & 0.0  & 0.0  & 0.0 \\
          & 4  & 4.4  & 12.8 & 14.3 & 16.3 & 19.4 \\
          & 16 & 31.6 & 65.9 & 70.6 & 71.7 & 76.1 \\
          & 48 & 46.6 & 89.7 & 95.8 & 95.6 & 100 \\
        \midrule
        \multirow{4}{*}{$S{=}51$--$500$}
          & 1  & 0.0  & 0.0  & 0.0  & 0.0  & 0.0 \\
          & 4  & 1.7  & 3.6  & 3.5  & 4.4  & 5.9 \\
          & 16 & 27.0 & 54.4 & 59.6 & 61.1 & 64.7 \\
          & 48 & 48.3 & 89.3 & 95.7 & 95.9 & 100 \\
        \bottomrule
    \end{tabular}
\end{table}

\section{The Decoding Zoo}
\label{app:decodingzoo}

Every decoding method used in this paper is a pair (loop-stopping rule, commit rule). We name a method \meth{loop\_commit}. This appendix fixes the semantics of each component, gives the full cross product of the two axes that we actually implement and sweep, and documents two relevant ablations that some of these methods additionally expose.

\subsection{Zoo overview}
\label{app:decodingzoo_overview}

Not all combinations of the loop-stopping and commit rules below are meaningful or implemented: \meth{sched} is only ever paired with \meth{uniform} (its schedule \emph{is} the loop budget, so it has nothing to add to a threshold commit rule), and the two adaptive loop rules (\meth{entropy}, \meth{hidden}) are paired with all three of \meth{uniform}, \meth{confidence}, \meth{entropy}. This gives the ten methods of Table~\ref{tab:decodingzoo_overview}.

\begin{table}[ht]
  \centering
  \small
  \caption{The decoding zoo. $\ell_t \le L$ is the number of loops run at outer step $t$; $U_t$ is the commit set; $n_{\mathrm{out}}$ is the (data-dependent) number of outer steps a T-invariant method takes. Threshold parameters are defined in Section~\ref{app:decodingzoo_loop}--Section~\ref{app:decodingzoo_commit}.}
  \label{tab:decodingzoo_overview}
  \begin{tabular}{@{}lllcl@{}}
    \toprule
    Method & Loop rule & Commit rule & Uses $T$? & Threshold(s) \\
    \midrule
    \schedunif{} & \meth{sched}: $L \!\to\! 1$ & \meth{uniform} & Yes & --- \\
    \fixunif{} & \meth{fixed}: $\ell_t = L$ & \meth{uniform} & Yes & --- \\
    \meth{fixed\_confidence} & \meth{fixed}: $\ell_t = L$ & \meth{confidence} & No & $c$ \\
    \fixent{} & \meth{fixed}: $\ell_t = L$ & \meth{entropy} & No & $\delta$ \\
    \entunif{} & \meth{entropy}-gain, $\le L$ & \meth{uniform} & Yes & $\lambda$ \\
    \entent{} & \meth{entropy}-gain, $\le L$ & \meth{entropy} & No & $\lambda$, $\delta$ \\
    \entconf{} & \meth{entropy}-gain, $\le L$ & \meth{confidence} & No & $\lambda$, $c$ \\
    \meth{hidden\_uniform} & \meth{hidden}-drift, $\le L$ & \meth{uniform} & Yes & $\varepsilon$ \\
    \hident{} & \meth{hidden}-drift, $\le L$ & \meth{entropy} & No & $\varepsilon$, $\delta$ \\
    \meth{hidden\_confidence} & \meth{hidden}-drift, $\le L$ & \meth{confidence} & No & $\varepsilon$, $c$ \\
    \bottomrule
  \end{tabular}
\end{table}

The nominal step count $T$ enters in only two places: the \meth{uniform} commit rule, which reveals $|M_0|/T$ positions per step, and the \meth{sched} loop schedule, which anneals over $T$ steps. Every method pairing a \meth{fixed} or adaptive loop rule with a \emph{threshold} commit rule therefore never reads $T$: its outer loop runs until no position is masked, and both its quality and its $\NFE$ are constant along the $T$ axis, and it is why those methods are reported once per $L$ rather than once per $(L,T)$ pair. 

Compute is billed as $\NFE = \sum_t \ell_t$. For the two \meth{uniform}-commit, T-scheduled methods with a non-adaptive loop rule, the number of outer steps is $\min(T,|M_0|)$, since \meth{uniform} commits at least one position per step; \fixunif{} therefore bills $\NFE = L\min(T,|M_0|)$, which equals $TL$ whenever $T\le|M_0|$ (\schedunif{} instead averages to $\NFE \approx T(L{+}1)/2$). For every other method $\NFE = \sum_t \ell_t \le n_{\mathrm{out}} L$, with equality only for the two \meth{fixed}-loop, threshold-commit methods (\meth{fixed\_confidence}, \fixent{}); the four \meth{entropy}- and \meth{hidden}-loop methods with a threshold commit rule additionally have $\ell_t \le L$ per step, so their realized $\NFE$ is data-dependent on \emph{both} axes at once.

Unless noted otherwise, decoding uses the \fixunif{} policy: over $T$ outer denoising steps the remaining masked positions are scored by top-$k$ confidence, and at each step a nearly equal share of them is committed (highest confidence first). Token values are drawn from temperature-scaled top-$k$ logits, while already-revealed (and prompt) tokens stay fixed. Per-dataset evaluation protocols are given in Appendix~\ref{app:datasets}, and the temperature and top-$k$ values in Table~\ref{tab:hp-choices}.

\subsection{Loop-stopping rules (how many inner loops at a state)}
\label{app:decodingzoo_loop}

At outer step $t$ the policy executes $\ell_t \le L$ loops at the frozen state $s_t$, where $L$ is the loop budget and $\ell_t$ is chosen by one of:

\begin{itemize}[leftmargin=1.4em, itemsep=2pt]
    \item \meth{fixed} --- always $\ell_t = L$. Non-adaptive; bills $\NFE = L \times T$ exactly.
    \item \meth{sched} --- a decreasing schedule that anneals the per-step loop count linearly from $L$ at $t=0$ down to $1$ at $t=T-1$, on the premise that the early, mostly-masked states need more recursion to resolve than the late, mostly-committed ones. This gives a mean depth of roughly $(L{+}1)/2$.
    \item \meth{entropy} --- \emph{entropy-gain} stopping: run the first loop unconditionally, then after each loop $\ell\geq2$ compute the mean Shannon entropy over still-masked positions $\bar H(q^{(\ell)})$, and continue while the gain
    \[
        \bar H(q^{(\ell-1)}) - \bar H(q^{(\ell)}) \ \ge\ \lambda.
    \]
    \item \meth{hidden} --- \emph{hidden-drift} stopping: continue while the relative hidden-state drift
    \[
        \frac{\Vert \mathbf{h}^{(\ell)} - \mathbf{h}^{(\ell-1)}\Vert}{\Vert \mathbf{h}^{(\ell)}\Vert} \ \ge\ \varepsilon.
    \] 
\end{itemize}

Adaptive rules ($\meth{entropy}$, $\meth{hidden}$) never exceed the budget $L$, so $\ell_t \le L$ always. Paired with \meth{uniform} commits the outer loop runs exactly $T$ steps, giving $\NFE \le LT$ with the gap being the compute the rule saves; paired with a threshold commit rule the number of outer steps $n_{\mathrm{out}}$ is data-dependent, so the bound is $\NFE \le n_{\mathrm{out}} L$ instead. When an \meth{entropy}-loop rule stops, the marginals it deploys may be taken from the last loop, the previous loop, or the best-scoring loop seen (the \emph{commit-loop policy}, see ablation in Appendix~\ref{app:decodingzoo_ablations}). Note that a small entropy gain does not imply that the two loops produce similar distributions: entropy is a scalar summary, and two distributions with equal entropy can be far apart in total variation. The choice is therefore settled empirically, not by the analysis.

\subsection{Commit rules (which masked positions to reveal)}
\label{app:decodingzoo_commit}

Given the deployed marginals $q_i$ at the masked positions, the commit set $U_t$ is chosen by one of:

\begin{itemize}[leftmargin=1.4em, itemsep=2pt]
    \item \meth{uniform} --- reveal a fixed number $k = |M_0|/T$ of positions per step, taken in order of decreasing confidence, so that all $n$ positions are committed in $T$ steps. This method carries no dependence certificate: it fixes $|U_t|$ regardless of realized dependence.
    \item \meth{confidence} --- reveal every position whose top-1 probability exceeds a threshold $c$. This is the standard MaskGIT-style schedule of~\citet{chang2022maskgit}.
    \item \meth{entropy} --- reveal the largest low-entropy set satisfying the budget $\sum_{i \in U} H_i \le \delta$, built greedily by revealing the lowest-entropy masked positions first. 
\end{itemize}

Both threshold rules (\meth{confidence}, \meth{entropy}) force at least one commit (either the most confident position, or the single lowest-entropy one) when no candidate qualifies, so decoding terminates in at most $n$ steps. 

\begin{algorithm}[h]
\caption{Entropy-guided recursive decoding (\entent{})}
\label{alg:entent}
\small
\setboolean{ALC@noend}{true}
\begin{algorithmic}[1]
\STATE {\bfseries Input:} R-\MDM{} $(f_\theta, W_\theta)$, state $s_0=(x_O, M_0)$, loop budget $L$, thresholds $\lambda, \delta$
\WHILE{$M_t \neq \emptyset$}
  \FOR{$\ell = 1, \dots, L$}
    \STATE compute $\mathbf{h}^{(\ell)}$ and marginals $q^{(\ell)}_i(\cdot \mid s_t)$ for $i \in M_t$ \hfill (\Eqref{eq:hloop})
    \STATE \textbf{if} $\ell \ge 2$ \textbf{and} $\bar H(q^{(\ell-1)}) - \bar H(q^{(\ell)}) < \lambda$ \textbf{then break} \hfill $\triangleright$ loop stopping
  \ENDFOR
  \STATE $U_t \gets$ lowest-entropy positions of $M_t$ with $\sum_{i \in U_t} H(q^{(\ell)}_i) \le \delta$, at least one \hfill $\triangleright$ commit
  \STATE sample $x_i \sim q^{(\ell)}_i$ independently for $i \in U_t$; \ $M_{t+1} \gets M_t \setminus U_t$
\ENDWHILE
\STATE {\bfseries return} the completed sequence
\end{algorithmic}
\end{algorithm}

\subsection{Ablation on the decoding mechanisms}
\label{app:decodingzoo_ablations}

We perform two focused ablations to isolate design choices in the adaptive recursive decoder.  Both experiments use the Sudoku generation setting and evaluate 50 generated boards per configuration.  We report Valid Puzzle Rate (VPR) as the task-quality metric and the mean number of recursive forward passes as the compute metric.  The latter is important because adaptive decoding policies do not necessarily spend the full nominal recursion budget: they stop once the mean entropy gain of the inner refinement loop falls below $\lambda$.  Thus, the relevant comparison is not only which method reaches high validity, but which method reaches it at the lowest effective number of function evaluations (NFE).

\paragraph{Commit-Loop Policy}

For entropy-controlled recursive decoders, the inner loop runs until the average entropy over still-masked positions stops decreasing by more than a threshold $\lambda$.  If this stopping condition triggers after loop $\ell$, there are two nearby predictive distributions available: the previous distribution $q^{\ell-1}$ and the current distribution $q^\ell$.  We ablate three policies for choosing which distribution is used to sample and commit tokens:
\[
\texttt{prev}: q^{\ell-1}, \qquad
\texttt{curr}: q^\ell, \qquad
\texttt{best}: 
\begin{cases}
q^\ell, & \text{if entropy improved at the confirming step},\\
q^{\ell-1}, & \text{otherwise}.
\end{cases}
\]
The \texttt{prev} policy is conservative: it avoids committing from the step that first indicates saturation.  The \texttt{curr} policy is more aggressive: it uses the latest refined logits, even though the marginal entropy improvement has become small.  The \texttt{best} policy is an intermediate rule that attempts to keep the better of the two local iterates.

\begin{figure}[t]
    \centering
    \includegraphics[width=\linewidth]{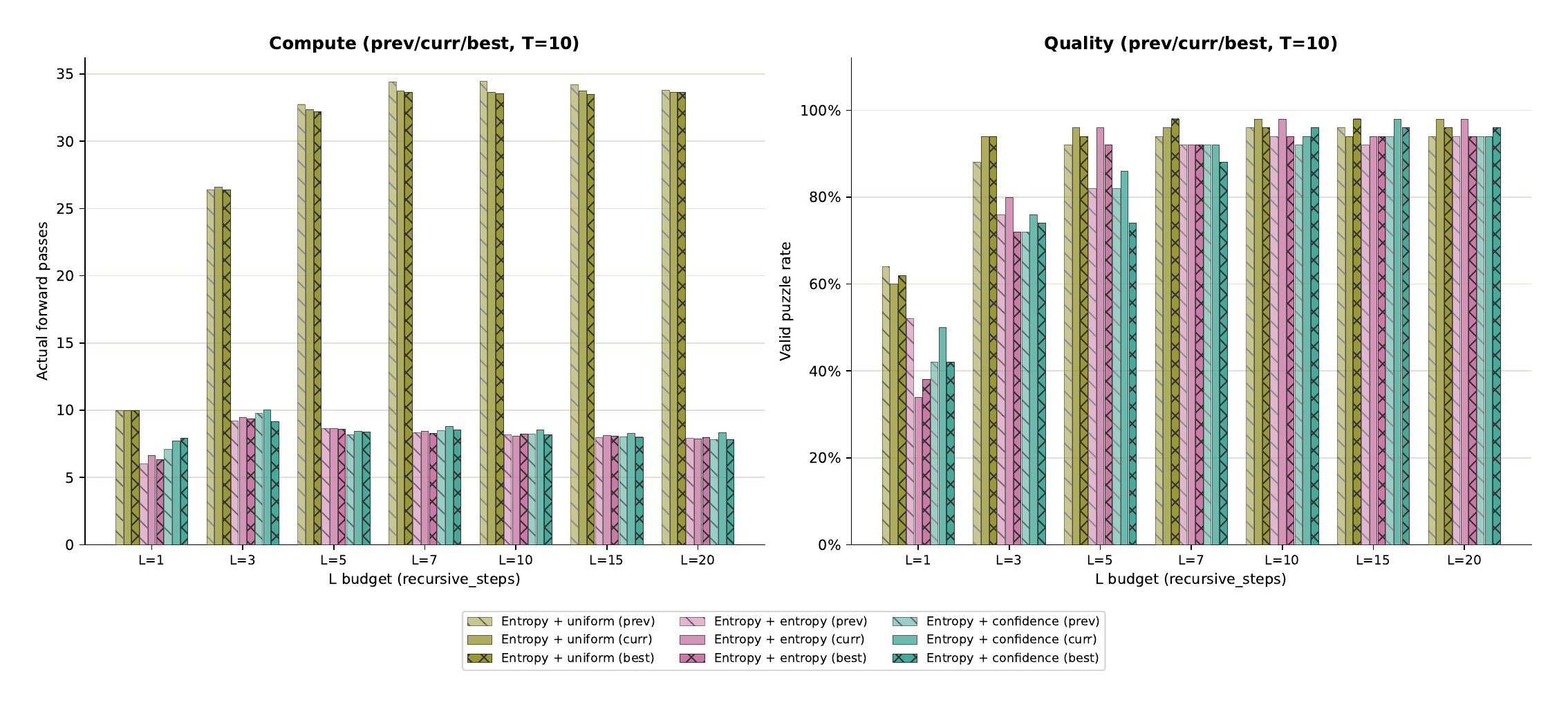}
    \caption{Commit-loop policy ablation for entropy-controlled recursive decoding.}
    \label{fig:curr_prev_best_ablation}
\end{figure}

The ablation (see Figure~\ref{fig:curr_prev_best_ablation}) supports \texttt{curr} as the default policy: it achieves the strongest performance in the adaptive entropy-budget setting, while remaining competitive under fixed-$T$ decoding. This suggests that using the current loop's logits for commitment is preferable to reverting to stale logits, particularly when commit size is adaptively determined from model uncertainty.

\paragraph{Entropy versus EB Commit Rule}

We next compare the standard entropy-budget commit rule against an EB-style alternative~\citep{benhamu2025ebsampler}.  The standard entropy rule sorts still-masked positions by predictive entropy and commits the largest low-entropy set $S$ satisfying
\[
    \sum_{i \in S} H(q_i) \leq \delta,
\]
with at least one token committed at every outer step.  The EB variant replaces this with the looser bound
\[
    \sum_{i \in S} H(q_i) - \max_{i \in S} H(q_i) \leq \gamma,
\]
using the same numerical budget value as the standard entropy rule.  Intuitively, the EB rule permits larger simultaneous commits by subtracting the largest individual uncertainty term.  This can reduce compute, but it also risks committing mutually incompatible tokens earlier.

\begin{figure}[t]
    \centering
    \includegraphics[trim={0 0 0 25.1cm},clip,width=0.9\linewidth]{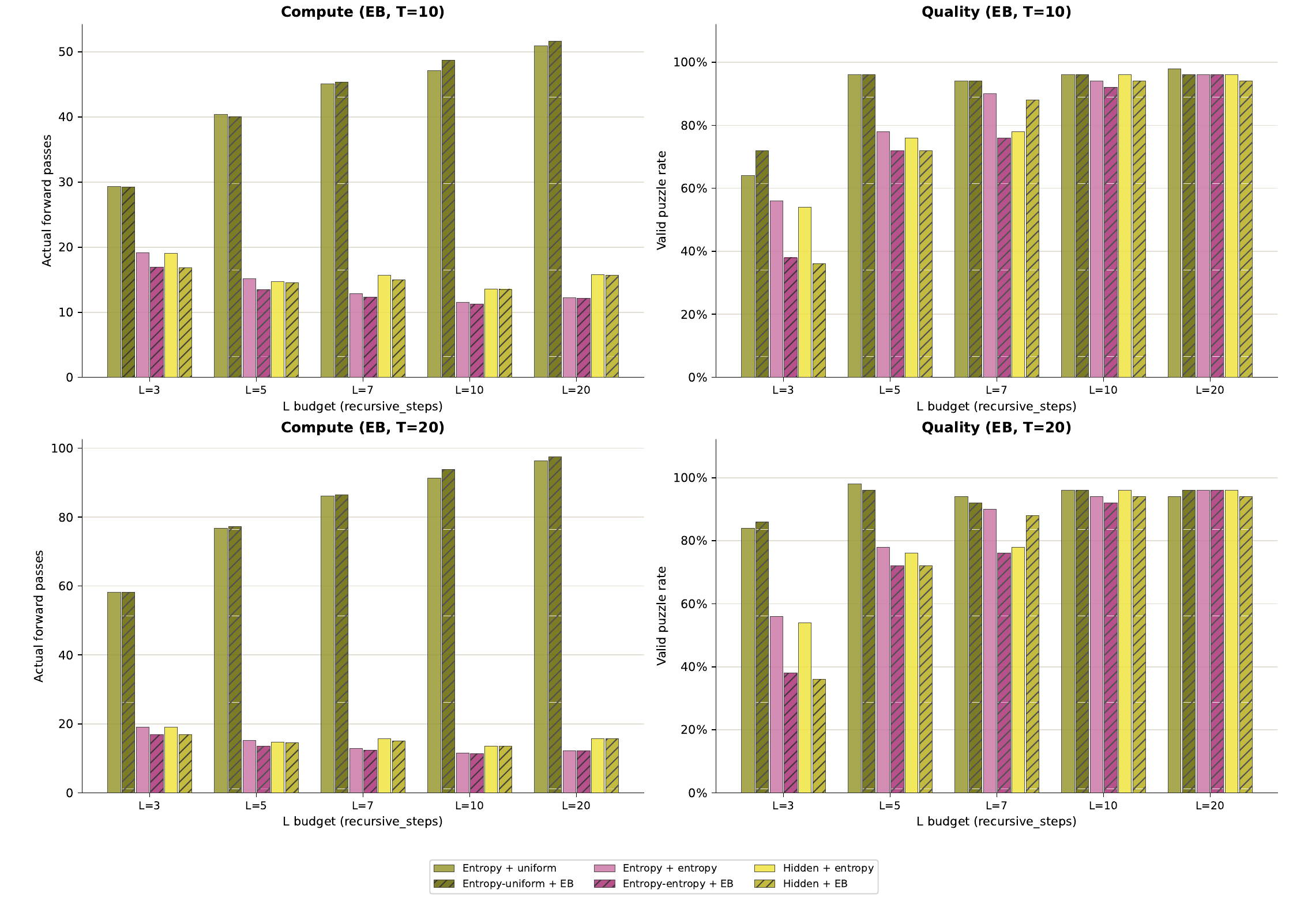}
    \caption{Comparison between standard entropy-budget decoding and the EB commit rule.}
    \label{fig:entropy_vs_eb_ablation}
\end{figure}

The results (see Figure~\ref{fig:entropy_vs_eb_ablation}) do not show a consistent benefit from the EB rule. While \texttt{entropy} and \texttt{entropy\_eb} perform similarly, the standard entropy budget is more reliable for adaptive entropy and hidden-entropy decoding: \texttt{adaptive\_entropy} reaches 0.90 and 0.94 VPR at $L=7$ and $L=10$, compared with 0.76 and 0.92 for its EB variant, with similar trends for hidden entropy. This suggests that the stricter sum-entropy criterion better preserves validity under Sudoku's global constraints, where parallel commitment of uncertain cells can create irrecoverable conflicts. We therefore use the standard entropy rule.

\subsection{Hyperparameter tuning}
\label{sec:hp-tuning}

We tune four decoding hyperparameters per dataset:

\begin{itemize}
    \item $\lambda$ on \texttt{entropy\_*}, $\lambda \in \{10^{-3}, 5{\times}10^{-3}, 10^{-2}, 5{\times}10^{-2}, 10^{-1}, 5{\times}10^{-1}\}$
    \item $\delta$ on \texttt{*\_entropy}, $\delta \in \{0.25, 0.5, 1, 2, 4, 8\}$,
    \item $\varepsilon$ on \texttt{hidden\_*}, $\varepsilon \in \{0.01, 0.05, 0.1, 0.2, 0.5, 1.0\}$,
    \item $c$ on \texttt{*\_confidence}, $c \in \{0.5, 0.7, 0.8, 0.9, 0.95, 0.99\}$.
\end{itemize}

We pick one value per dataset from quality--NFE plots on the validation split, favoring near-frontier settings at moderate $L$ and lower NFE.

\begin{table}[h]
  \centering
  \caption{Decoding settings per dataset. The thresholds $\lambda$, $\delta$, $\varepsilon$ and $c$ are
  selected as described above; the sampling temperature and top-$k$ are fixed per dataset and not tuned.}
  \label{tab:hp-choices}
  \small
  \begin{tabular}{lcccccc}
    \toprule
    Dataset & $\lambda$ & $\delta$ & $\varepsilon$ & $c$ & Temp. & top-$k$ \\
    \midrule
    Sudoku 9$\times$9   & 0.01 & 0.25 & 0.20 & 0.95 & 0.8 & 2 \\
    Countdown-3         & 0.1 & 0.25 & 0.50 & 0.90 & 0.5 & 1 \\
    Countdown-4         & 0.01 & 0.25 & 0.10 & 0.95 & 0.5 & 1 \\
    Countdown-5         & 0.05 & 0.50 & 0.10 & 0.95 & 0.5 & 1 \\
    RNA                 & 0.05 & 4.00 & 0.05 & 0.95 & 0.8 & 2 \\
    TinyGSM             & 0.01 & 2.00 & 0.05 & 0.95 & 0.8 & 1 \\
    \bottomrule
  \end{tabular}
\end{table}

\clearpage
\section{Extended Results}
\label{app:extendedresults}

\subsection{Sudoku 9$\times$9: Full Tables}
\label{app:sudoku_full}
 
Table~\ref{tab:sudoku_full} presents the comprehensive Sudoku 9$\times$9 results across recursion depth and denoising step budget, for the all-loops objective $\mathcal{L}_{\text{all}}$ used throughout the paper. Latency and throughput figures are measured at batch size 1 on a single A100. Rows labelled \emph{Baseline} correspond to $(6 \otimes 1)$ without any recursion, and rows labeled $k$~\emph{rec steps} to a $(6 \otimes k)$ model. 

\begin{table*}[ht]
    \caption{Sudoku $9\times9$: valid puzzle rate, soft constraint loss, latency and throughput across recursion depths and denoising budgets. All models are trained with the all-loops objective $\mathcal{L}_{\mathrm{all}}$; the comparison against $\mathcal{L}_{\mathrm{final}}$ is in Appendix~\ref{app:loss_mode_ablation}. We report mean and standard deviation across 5 sampling runs of 100 samples each.}
    \label{tab:sudoku_full}
    \vspace{-6pt}
    \centering
    \resizebox{0.72\linewidth}{!}{
    \footnotesize
    \begin{tabular}{l ccccc}
    \toprule
    \textbf{Model} & \textbf{Steps} & \textbf{VPR \%} & \textbf{SCL} & \textbf{Latency (s)} & \textbf{Throughput} \\
    \midrule
    \multirow{5}{*}{Baseline} & 1 & 0.0\% $\pm$ 0.0 & 3.077 $\pm$ 0.039 & 0.007 & 149.6 \\
     & 5 & 65.2\% $\pm$ 6.9 & 0.226 $\pm$ 0.052 & 0.022 & 44.9 \\
     & 10 & 85.0\% $\pm$ 2.7 & 0.088 $\pm$ 0.024 & 0.043 & 23.3 \\
     & 20 & 92.2\% $\pm$ 2.7 & 0.031 $\pm$ 0.017 & 0.084 & 11.9 \\
     & 40 & 95.8\% $\pm$ 1.9 & 0.011 $\pm$ 0.006 & 0.169 & 5.9 \\
    \midrule
    \multirow{5}{*}{2 rec steps} & 1 & 0.0\% $\pm$ 0.0 & 2.443 $\pm$ 0.080 & 0.010 & 102.1 \\
     & 5 & 83.2\% $\pm$ 3.6 & 0.120 $\pm$ 0.026 & 0.039 & 25.8 \\
     & 10 & 91.2\% $\pm$ 1.6 & 0.055 $\pm$ 0.017 & 0.079 & 12.7 \\
     & 20 & 95.4\% $\pm$ 2.8 & 0.024 $\pm$ 0.015 & 0.153 & 6.6 \\
     & 40 & 96.4\% $\pm$ 1.8 & 0.016 $\pm$ 0.009 & 0.298 & 3.4 \\
    \midrule
    \multirow{5}{*}{3 rec steps} & 1 & 8.6\% $\pm$ 3.0 & 1.645 $\pm$ 0.051 & 0.020 & 50.5 \\
     & 5 & 92.4\% $\pm$ 2.5 & 0.056 $\pm$ 0.018 & 0.072 & 13.9 \\
     & 10 & 95.6\% $\pm$ 2.1 & 0.023 $\pm$ 0.015 & 0.131 & 7.7 \\
     & 20 & 96.0\% $\pm$ 2.2 & 0.024 $\pm$ 0.016 & 0.234 & 4.3 \\
     & 40 & 98.0\% $\pm$ 0.7 & 0.008 $\pm$ 0.005 & 0.456 & 2.2 \\
    \midrule
    \multirow{5}{*}{5 rec steps} & 1 & 21.2\% $\pm$ 2.4 & 1.093 $\pm$ 0.113 & 0.084 & 11.9 \\
     & 5 & 93.6\% $\pm$ 2.9 & 0.050 $\pm$ 0.026 & 0.210 & 4.8 \\
     & 10 & 97.4\% $\pm$ 1.3 & 0.016 $\pm$ 0.014 & 0.362 & 2.8 \\
     & 20 & 96.4\% $\pm$ 1.5 & 0.016 $\pm$ 0.007 & 0.696 & 1.4 \\
     & 40 & 98.0\% $\pm$ 1.6 & 0.008 $\pm$ 0.006 & 1.119 & 0.9 \\
    \midrule
    \multirow{5}{*}{10 rec steps} & 1 & 46.4\% $\pm$ 8.0 & 0.670 $\pm$ 0.156 & 0.036 & 27.8 \\
     & 5 & 95.2\% $\pm$ 2.4 & 0.034 $\pm$ 0.016 & 0.171 & 5.8 \\
     & 10 & 96.0\% $\pm$ 1.6 & 0.026 $\pm$ 0.016 & 0.504 & 2.0 \\
     & 20 & 97.2\% $\pm$ 1.3 & 0.017 $\pm$ 0.012 & 1.232 & 0.8 \\
     & 40 & 97.0\% $\pm$ 1.9 & 0.014 $\pm$ 0.010 & 3.457 & 0.3 \\
    \bottomrule
    \end{tabular}}
\end{table*}%

\newpage
Table~\ref{tab:sudoku_sample} presents the Sudoku 9$\times$9 results for the (\textit{Schedule}$10 \to 1$) training recursion depth and for different evaluated sampling recursion depths ($L_s$).

\begin{table*}[ht]
    \caption{Sudoku Performance: comprehensive comparison across sampling recursion depths ($L_s$) and decoding steps ($T$). We report mean and standard deviation across 5 sampling runs of 100 samples each.}
    \label{tab:sudoku_sample}
    \vspace{-6pt}
    \centering
    \resizebox{0.7\linewidth}{!}{
    \footnotesize
    \begin{tabular}{ll cccc}
    \toprule
    \textbf{Recursion $L_s$} & \textbf{Steps} & \textbf{VPR \%} & \textbf{SCL} & \textbf{Latency (s)} & \textbf{Throughput} \\
    \midrule
    \multirow{5}{*}{1 rec step} & 1  & 0.0\% $\pm$ 0.0    & 3.729 $\pm$ 0.034      & 0.018 & 55.0 \\
                              & 5  & 24.0\% $\pm$ 2.5   & 0.587 $\pm$ 0.033      & 0.045 & 22.3 \\
                              & 10 & 63.6\% $\pm$ 2.5   & 0.210 $\pm$ 0.034      & 0.087 & 11.5 \\
                              & 20 & 81.4\% $\pm$ 2.5   & 0.084 $\pm$ 0.008      & 0.143 & 7.0 \\
                              & 40 & 85.2\% $\pm$ 0.8   & 0.050 $\pm$ 0.003      & 0.286 & 3.5 \\
    \midrule
    \multirow{5}{*}{3 rec steps} & 1  & 3.8\% $\pm$ 1.3    & 1.791 $\pm$ 0.046      & 0.024 & 41.8 \\
                              & 5  & 95.6\% $\pm$ 0.9   & 0.025 $\pm$ 0.009      & 0.081 & 12.3 \\
                              & 10 & 95.8\% $\pm$ 1.8   & 0.021 $\pm$ 0.007      & 0.161 & 6.2 \\
                              & 20 & 98.2\% $\pm$ 0.8   & 0.012 $\pm$ 0.011      & 0.276 & 3.6 \\
                              & 40 & 99.0\% $\pm$ 1.0   & 0.004 $\pm$ 0.004      & 0.608 & 1.6 \\
    \midrule
    \multirow{5}{*}{5 rec steps} & 1  & 63.6\% $\pm$ 1.5   & 0.426 $\pm$ 0.019      & 0.032 & 30.9 \\
                              & 5  & 96.8\% $\pm$ 0.8   & 0.022 $\pm$ 0.007      & 0.120 & 8.3 \\
                              & 10 & 97.0\% $\pm$ 1.6   & 0.020 $\pm$ 0.007      & 0.235 & 4.2 \\
                              & 20 & 97.8\% $\pm$ 0.8   & 0.012 $\pm$ 0.006      & 0.446 & 2.2 \\
                              & 40 & 99.8\% $\pm$ 0.4   & 0.001 $\pm$ 0.002      & 0.940 & 1.1 \\
    \midrule
    \multirow{5}{*}{7 rec steps} & 1  & 89.8\% $\pm$ 0.8   & 0.113 $\pm$ 0.007      & 0.040 & 25.1 \\
                              & 5  & 97.4\% $\pm$ 0.9   & 0.021 $\pm$ 0.004      & 0.159 & 6.3 \\
                              & 10 & 96.4\% $\pm$ 0.9   & 0.015 $\pm$ 0.004      & 0.339 & 3.0 \\
                              & 20 & 96.2\% $\pm$ 0.8   & 0.018 $\pm$ 0.004      & 0.573 & 1.7 \\
                              & 40 & 97.6\% $\pm$ 0.5   & 0.011 $\pm$ 0.002      & 1.279 & 0.8 \\
    \midrule
    \multirow{5}{*}{10 rec steps} & 1  & 94.2\% $\pm$ 0.4   & 0.082 $\pm$ 0.013      & 0.049 & 20.4 \\
                              & 5  & 97.4\% $\pm$ 0.5   & 0.021 $\pm$ 0.002      & 0.212 & 4.7 \\
                              & 10 & 98.4\% $\pm$ 0.9   & 0.005 $\pm$ 0.003      & 0.435 & 2.3 \\
                              & 20 & 96.8\% $\pm$ 0.4   & 0.014 $\pm$ 0.004      & 0.806 & 1.2 \\
                              & 40 & 98.0\% $\pm$ 0.0   & 0.007 $\pm$ 0.000      & 1.565 & 0.6 \\
    \midrule
    \multirow{5}{*}{15 rec steps} & 1  & 95.0\% $\pm$ 0.0   & 0.062 $\pm$ 0.010      & 0.053 & 18.9 \\
                              & 5  & 97.0\% $\pm$ 0.0   & 0.016 $\pm$ 0.001      & 0.326 & 3.1 \\
                              & 10 & 97.2\% $\pm$ 0.4   & 0.014 $\pm$ 0.002      & 0.613 & 1.6 \\
                              & 20 & 98.0\% $\pm$ 0.0   & 0.010 $\pm$ 0.002      & 1.183 & 0.8 \\
                              & 40 & 98.0\% $\pm$ 0.0   & 0.007 $\pm$ 0.000      & 1.822 & 0.5 \\
    \midrule
    \multirow{5}{*}{20 rec steps} & 1  & 95.0\% $\pm$ 0.0   & 0.060 $\pm$ 0.006      & 0.071 & 14.0 \\
                              & 5  & 97.0\% $\pm$ 0.0   & 0.018 $\pm$ 0.003      & 0.373 & 2.7 \\
                              & 10 & 97.6\% $\pm$ 0.5   & 0.007 $\pm$ 0.002      & 0.726 & 1.4 \\
                              & 20 & 97.0\% $\pm$ 0.0   & 0.018 $\pm$ 0.003      & 1.579 & 0.6 \\
                              & 40 & 98.0\% $\pm$ 0.0   & 0.007 $\pm$ 0.000      & 2.416 & 0.4 \\
    \bottomrule
    \end{tabular}}
\end{table*}

\clearpage
\subsection{Sudoku 9$\times$9: Depth Scaling Without Recursion}
\label{app:depth_full}
 
Table~\ref{tab:sudoku_extended_layers_full} reports Valid Puzzle Rate and Soft Constraint Loss metrics for non-recursive baselines with 6, 8, 10, 12, 18, 24,
and 30 transformer layers. Parameter counts range from 10.6M to 53.1M. All models share the same $d{=}384$, $H{=}6$ architecture and are trained for the same number of gradient steps as the recursive models.
 
Notably, increasing depth beyond 18 layers produces only marginal improvements on VPR (96.6\%$\to$97.2\%$\to$97.2\% for 18$\to$24$\to$30 layers at $T{=}40$), whereas increasing $L$ from 1 to 5 at fixed 6-layer parameter count produces a larger gain (95.8\%$\to$98.0\%).

\begin{table*}[ht]
    \caption{Sudoku performance for baseline models without recursion across different depths (6--30 layers). We report mean and standard deviation across 5 sampling runs of 100 samples each.}
    \label{tab:sudoku_extended_layers_full}
    \vspace{-6pt}
    \centering
    \resizebox{0.7\linewidth}{!}{
    \footnotesize
    \begin{tabular}{l cccccc}
    \toprule
    \textbf{Model} & \textbf{Steps} & \textbf{VPR \%} & \textbf{SCL} & \textbf{Latency (s)} & \textbf{Throughput} \\
    \midrule
    \multirow{5}{*}{6 layers (10.6M)} & 1 & 0.0\% $\pm$ 0.0 & 3.077 $\pm$ 0.039 & 0.007 & 149.6 \\
     & 5 & 65.2\% $\pm$ 6.9 & 0.226 $\pm$ 0.052 & 0.022 & 44.9 \\
     & 10 & 85.0\% $\pm$ 2.7 & 0.088 $\pm$ 0.024 & 0.043 & 23.3 \\
     & 20 & 92.2\% $\pm$ 2.7 & 0.031 $\pm$ 0.017 & 0.084 & 11.9 \\
     & 40 & 95.8\% $\pm$ 1.9 & 0.011 $\pm$ 0.006 & 0.169 & 5.9 \\
    \midrule
    \multirow{5}{*}{8 layers (14.2M)} & 1  & 0.0\% $\pm$ 0.0 & 3.209 $\pm$ 0.053 & 0.010 & 104.8 \\
     & 5  & 58.6\% $\pm$ 6.4 & 0.291 $\pm$ 0.038 & 0.033 & 30.6  \\
     & 10 & 83.0\% $\pm$ 3.7 & 0.099 $\pm$ 0.018 & 0.072 & 13.9  \\
     & 20 & 90.2\% $\pm$ 3.5 & 0.045 $\pm$ 0.016 & 0.151 & 6.6   \\
     & 40 & 94.4\% $\pm$ 2.0 & 0.020 $\pm$ 0.007 & 0.370 & 2.7   \\
    \midrule
    \multirow{5}{*}{10 layers (17.7M)} & 1  & 0.0\% $\pm$ 0.0 & 3.100 $\pm$ 0.098 & 0.010 & 102.4  \\
     & 5  & 60.4\% $\pm$ 5.9 & 0.302 $\pm$ 0.055 & 0.046 & 22.0  \\
     & 10 & 83.8\% $\pm$ 3.0 & 0.098 $\pm$ 0.016 & 0.094 & 10.6  \\
     & 20 & 91.4\% $\pm$ 4.3 & 0.047 $\pm$ 0.028 & 0.211 & 4.7   \\
     & 40 & 94.6\% $\pm$ 1.1 & 0.024 $\pm$ 0.004 & 0.470 & 2.1   \\
    \midrule
    \multirow{5}{*}{12 layers (21.2M)} & 1  & 0.0\% $\pm$ 0.0 & 3.025 $\pm$ 0.057 & 0.010 & 101.5 \\
     & 5  & 65.4\% $\pm$ 3.4 & 0.262 $\pm$ 0.034 & 0.040 & 24.8  \\
     & 10 & 87.6\% $\pm$ 4.8 & 0.064 $\pm$ 0.022 & 0.082 & 12.1  \\
     & 20 & 93.6\% $\pm$ 2.9 & 0.031 $\pm$ 0.015 & 0.276 & 3.6   \\
     & 40 & 93.8\% $\pm$ 1.9 & 0.018 $\pm$ 0.006 & 1.005 & 1.0   \\
    \midrule
    \multirow{5}{*}{18 layers (31.9M)} & 1  & 0.0\% $\pm$ 0.0 & 2.622 $\pm$ 0.083 & 0.023 & 43.0  \\
     & 5  & 80.6\% $\pm$ 3.2 & 0.144 $\pm$ 0.021 & 0.085 & 11.7  \\
     & 10 & 95.2\% $\pm$ 2.4 & 0.028 $\pm$ 0.017 & 0.187 & 5.4   \\
     & 20 & 94.8\% $\pm$ 1.5 & 0.027 $\pm$ 0.008 & 0.535 & 1.9   \\
     & 40 & 96.6\% $\pm$ 0.6 & 0.014 $\pm$ 0.005 & 1.112 & 0.9   \\
    \midrule
    \multirow{5}{*}{24 layers (42.5M)} & 1  & 0.2\% $\pm$ 0.5 & 2.710 $\pm$ 0.056 & 0.048 & 21.1  \\
     & 5  & 82.8\% $\pm$ 3.6 & 0.127 $\pm$ 0.025 & 0.167 & 6.0   \\
     & 10 & 92.2\% $\pm$ 2.8 & 0.042 $\pm$ 0.019 & 0.365 & 2.7   \\
     & 20 & 95.0\% $\pm$ 2.5 & 0.023 $\pm$ 0.013 & 0.724 & 1.4   \\
     & 40 & 97.2\% $\pm$ 0.8 & 0.008 $\pm$ 0.002 & 1.435 & 0.7   \\
    \midrule
    \multirow{5}{*}{30 layers (53.1M)} & 1  & 0.2\% $\pm$ 0.5 & 2.624 $\pm$ 0.093 & 0.050 & 19.9  \\
     & 5  & 77.8\% $\pm$ 2.2 & 0.159 $\pm$ 0.018 & 0.160 & 6.2   \\
     & 10 & 91.4\% $\pm$ 1.3 & 0.044 $\pm$ 0.010 & 0.284 & 3.5   \\
     & 20 & 95.0\% $\pm$ 2.0 & 0.021 $\pm$ 0.006 & 0.547 & 1.8   \\
     & 40 & 97.2\% $\pm$ 0.8 & 0.012 $\pm$ 0.004 & 1.059 & 0.9   \\
    \bottomrule
    \end{tabular}}
\end{table*}

\subsection{Sudoku 9$\times$9: Cross-Recursion Trade-off and Efficient Compute}
\label{app:crossrec_table}
 
\begin{figure}[t]
    \centering
    \includegraphics[width=0.65\textwidth]{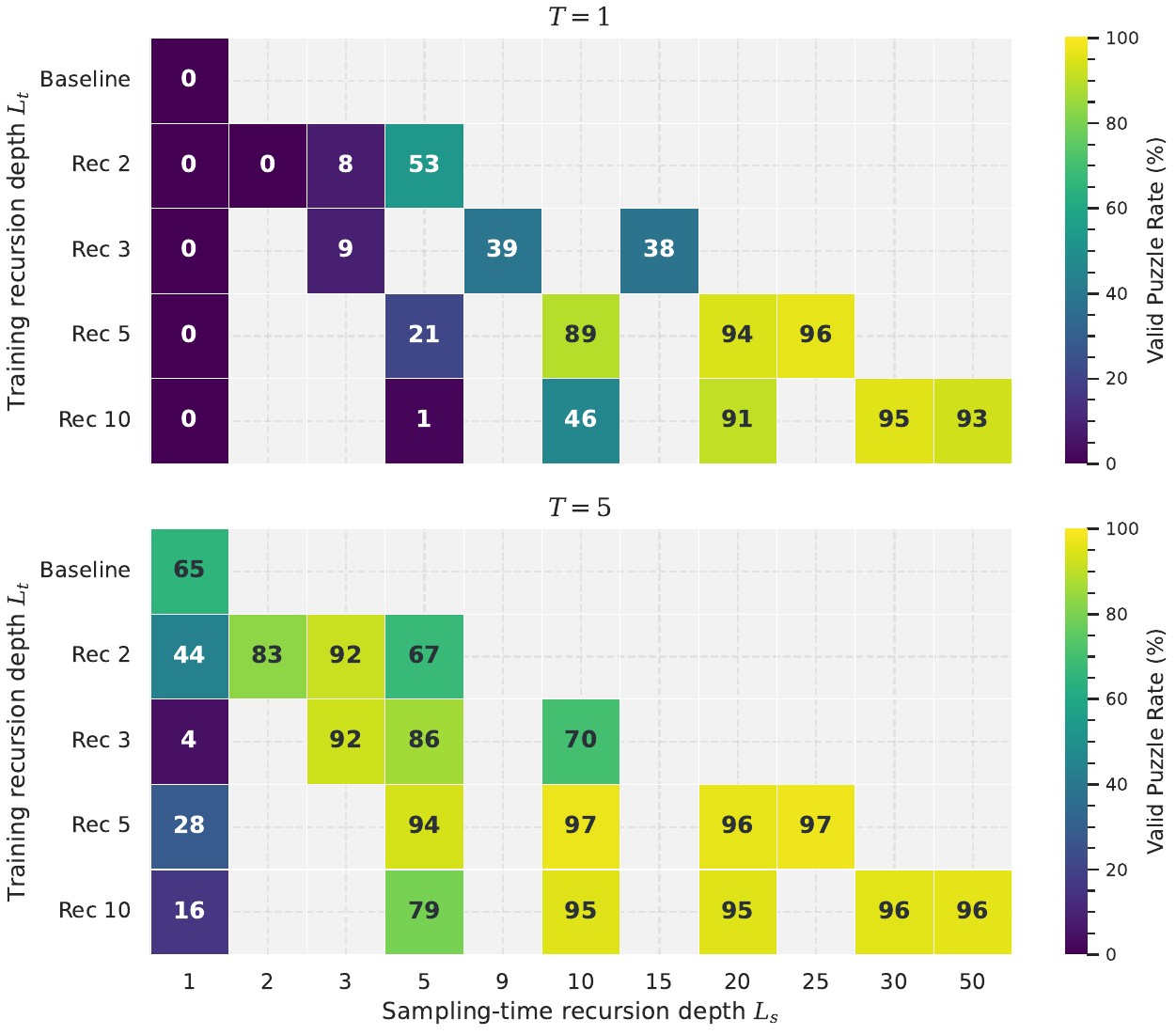}
    \caption{Cross-recursion trade-off between training recursion depth ($L_t$) and sampling recursion depth ($L_s$) for $9 \times 9$ Sudoku tasks. The top heatmap shows one-step decoding, while the bottom shows 5 decoding steps.}
    \label{fig:crossrec_heatmap}
\end{figure}

Figure~\ref{fig:crossrec_heatmap} shows the effect of varying the training loop count $L_t$ and sampling loop count $L_s$ on Sudoku $9\times9$ validity under tight decoding budgets ($T=1, 5$). Interestingly, for one-shot generation ($T=1$), increasing $L_s$ above $L_t$ consistently improves performance for all models. For $T=5$ decoding steps, moderate extrapolation still helps, but performance eventually degrades when $L_s \gg L_t$, suggesting that representations are calibrated to converge in approximately $L_t$ loops, which could also be explained by the step embedding. Critically, performance degrades sharply when $L_s \ll L_t$ (e.g.\ $L_s{=}1$ for an $L_t{=}5$ model), confirming that early-loop representations are optimized to be refined by subsequent loops rather than decoded directly. Overall, this flexibility means that recursive loops and denoising steps can be traded off continuously at inference time. In Section \ref{app:rec_steps_ablation}, we further investigate alternative schedules for the number of recursive steps using training, showing that additional performance gains can be obtained by varying the number of recursive steps during training.

Table~\ref{tab:sudoku_cross_recursion_final} extends the cross-recursion analysis by reporting both Valid Puzzle Rate and Soft Constraint Loss across training depths $L_t \in \{1,2,3,5,10\}$ and a range of sampling depths $L_s$.
 
Note that the $(L_t = 2, L_s = 3)$ setting, which uses one more loop at sampling than training, achieves 91.6\% at $T{=}5$ already surpassing the $L_t{=}2$ baseline at $T{=}5$ (83.2\%) and comparable to the $L_t{=}3$ model at $T{=}5$ (92.4\%). This suggests that moderate extrapolation is possible, likely because the step embedding (Eq.~\ref{eq:step_embed}) is parameterized by normalized progress $s_\ell \in [0,1]$ rather than the raw loop index.

\begin{table*}[ht]
    \centering
    \caption{Sudoku Performance: Cross-Recursion Trade-off. Training Recursion Depth vs. Sampling Recursion Depth across various decoding steps ($T$). We report mean and standard deviation across 5 sampling runs of 100 samples each.}
    \label{tab:sudoku_cross_recursion_final}
    \vspace{-6pt}
    \resizebox{\linewidth}{!}{
    \footnotesize
    \begin{tabular}{llcccccccccc}
    \toprule
    \textbf{Tr. Rec} & \textbf{Sa. Rec} & \multicolumn{2}{c}{$T=1$} & \multicolumn{2}{c}{$T=5$} & \multicolumn{2}{c}{$T=10$} & \multicolumn{2}{c}{$T=20$} & \multicolumn{2}{c}{$T=40$} \\
    \cmidrule(lr){3-4} \cmidrule(lr){5-6} \cmidrule(lr){7-8} \cmidrule(lr){9-10} \cmidrule(lr){11-12}
    \textbf{Steps} & \textbf{Steps} & \textbf{VPR \%} & \textbf{SCL} & \textbf{VPR \%} & \textbf{SCL} & \textbf{VPR \%} & \textbf{SCL} & \textbf{VPR \%} & \textbf{SCL} & \textbf{VPR \%} & \textbf{SCL} \\
    \midrule
    \textbf{Baseline} & 1 & 0.0 $\pm$ 0.0 & 3.077 $\pm$ 0.039 & 65.2 $\pm$ 6.9 & 0.226 $\pm$ 0.052 & 85.0 $\pm$ 2.7 & 0.088 $\pm$ 0.024 & 92.2 $\pm$ 2.7 & 0.031 $\pm$ 0.017 & 95.8 $\pm$ 1.9 & 0.011 $\pm$ 0.006 \\
    \midrule
    \multirow{4}{*}{\textbf{Rec 2}} 
     & 1 & 0.0 $\pm$ 0.0 & 3.355 $\pm$ 0.069 & 43.6 $\pm$ 3.0 & 0.415 $\pm$ 0.048 & 77.6 $\pm$ 2.5 & 0.129 $\pm$ 0.017 & 88.2 $\pm$ 3.8 & 0.052 $\pm$ 0.024 & 92.6 $\pm$ 3.2 & 0.023 $\pm$ 0.011 \\
     & 2 & 0.0 $\pm$ 0.0 & 2.443 $\pm$ 0.080 & 83.2 $\pm$ 3.6 & 0.120 $\pm$ 0.026 & 91.2 $\pm$ 1.6 & 0.055 $\pm$ 0.017 & 95.4 $\pm$ 2.8 & 0.024 $\pm$ 0.015 & 96.4 $\pm$ 1.8 & 0.016 $\pm$ 0.009 \\
     & 3 & 8.2 $\pm$ 3.1 & 1.625 $\pm$ 0.042 & 91.6 $\pm$ 2.7 & 0.060 $\pm$ 0.025 & 96.2 $\pm$ 2.2 & 0.026 $\pm$ 0.015 & 96.2 $\pm$ 1.1 & 0.021 $\pm$ 0.008 & 98.4 $\pm$ 0.9 & 0.007 $\pm$ 0.004 \\
     & 5 & 52.8 $\pm$ 5.0 & 0.658 $\pm$ 0.110 & 67.2 $\pm$ 5.6 & 0.153 $\pm$ 0.023 & 64.6 $\pm$ 3.2 & 0.160 $\pm$ 0.027 & 67.6 $\pm$ 2.7 & 0.137 $\pm$ 0.019 & 67.0 $\pm$ 6.5 & 0.130 $\pm$ 0.019 \\ 
    \midrule
    \multirow{4}{*}{\textbf{Rec 3}} 
     & 1 & 0.0 $\pm$ 0.0 & 4.485 $\pm$ 0.058 & 4.0 $\pm$ 3.9 & 0.884 $\pm$ 0.020 & 24.2 $\pm$ 3.1 & 0.457 $\pm$ 0.029 & 38.6 $\pm$ 7.1 & 0.277 $\pm$ 0.028 & 50.8 $\pm$ 9.3 & 0.165 $\pm$ 0.043 \\
     & 3 & 8.6 $\pm$ 3.0 & 1.645 $\pm$ 0.051 & 92.4 $\pm$ 2.5 & 0.056 $\pm$ 0.018 & 95.6 $\pm$ 2.1 & 0.023 $\pm$ 0.015 & 96.0 $\pm$ 2.2 & 0.024 $\pm$ 0.016 & 98.0 $\pm$ 0.7 & 0.008 $\pm$ 0.005 \\
    & 9 & 38.8 $\pm$ 5.5 & 1.304 $\pm$ 0.188 & 86.2 $\pm$ 3.0 & 0.174 $\pm$ 0.040 & 91.6 $\pm$ 4.4 & 0.073 $\pm$ 0.035 & 93.6 $\pm$ 3.4 & 0.044 $\pm$ 0.025 & 95.2 $\pm$ 1.3 & 0.033 $\pm$ 0.021 \\
    & 15 & 38.0 $\pm$ 3.7 & 1.796 $\pm$ 0.285 & 70.0 $\pm$ 6.6 & 0.817 $\pm$ 0.325 & 80.0 $\pm$ 5.4 & 0.424 $\pm$ 0.168 & 84.6 $\pm$ 1.1 & 0.260 $\pm$ 0.067 & 82.8 $\pm$ 2.7 & 0.290 $\pm$ 0.037 \\
    \midrule
    \multirow{5}{*}{\textbf{Rec 5}} 
     & 1 & 0.0 $\pm$ 0.0 & 3.645 $\pm$ 0.060 & 28.4 $\pm$ 5.8 & 0.605 $\pm$ 0.054 & 61.6 $\pm$ 4.6 & 0.219 $\pm$ 0.035 & 80.4 $\pm$ 2.5 & 0.097 $\pm$ 0.022 & 83.4 $\pm$ 3.3 & 0.061 $\pm$ 0.014 \\
     & 5 & 21.2 $\pm$ 2.4 & 1.093 $\pm$ 0.113 & 93.6 $\pm$ 2.9 & 0.050 $\pm$ 0.026 & 97.4 $\pm$ 1.3 & 0.016 $\pm$ 0.014 & 96.4 $\pm$ 1.5 & 0.016 $\pm$ 0.007 & 98.0 $\pm$ 1.6 & 0.008 $\pm$ 0.006 \\
     & 10 & 89.0 $\pm$ 2.5 & 0.127 $\pm$ 0.033 & 97.0 $\pm$ 1.4 & 0.022 $\pm$ 0.011 & 95.8 $\pm$ 2.5 & 0.025 $\pm$ 0.013 & 97.6 $\pm$ 0.9 & 0.012 $\pm$ 0.004 & 98.4 $\pm$ 0.9 & 0.008 $\pm$ 0.007 \\
     & 20 & 94.4 $\pm$ 2.9 & 0.080 $\pm$ 0.052 & 96.2 $\pm$ 1.9 & 0.027 $\pm$ 0.018 & 97.4 $\pm$ 1.3 & 0.018 $\pm$ 0.011 & 98.2 $\pm$ 0.8 & 0.011 $\pm$ 0.007 & 97.4 $\pm$ 2.4 & 0.012 $\pm$ 0.012 \\
     & 25 & 96.4 $\pm$ 2.1 & 0.050 $\pm$ 0.027 & 97.4 $\pm$ 1.1 & 0.018 $\pm$ 0.013 & 98.4 $\pm$ 1.8 & 0.009 $\pm$ 0.012 & 97.4 $\pm$ 1.7 & 0.014 $\pm$ 0.010 & 99.2 $\pm$ 0.8 & 0.002 $\pm$ 0.002 \\
    \midrule
    \multirow{6}{*}{\textbf{Rec 10}} 
    & 1 & 0.0 $\pm$ 0.0 & 3.935 $\pm$ 0.044 & 16.0 $\pm$ 3.1 & 0.686 $\pm$ 0.035 & 51.6 $\pm$ 5.0 & 0.259 $\pm$ 0.041 & 70.4 $\pm$ 3.8 & 0.132 $\pm$ 0.012 & 73.8 $\pm$ 4.6 & 0.081 $\pm$ 0.015 \\
     & 5 & 1.2 $\pm$ 0.8 & 2.186 $\pm$ 0.097 & 79.4 $\pm$ 0.9 & 0.163 $\pm$ 0.025 & 88.4 $\pm$ 2.7 & 0.069 $\pm$ 0.027 & 92.4 $\pm$ 1.5 & 0.048 $\pm$ 0.010 & 94.8 $\pm$ 1.9 & 0.026 $\pm$ 0.005 \\
     & 10 & 46.4 $\pm$ 8.0 & 0.670 $\pm$ 0.156 & 95.2 $\pm$ 2.4 & 0.034 $\pm$ 0.016 & 96.0 $\pm$ 1.6 & 0.026 $\pm$ 0.016 & 97.2 $\pm$ 1.3 & 0.017 $\pm$ 0.012 & 97.0 $\pm$ 1.9 & 0.014 $\pm$ 0.010 \\
     & 20 & 90.6 $\pm$ 2.1 & 0.120 $\pm$ 0.050 & 95.0 $\pm$ 2.1 & 0.041 $\pm$ 0.019 & 96.4 $\pm$ 3.1 & 0.026 $\pm$ 0.018 & 98.4 $\pm$ 0.9 & 0.006 $\pm$ 0.003 & 97.6 $\pm$ 1.5 & 0.013 $\pm$ 0.009 \\
     & 30 & 95.2 $\pm$ 2.5 & 0.098 $\pm$ 0.053 & 95.6 $\pm$ 1.5 & 0.026 $\pm$ 0.009 & 95.6 $\pm$ 2.0 & 0.021 $\pm$ 0.009 & 97.0 $\pm$ 1.0 & 0.018 $\pm$ 0.007 & 98.4 $\pm$ 1.1 & 0.007 $\pm$ 0.006 \\
     & 50 & 92.8 $\pm$ 2.5 & 0.122 $\pm$ 0.042 & 96.0 $\pm$ 1.0 & 0.024 $\pm$ 0.004 & 97.2 $\pm$ 1.9 & 0.018 $\pm$ 0.012 & 97.8 $\pm$ 2.3 & 0.012 $\pm$ 0.013 & 98.8 $\pm$ 0.8 & 0.005 $\pm$ 0.003 \\
    \bottomrule
    \end{tabular}
    }
\end{table*}

Having established that in our recursive paradigm the number of recursive steps can be flexibly changed at inference time, we now proceed to investigate to what extent using recursive steps (rather than denoising steps) can lead to more efficient inference. We quantify ``efficiency'' using the total number of forward passes through the model, ($x = T \times L$). Specifically, we compare a non-recursive baseline ($L_t = 1$) against a looped model trained with a reverse curriculum (\emph{Schedule $10 \rightarrow 1$}), evaluated at multiple inference depths $L_s \in \{1, 3, 5, 7, 10\}$.

 \begin{figure}
    \centering
    \includegraphics[width=0.6\linewidth]{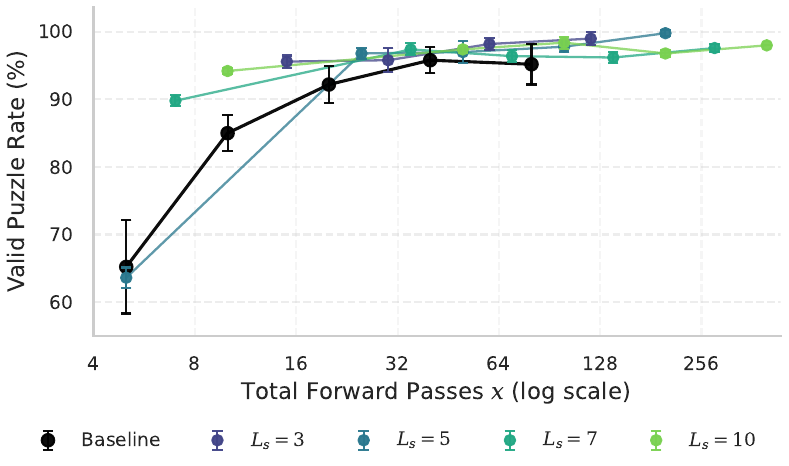}
    \caption{Valid Puzzle Rate versus total computational budget (forward passes $x = T \times L$) on Sudoku $9 \times 9$.}

    \label{fig:denoising-recursive-steps}
\end{figure}

Figure \ref{fig:denoising-recursive-steps} shows that for a fixed compute budget $x$, allocating computation to recursive inference depth $L_s$ can be more efficient than increasing decoding steps $T$ in a non-recursive model. At $x = 10$ forward passes, the baseline ($T{=}10$) achieves $85.0\%$ validity, whereas the looped model with $L_s{=}10, T{=}1$ reaches $94.2\%$, a $9.2\%$ absolute improvement under identical compute. At $x = 15$, the configuration $L_s{=}3, T{=}5$ achieves $95.6\%$, matching the baseline performance at $x = 40$ ($95.8\%$), corresponding to a $2.7\times$ reduction in compute. Beyond efficiency, recursion also raises the performance ceiling: while the baseline reaches $95.8\%$ validity at $x = 40$, recursive models exceed this level already at $T{=}5$ ($96.8\%$ for $L_s{=}5$ and $97.4\%$ for $L_s{=}7$), and reach up to $99.8\%$ ($L_s{=}5$, $T{=}40$).

\subsection{Sudoku 25$\times$25: Effect of Recursion and Masking}
\label{app:sudoku25}
 
\begin{figure*}[h]
    \centering
    \includegraphics[width=\textwidth]{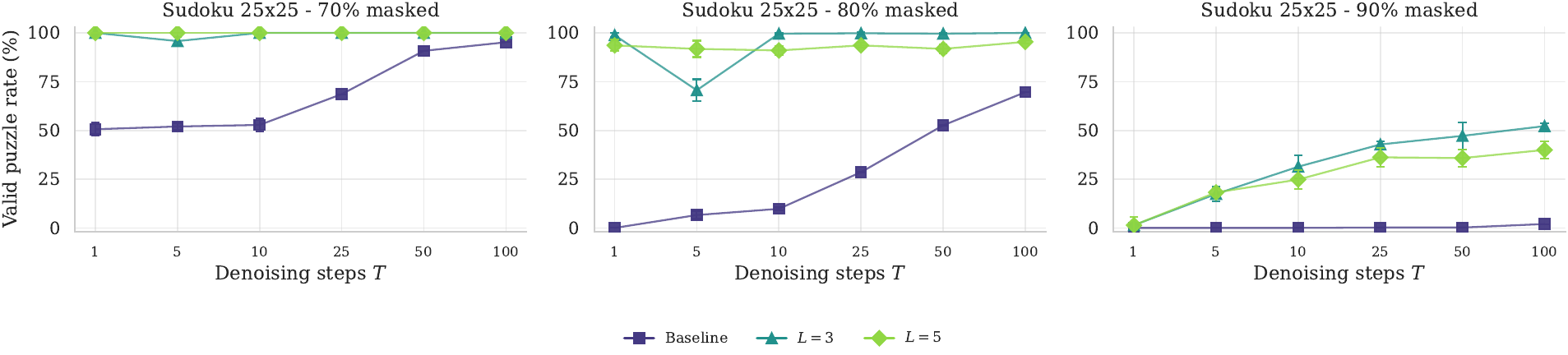}
    \caption{Effect of recursive refinement depth (fixed $L$ at training and sampling) on $25\times25$ Sudoku under different masking conditions. Across all masking regimes, recursion improves VPR.}
    \label{fig:sudoku25}
    \vspace{-0.3cm}
\end{figure*}

Table~\ref{tab:bigsudoku} and Figure~\ref{fig:sudoku25} show Valid Puzzle Rate and Soft Constraint Loss for
the baseline and recursive models on the 25$\times$25 Sudoku task under three
masking regimes: 70\%, 80\%, and 90\% of cells masked at inference time.
 
At 70\% masking, the task is tractable for all models at high $T$: the baseline reaches 95.2\% VPR at $T{=}100$. However, both $L{=}3$ and $L{=}5$ achieve 100\% at $T{=}1$, a single denoising step (i.e., $L$ passes of the $6$-layer block), whereas the baseline does not reach 100\% at any tested budget. At 80\% masking, the baseline degrades severely (69.6\% VPR at $T{=}100$), while $L{=}3$ achieves 100\% VPR at $T{=}100$ and 99.6\% at $T{=}10$. At 90\% masking, all models struggle, but the gap between baseline (2.0\% at $T{=}100$) and $L{=}3$ (52.2\%) reveals the qualitative benefit of recursive loops for highly under-determined constraint satisfaction.
 
These results suggest that the benefits of recursive depth are not simply additive with problem size but compound with masking difficulty: as fewer clues are provided, the model's need to propagate global constraints increases, and recursive loops (each a full pass through the $6$-layer bidirectional block) provide increasingly valuable additional computation.

\begin{table*}[ht]
    \caption{Sudoku 25 $\times$ 25 generative performance: Comparison between different percentages of masks (M). We report mean and standard deviation across 5 sampling runs of 100 samples each.}
    \label{tab:bigsudoku}
    \vspace{-6pt}
    \centering
    \resizebox{0.95\linewidth}{!}{
    \footnotesize
    \begin{tabular}{ll c c c c c c c c}
    \toprule
    & & \multicolumn{2}{c}{\textbf{M = 70\%}} & & \multicolumn{2}{c}{\textbf{M = 80\%}} & & \multicolumn{2}{c}{\textbf{M = 90\%}}\\
    \cmidrule(lr){3-4} \cmidrule(lr){6-7} \cmidrule(lr){9-10}
    \textbf{Model} & \textbf{Steps} & \textbf{VPR \%} & \textbf{SCL} & & \textbf{VPR \%} & \textbf{SCL} & & \textbf{VPR \%} & \textbf{SCL} \\
    \midrule
    \multirow{6}{*}{Baseline} & 1 & 50.6\% $\pm$ 3.3 & 0.189 $\pm$ 0.021 & & 0.0\% $\pm$ 0.0 & 4.587 $\pm$ 0.127 & & 0.0\% $\pm$ 0.0 & 24.978 $\pm$ 0.226  \\
     & 5 & 52.0\% $\pm$ 1.6 & 0.225 $\pm$ 0.008 & & 6.6\% $\pm$ 1.8 & 1.031 $\pm$ 0.080 & & 0.0\% $\pm$ 0.0 & 12.286 $\pm$ 0.453 \\
     & 10 & 52.8\% $\pm$ 3.4 & 0.221 $\pm$ 0.023 & & 9.8\% $\pm$ 0.8 & 0.719 $\pm$ 0.034 & & 0.0\% $\pm$ 0.0 & 6.363 $\pm$ 0.240 \\
     & 25 & 68.6\% $\pm$ 3.9 & 0.116 $\pm$ 0.026 & & 28.6\% $\pm$ 5.2 & 0.335 $\pm$ 0.032 & & 0.2\% $\pm$ 0.5 & 3.272 $\pm$ 0.188 \\
     & 50 & 90.8\% $\pm$ 2.5 & 0.026 $\pm$ 0.006 & & 52.6\% $\pm$ 4.2 & 0.142 $\pm$ 0.007 & & 0.2\% $\pm$ 0.5 & 2.060 $\pm$ 0.058 \\
     & 100 & 95.2\% $\pm$ 2.4 & 0.008 $\pm$ 0.004 & & 69.6\% $\pm$ 5.6 & 0.432 $\pm$ 0.040 & & 2.0\% $\pm$ 1.6 & 1.428 $\pm$ 0.120\\
    \midrule
    \multirow{6}{*}{3 rec. steps} & 1 & 100.0\% $\pm$ 0.0 & 0.000 $\pm$ 0.000 & & 98.8\% $\pm$ 1.1 & 0.018 $\pm$ 0.019 & & 1.4\% $\pm$ 1.3 & 10.426 $\pm$ 0.323  \\
     & 5 & 95.8\% $\pm$ 2.2 & 0.007 $\pm$ 0.004 & & 70.6\% $\pm$ 5.6 & 0.082 $\pm$ 0.029 & & 17.4\% $\pm$ 3.7 & 2.803 $\pm$ 0.243 \\
     & 10 & 100.0\% $\pm$ 0.0 & 0.000 $\pm$ 0.000 & & 99.6\% $\pm$ 0.6 & 0.001 $\pm$ 0.001 & & 31.4\% $\pm$ 5.9 & 0.256 $\pm$ 0.050 \\
     & 25 & 100.0\% $\pm$ 0.0 & 0.000 $\pm$ 0.000 & & 99.8\% $\pm$ 0.5 & 0.001 $\pm$ 0.001 & & 42.8\% $\pm$ 1.5 & 0.680 $\pm$ 0.056 \\
     & 50 & 100.0\% $\pm$ 0.0 & 0.000 $\pm$ 0.000 & & 99.6\% $\pm$ 0.6 & 0.002 $\pm$ 0.003 & & 47.2\% $\pm$ 6.8 & 0.480 $\pm$ 0.058 \\
     & 100 & 100.0\% $\pm$ 0.0 & 0.000 $\pm$ 0.000 & & 100.0\% $\pm$ 0.0 & 0.000 $\pm$ 0.000 & & 52.2\% $\pm$ 1.6 & 0.430 $\pm$ 0.051\\
    \midrule
    \multirow{6}{*}{5 rec. steps} & 1 & 100.0\% $\pm$ 0.0 & 0.000 $\pm$ 0.000 & & 93.6\% $\pm$ 2.8 & 0.083 $\pm$ 0.035 & & 1.4\% $\pm$ 1.7 & 9.298 $\pm$ 0.754  \\
     & 5 & 100.0\% $\pm$ 0.0 & 0.000 $\pm$ 0.000 & & 91.8\% $\pm$ 2.3 & 0.118 $\pm$ 0.030 & & 18.2\% $\pm$ 4.4 & 2.891 $\pm$ 0.187  \\
     & 10 &  100.0\% $\pm$ 0.0 & 0.000 $\pm$ 0.000 & & 91.0\% $\pm$ 2.5 & 0.101 $\pm$ 0.027 & & 24.8\% $\pm$ 3.5 & 1.691 $\pm$ 0.158\\
     & 25 & 100.0\% $\pm$ 0.0 & 0.000 $\pm$ 0.000 & & 93.6\% $\pm$ 1.3 & 0.039 $\pm$ 0.012 & & 36.2\% $\pm$ 1.1 & 0.936 $\pm$ 0.093\\
     & 50 & 100.0\% $\pm$ 0.0 & 0.000 $\pm$ 0.000 & & 91.8\% $\pm$ 2.6 & 0.030 $\pm$ 0.013 & & 35.8\% $\pm$ 4.8 & 0.835 $\pm$ 0.168\\
     & 100 & 100.0\% $\pm$ 0.0 & 0.000 $\pm$ 0.000 & & 95.4\% $\pm$ 5.5 & 0.017 $\pm$ 0.001 & & 40.0\% $\pm$ 3.7 & 0.600 $\pm$ 0.088\\
    \bottomrule
    \end{tabular}}
\end{table*}

\subsection{Countdown: Extended Results}
\label{app:countdown_full}
 
Table~\ref{tab:countdown_rec} reports Countdown performance for recursive models and the baseline across $k \in \{3,4,5\}$ operands, varying both recursion depth and denoising step budget. Table~\ref{tab:countdown_layers} reports the equivalent depth-scaling analysis for non-recursive models with 3, 6, 9, 15, and 30 layers.
 
For Countdown-3, the task is tractable even for the baseline at high $T$ (96.0\% RTR at $T{=}30$), but recursion substantially accelerates convergence: $L{=}3$ reaches 92.4\% at $T{=}10$, a step budget at which the baseline achieves only 59.4\%. For Countdown-4 and -5, the baseline saturates at substantially lower accuracy than the recursive models, suggesting that the difficulty of multi-step arithmetic planning grows faster than what additional denoising steps alone can resolve.

For the depth-scaling analysis (Table~\ref{tab:countdown_layers}), results show that a 15-layer non-recursive model (26.7M) on Countdown-4 reaches 80.0\% RTR at $T{=}30$, while the $(3 \otimes 5)$ model (5.5M) achieves 81.8\%, comparable performance at about one-fifth of the parameter count.

\begin{table*}[!h]
    \caption{Countdown performance with different recursive steps. We report mean and standard deviation across 5 sampling runs of 100 samples each as well as the total sampling time.}
    \label{tab:countdown_rec}
    \vspace{-6pt}
    \centering
    \resizebox{0.85\linewidth}{!}{
    \footnotesize
    \begin{tabular}{lccccccc}
    \toprule
    \textbf{Model} & \textbf{Dec. steps} & \textbf{RTR \%} & \textbf{PPF \%} & \textbf{LAF \%} & \textbf{TRN} & \textbf{Sampling time (s)} \\
    \midrule
    \multicolumn{7}{c}{\textbf{Countdown 3}} \\ \midrule
    \multirow{5}{*}{Baseline} & 1 & 7.6\% $\pm$ 1.1 & 19.0\% $\pm$ 1.2 & 26.2\% $\pm$ 2.7 & 1.046 $\pm$ 0.300 & 2.79 \\
     & 5 & 44.8\% $\pm$ 3.0 & 51.0\% $\pm$ 3.4 & 64.5\% $\pm$ 4.1 & 0.520 $\pm$ 0.027 & 8.27 \\
     & 10 & 59.4\% $\pm$ 5.0 & 63.5\% $\pm$ 4.2 & 75.8\% $\pm$ 3.9 & 0.346 $\pm$ 0.061 & 7.32 \\
     & 20 & 93.2\% $\pm$ 1.6 & 94.0\% $\pm$ 1.7 & 94.3\% $\pm$ 1.4 & 0.068 $\pm$ 0.033 & 41.51 \\
     & 30 & 96.0\% $\pm$ 0.7 & 96.3\% $\pm$ 0.5 & 96.7\% $\pm$ 0.6 & 0.037 $\pm$ 0.011 & 56.41 \\
    \midrule
    \multirow{5}{*}{3 recursive steps} & 1 & 67.0\% $\pm$ 4.6 & 70.1\% $\pm$ 5.6 & 76.9\% $\pm$ 4.6 & 0.307 $\pm$ 0.053 & 3.52 \\
     & 5  & 87.6\% $\pm$ 1.8 & 89.2\% $\pm$ 2.0 & 92.9\% $\pm$ 1.9 & 0.145 $\pm$ 0.086 & 14.43 \\
     & 10 & 92.4\% $\pm$ 2.5 & 93.2\% $\pm$ 2.7 & 95.5\% $\pm$ 2.0 & 0.061 $\pm$ 0.017 & 28.24 \\
     & 20 & 97.4\% $\pm$ 1.5 & 97.6\% $\pm$ 1.6 & 98.0\% $\pm$ 1.2 & 0.021 $\pm$ 0.012 & 52.85 \\
     & 30 & 97.8\% $\pm$ 1.6 & 98.2\% $\pm$ 1.5 & 98.3\% $\pm$ 1.3 & 0.030 $\pm$ 0.024 & 55.96 \\
     \midrule
    \multirow{5}{*}{5 recursive steps} & 1 & 67.4\% $\pm$ 2.1 & 70.3\% $\pm$ 2.0 & 79.5\% $\pm$ 1.5 & 0.317 $\pm$ 0.088 & 5.57 \\
     & 5  & 85.8\% $\pm$ 6.8 & 86.6\% $\pm$ 6.0 & 93.0\% $\pm$ 3.7 & 0.122 $\pm$ 0.059 & 23.32 \\
     & 10 & 91.0\% $\pm$ 1.6 & 91.5\% $\pm$ 1.5 & 95.4\% $\pm$ 0.9 & 0.071 $\pm$ 0.015 & 46.97 \\
     & 20 & 98.4\% $\pm$ 0.9 & 98.7\% $\pm$ 0.8 & 98.7\% $\pm$ 0.8 & 0.012 $\pm$ 0.011 & 86.52 \\
     & 30 & 98.2\% $\pm$ 1.9 & 98.5\% $\pm$ 1.5 & 98.8\% $\pm$ 0.9 & 0.016 $\pm$ 0.016 & 91.50 \\
     \midrule
    \multicolumn{7}{c}{\textbf{Countdown 4}} \\ \midrule
    \multirow{5}{*}{Baseline} & 1 & 0.0\% $\pm$ 0.0 & 1.9\% $\pm$ 0.6 & 4.1\% $\pm$ 1.1 & 1.054 $\pm$ 0.169 & 2.55 \\
     & 5 & 6.6\% $\pm$ 3.1 & 16.7\% $\pm$ 3.2 & 36.2\% $\pm$ 1.2 & 1.001 $\pm$ 0.231 & 5.67 \\
     & 10 & 18.0\% $\pm$ 2.5 & 31.6\% $\pm$ 3.5 & 49.3\% $\pm$ 3.6 & 0.817 $\pm$ 0.140 & 10.82 \\
     & 20 & 38.2\% $\pm$ 6.2 & 52.3\% $\pm$ 4.4 & 64.5\% $\pm$ 3.8 & 0.648 $\pm$ 0.206 & 21.36 \\
     & 30 & 44.8\% $\pm$ 6.0 & 59.5\% $\pm$ 5.2 & 71.5\% $\pm$ 3.3 & 0.626 $\pm$ 0.046 & 31.88 \\
    \midrule
    \multirow{5}{*}{3 recursive steps} & 1 & 0.8\% $\pm$ 0.8 & 5.5\% $\pm$ 1.2 & 15.0\% $\pm$ 1.7 & 1.243 $\pm$ 0.401 & 4.51 \\
     & 5  & 11.6\% $\pm$ 2.4 & 28.4\% $\pm$ 2.3 & 53.2\% $\pm$ 2.2 & 0.908 $\pm$ 0.180 & 16.52 \\
     & 10 & 39.6\% $\pm$ 1.8 & 56.9\% $\pm$ 1.7 & 72.6\% $\pm$ 1.0 & 0.593 $\pm$ 0.143 & 32.83 \\
     & 20 & 68.6\% $\pm$ 6.2 & 77.6\% $\pm$ 4.0 & 85.1\% $\pm$ 3.0 & 0.364 $\pm$ 0.150 & 63.77 \\
     & 30 & 76.8\% $\pm$ 2.6 & 84.4\% $\pm$ 1.6 & 89.6\% $\pm$ 1.0 & 0.256 $\pm$ 0.131 & 83.95 \\
     \midrule
    \multirow{5}{*}{5 recursive steps} & 1 & 0.6\% $\pm$ 0.6 & 6.7\% $\pm$ 2.6 & 17.2\% $\pm$ 1.9 & 1.021 $\pm$ 0.145 & 7.46 \\
     & 5  & 15.4\% $\pm$ 3.0 & 30.9\% $\pm$ 2.0 & 54.2\% $\pm$ 1.6 & 0.802 $\pm$ 0.075 & 28.37 \\
     & 10 & 37.0\% $\pm$ 2.9 & 53.7\% $\pm$ 3.1 & 69.9\% $\pm$ 2.0 & 0.580 $\pm$ 0.088 & 50.52 \\
     & 20 & 70.2\% $\pm$ 7.0 & 78.3\% $\pm$ 5.2 & 85.6\% $\pm$ 3.3 & 0.269 $\pm$ 0.055 & 93.76 \\
     & 30 & 81.8\% $\pm$ 2.5 & 86.9\% $\pm$ 1.0 & 90.4\% $\pm$ 1.1 & 0.212 $\pm$ 0.116 & 137.65 \\
     \midrule
     \multirow{5}{*}{10 recursive steps} & 1 & 0.8\% $\pm$ 0.8 & 5.6\% $\pm$ 2.4 & 14.8\% $\pm$ 3.2 & 0.876 $\pm$ 0.043 & 10.50 \\
     & 5 & 12.8\% $\pm$ 2.2 & 27.6\% $\pm$ 1.2 & 48.1\% $\pm$ 1.9 & 0.848 $\pm$ 0.118 & 46.15 \\
     & 10 & 43.6\% $\pm$ 4.2 & 56.7\% $\pm$ 4.1 & 70.2\% $\pm$ 3.5 & 0.676 $\pm$ 0.317 & 91.65 \\
     & 20 & 74.4\% $\pm$ 5.3 & 81.1\% $\pm$ 4.6 & 86.1\% $\pm$ 2.8 & 0.261 $\pm$ 0.102 & 179.72 \\
     & 30 & 86.2\% $\pm$ 4.1 & 89.7\% $\pm$ 3.0 & 91.7\% $\pm$ 2.5 & 0.211 $\pm$ 0.184 & 264.44 \\
     \midrule
    \multicolumn{7}{c}{\textbf{Countdown 5}} \\ \midrule
    \multirow{6}{*}{Baseline} & 1 & 0.0\% $\pm$ 0.0 & 0.3\% $\pm$ 0.3 & 1.3\% $\pm$ 0.6 & 0.925 $\pm$ 0.112 & 2.75 \\
     & 5 & 0.2\% $\pm$ 0.5 & 11.1\% $\pm$ 1.3 & 33.7\% $\pm$ 0.9 & 1.023 $\pm$ 0.192 & 5.68 \\
     & 10 & 3.8\% $\pm$ 2.4 & 20.9\% $\pm$ 2.8 & 47.0\% $\pm$ 2.9 & 0.909 $\pm$ 0.069 & 11.23 \\
     & 20 & 9.6\% $\pm$ 1.8 & 28.3\% $\pm$ 2.3 & 57.4\% $\pm$ 1.7 & 0.886 $\pm$ 0.161 & 21.74 \\
     & 30 & 15.6\% $\pm$ 5.1 & 37.6\% $\pm$ 4.1 & 63.3\% $\pm$ 3.0 & 0.865 $\pm$ 0.229 & 32.22 \\
     & 40 & 17.0\% $\pm$ 2.6 & 40.5\% $\pm$ 4.4 & 65.1\% $\pm$ 3.4 & 0.889 $\pm$ 0.168 & 166.18 \\
    \midrule
    \multirow{6}{*}{3 recursive steps} & 1 & 0.0\% $\pm$ 0.0 & 0.5\% $\pm$ 0.4 & 1.6\% $\pm$ 0.6 & 1.036 $\pm$ 0.229 & 3.50 \\
     & 5  & 4.8\% $\pm$ 1.8 & 20.9\% $\pm$ 1.2 & 47.6\% $\pm$ 1.7 & 0.971 $\pm$ 0.169 & 14.00 \\
     & 10 & 10.8\% $\pm$ 2.2 & 31.0\% $\pm$ 1.6 & 60.9\% $\pm$ 1.9 & 0.758 $\pm$ 0.068 & 28.00 \\
     & 20 & 21.6\% $\pm$ 4.2 & 42.6\% $\pm$ 5.5 & 70.9\% $\pm$ 2.5 & 0.913 $\pm$ 0.153 & 55.50 \\
     & 30 & 36.6\% $\pm$ 4.9 & 57.5\% $\pm$ 2.2 & 79.6\% $\pm$ 1.0 & 0.746 $\pm$ 0.160 & 83.00 \\
     & 40 & 39.6\% $\pm$ 7.0 & 60.3\% $\pm$ 6.0 & 80.8\% $\pm$ 3.7 & 0.782 $\pm$ 0.125 & 107.00 \\
     \midrule
    \multirow{6}{*}{5 recursive steps} & 1 & 0.0\% $\pm$ 0.0 & 0.7\% $\pm$ 0.4 & 2.6\% $\pm$ 0.6 & 1.136 $\pm$ 0.123 & 6.42 \\
     & 5 & 4.8\% $\pm$ 1.9 & 23.6\% $\pm$ 1.9 & 47.0\% $\pm$ 4.5 & 1.051 $\pm$ 0.275 & 24.82 \\
     & 10 & 14.2\% $\pm$ 3.4 & 34.0\% $\pm$ 2.1 & 59.6\% $\pm$ 2.1 & 0.864 $\pm$ 0.177 & 48.76 \\
     & 20 & 24.6\% $\pm$ 7.2 & 44.6\% $\pm$ 6.4 & 69.7\% $\pm$ 3.8 & 0.822 $\pm$ 0.303 & 97.16 \\
     & 30 & 44.4\% $\pm$ 4.7 & 62.1\% $\pm$ 4.8 & 80.4\% $\pm$ 2.1 & 0.585 $\pm$ 0.122 & 141.88 \\
     & 40 & 46.8\% $\pm$ 6.5 & 64.3\% $\pm$ 4.3 & 81.3\% $\pm$ 1.3 & 0.586 $\pm$ 0.270 & 174.04 \\
     \midrule
    \multirow{6}{*}{10 recursive steps} & 1 & 0.2\% $\pm$ 0.5 & 1.2\% $\pm$ 1.0 & 3.5\% $\pm$ 0.9 & 1.107 $\pm$ 0.330 & 10.50 \\
 & 5 & 5.8\% $\pm$ 1.3 & 22.4\% $\pm$ 1.9 & 44.2\% $\pm$ 2.8 & 0.856 $\pm$ 0.204 & 43.51 \\
 & 10 & 14.4\% $\pm$ 4.3 & 35.1\% $\pm$ 4.1 & 58.2\% $\pm$ 4.7 & 0.880 $\pm$ 0.166 & 86.57 \\
 & 20 & 28.4\% $\pm$ 4.4 & 49.3\% $\pm$ 4.2 & 69.4\% $\pm$ 2.5 & 0.654 $\pm$ 0.033 & 173.24 \\
 & 30 & 56.4\% $\pm$ 5.0 & 71.4\% $\pm$ 4.2 & 82.0\% $\pm$ 2.0 & 0.671 $\pm$ 0.355 & 260.25 \\
 & 40 & 53.8\% $\pm$ 4.3 & 69.6\% $\pm$ 3.4 & 80.4\% $\pm$ 2.0 & 0.538 $\pm$ 0.208 & 330.43 \\
    \bottomrule
    \end{tabular}}
\end{table*}

\begin{table*}[h!]
    \caption{Countdown performance with models of different depth. We report mean and standard deviation across 5 sampling runs of 100 samples each as well as the total sampling time.}
    \label{tab:countdown_layers}
    \vspace{-6pt}
    \centering
    \resizebox{0.78\linewidth}{!}{
    \footnotesize
    \begin{tabular}{lccccccc}
    \toprule
    \textbf{Model (params)} & \textbf{Dec. steps} & \textbf{RTR \%} & \textbf{PPF \%} & \textbf{LAF \%} & \textbf{TRN} & \textbf{Sampling time (s)} \\
    \midrule
    \multicolumn{7}{c}{\textbf{Countdown 3}} \\ \midrule
    \multirow{5}{*}{3 layers (5.5M)}  & 1 & 7.6\% $\pm$ 1.1 & 19.0\% $\pm$ 1.2 & 26.2\% $\pm$ 2.7 & 1.046 $\pm$ 0.300 & 2.79 \\
     & 5 & 44.8\% $\pm$ 3.0 & 51.0\% $\pm$ 3.4 & 64.5\% $\pm$ 4.1 & 0.520 $\pm$ 0.027 & 8.27 \\
     & 10 & 59.4\% $\pm$ 5.0 & 63.5\% $\pm$ 4.2 & 75.8\% $\pm$ 3.9 & 0.346 $\pm$ 0.061 & 7.32 \\
     & 20 & 93.2\% $\pm$ 1.6 & 94.0\% $\pm$ 1.7 & 94.3\% $\pm$ 1.4 & 0.068 $\pm$ 0.033 & 41.51 \\
     & 30 & 96.0\% $\pm$ 0.7 & 96.3\% $\pm$ 0.5 & 96.7\% $\pm$ 0.6 & 0.037 $\pm$ 0.011 & 56.41 \\
    \midrule
    \multirow{5}{*}{6 layers (10.8M)} & 1 & 20.0\% $\pm$ 4.1 & 28.5\% $\pm$ 3.5 & 39.9\% $\pm$ 4.8 & 0.865 $\pm$ 0.201 & 2.52 \\
     & 5 & 56.8\% $\pm$ 6.5 & 59.9\% $\pm$ 6.4 & 73.2\% $\pm$ 4.9 & 0.413 $\pm$ 0.093 & 9.36 \\
     & 10 & 74.4\% $\pm$ 5.8 & 76.4\% $\pm$ 5.9 & 85.4\% $\pm$ 4.0 & 0.341 $\pm$ 0.186 & 18.48 \\
     & 20 & 97.2\% $\pm$ 1.8 & 97.4\% $\pm$ 1.5 & 97.6\% $\pm$ 1.4 & 0.025 $\pm$ 0.014 & 34.41 \\
     & 30 & 97.0\% $\pm$ 1.2 & 97.3\% $\pm$ 1.2 & 97.8\% $\pm$ 1.2 & 0.028 $\pm$ 0.013 & 35.78 \\
    \midrule
    \multirow{5}{*}{9 layers (16.1M)} & 1 & 36.0\% $\pm$ 4.6 & 43.1\% $\pm$ 3.4 & 54.8\% $\pm$ 2.6 & 0.558 $\pm$ 0.043 & 4.89 \\
     & 5 & 72.4\% $\pm$ 3.6 & 75.0\% $\pm$ 3.5 & 83.2\% $\pm$ 2.6 & 0.389 $\pm$ 0.261 & 13.17 \\
     & 10 & 86.6\% $\pm$ 2.7 & 87.6\% $\pm$ 2.1 & 92.1\% $\pm$ 2.0 & 0.121 $\pm$ 0.013 & 26.12 \\
     & 20 & 96.0\% $\pm$ 1.9 & 96.5\% $\pm$ 1.7 & 97.2\% $\pm$ 1.8 & 0.036 $\pm$ 0.015 & 48.75 \\
     & 30 & 95.0\% $\pm$ 1.9 & 95.7\% $\pm$ 1.4 & 96.2\% $\pm$ 1.4 & 0.131 $\pm$ 0.208 & 51.51 \\
     \midrule
    \multirow{5}{*}{15 layers (26.7M)} & 1 & 59.0\% $\pm$ 3.9 & 64.3\% $\pm$ 4.0 & 73.4\% $\pm$ 3.0 & 0.405 $\pm$ 0.090 & 5.67 \\
     & 5 & 88.0\% $\pm$ 1.4 & 90.0\% $\pm$ 0.7 & 91.6\% $\pm$ 0.8 & 0.119 $\pm$ 0.049 & 20.54 \\
     & 10 & 94.4\% $\pm$ 4.2 & 95.1\% $\pm$ 3.6 & 96.1\% $\pm$ 2.8 & 0.052 $\pm$ 0.037 & 40.93 \\
     & 20 & 95.6\% $\pm$ 2.2 & 96.4\% $\pm$ 2.0 & 96.8\% $\pm$ 1.6 & 0.036 $\pm$ 0.025 & 76.25 \\
     & 30 & 95.2\% $\pm$ 1.3 & 96.0\% $\pm$ 1.2 & 96.6\% $\pm$ 1.1 & 0.038 $\pm$ 0.013 & 83.87 \\
    \midrule
    \multicolumn{7}{c}{\textbf{Countdown 4}} \\ \midrule
    \multirow{5}{*}{3 layers (5.5M)} & 1 & 0.0\% $\pm$ 0.0 & 1.9\% $\pm$ 0.6 & 4.1\% $\pm$ 1.1 & 1.054 $\pm$ 0.169 & 2.55 \\
     & 5 & 6.6\% $\pm$ 3.1 & 16.7\% $\pm$ 3.2 & 36.2\% $\pm$ 1.2 & 1.001 $\pm$ 0.231 & 5.67 \\
     & 10 & 18.0\% $\pm$ 2.5 & 31.6\% $\pm$ 3.5 & 49.3\% $\pm$ 3.6 & 0.817 $\pm$ 0.140 & 10.82 \\
     & 20 & 38.2\% $\pm$ 6.2 & 52.3\% $\pm$ 4.4 & 64.5\% $\pm$ 3.8 & 0.648 $\pm$ 0.206 & 21.36 \\
     & 30 & 44.8\% $\pm$ 6.0 & 59.5\% $\pm$ 5.2 & 71.5\% $\pm$ 3.3 & 0.626 $\pm$ 0.046 & 31.88 \\
    \midrule
    \multirow{5}{*}{6 layers (10.8M)} & 1 & 0.4\% $\pm$ 0.6 & 3.8\% $\pm$ 1.0 & 10.7\% $\pm$ 1.2 & 1.003 $\pm$ 0.180 &  3.33 \\
     & 5 & 10.2\% $\pm$ 3.6 & 24.0\% $\pm$ 4.8 & 47.9\% $\pm$ 1.2 & 0.915 $\pm$ 0.129 & 13.44 \\
     & 10 & 34.8\% $\pm$ 4.3 & 48.9\% $\pm$ 5.0 & 65.1\% $\pm$ 3.2 & 0.755 $\pm$ 0.269 & 26.62 \\
     & 20 & 59.8\% $\pm$ 3.1 & 71.7\% $\pm$ 3.7 & 79.2\% $\pm$ 3.3 & 0.385 $\pm$ 0.058 &  52.90\\
     & 30 & 68.4\% $\pm$ 4.0 & 77.6\% $\pm$ 2.3 & 83.1\% $\pm$ 1.3 & 0.445 $\pm$ 0.255 &  76.06\\
    \midrule
    \multirow{5}{*}{9 layers (16.1M)} & 1 & 0.2\% $\pm$ 0.5 & 5.7\% $\pm$ 1.7 & 14.1\% $\pm$ 2.1 & 1.023 $\pm$ 0.108 & 5.19\\
     & 5 & 15.0\% $\pm$ 3.5 & 28.0\% $\pm$ 2.8 & 51.3\% $\pm$ 2.8 & 0.837 $\pm$ 0.140 & 20.07\\
     & 10 & 38.8\% $\pm$ 2.7 & 51.9\% $\pm$ 1.5 & 68.2\% $\pm$ 1.9 & 0.591 $\pm$ 0.071 & 42.87 \\
     & 20 & 64.2\% $\pm$ 4.4 & 73.0\% $\pm$ 3.5 & 80.0\% $\pm$ 2.1 & 0.359 $\pm$ 0.091 & 84.29 \\
     & 30 & 74.2\% $\pm$ 8.0 & 81.3\% $\pm$ 5.7 & 86.1\% $\pm$ 4.3 & 0.227 $\pm$ 0.086 & 106.25 \\
    \midrule
    \multirow{5}{*}{15 layers (26.7M)} & 1 & 1.4\% $\pm$ 1.1 & 9.1\% $\pm$ 2.1 & 24.2\% $\pm$ 2.2 & 0.999 $\pm$ 0.303 & 5.65 \\
     & 5 & 21.4\% $\pm$ 4.9 & 33.9\% $\pm$ 3.1 & 60.1\% $\pm$ 3.6 & 0.781 $\pm$ 0.124 & 22.96 \\
     & 10 & 48.2\% $\pm$ 2.6 & 59.6\% $\pm$ 2.2 & 75.5\% $\pm$ 1.8 & 0.453 $\pm$ 0.046 & 41.23 \\
     & 20 & 69.4\% $\pm$ 4.3 & 78.3\% $\pm$ 2.6 & 84.7\% $\pm$ 1.8 & 0.269 $\pm$ 0.066 & 81.46 \\
     & 30 & 80.0\% $\pm$ 2.1 & 85.7\% $\pm$ 1.9 & 89.5\% $\pm$ 2.3 & 0.183 $\pm$ 0.042 & 121.79 \\
    \midrule
    \multirow{5}{*}{30 layers (53.3M)} & 1 & 5.2\% $\pm$ 2.3 & 15.2\% $\pm$ 2.1 & 32.4\% $\pm$ 3.5 & 0.886 $\pm$ 0.075 & 8.69 \\
     & 5 & 32.4\% $\pm$ 4.9 & 42.6\% $\pm$ 3.7 & 67.2\% $\pm$ 3.7 & 0.616 $\pm$ 0.047 & 40.17 \\
     & 10 & 58.0\% $\pm$ 2.7 & 68.7\% $\pm$ 2.8 & 80.4\% $\pm$ 1.3 & 0.356 $\pm$ 0.051 & 79.95 \\
     & 20 & 69.6\% $\pm$ 2.6 & 78.2\% $\pm$ 1.8 & 85.4\% $\pm$ 2.0 & 0.362 $\pm$ 0.152 & 159.88 \\
     & 30 & 72.6\% $\pm$ 2.4 & 81.3\% $\pm$ 2.0 & 86.3\% $\pm$ 1.6 & 0.300 $\pm$ 0.144 & 231.88 \\
    \midrule
    \multicolumn{7}{c}{\textbf{Countdown 5}} \\ \midrule
    \multirow{6}{*}{3 layers (5.5M)} & 1 & 0.0\% $\pm$ 0.0 & 0.3\% $\pm$ 0.3 & 1.3\% $\pm$ 0.6 & 0.925 $\pm$ 0.112 & 2.75 \\
     & 5 & 0.2\% $\pm$ 0.5 & 11.1\% $\pm$ 1.3 & 33.7\% $\pm$ 0.9 & 1.023 $\pm$ 0.192 & 5.68 \\
     & 10 & 3.8\% $\pm$ 2.4 & 20.9\% $\pm$ 2.8 & 47.0\% $\pm$ 2.9 & 0.909 $\pm$ 0.069 & 11.23 \\
     & 20 & 9.6\% $\pm$ 1.8 & 28.3\% $\pm$ 2.3 & 57.4\% $\pm$ 1.7 & 0.886 $\pm$ 0.161 & 21.74 \\
     & 30 & 15.6\% $\pm$ 5.1 & 37.6\% $\pm$ 4.1 & 63.3\% $\pm$ 3.0 & 0.865 $\pm$ 0.229 & 32.22 \\
     & 40 & 17.0\% $\pm$ 2.6 & 40.5\% $\pm$ 4.4 & 65.1\% $\pm$ 3.4 & 0.889 $\pm$ 0.168 & 166.18 \\
    \midrule
    \multirow{5}{*}{6 layers (10.8M)} & 1 & 0.0\% $\pm$ 0.0 & 0.9\% $\pm$ 0.4 & 2.1\% $\pm$ 1.1 & 0.976 $\pm$ 0.162 & 2.51 \\
     & 5 & 1.8\% $\pm$ 0.8 & 17.5\% $\pm$ 2.3 & 45.7\% $\pm$ 3.3 & 0.923 $\pm$ 0.150 & 9.51 \\
     & 10 & 8.6\% $\pm$ 0.6 & 27.9\% $\pm$ 1.3 & 54.4\% $\pm$ 1.2 & 1.047 $\pm$ 0.397 & 18.40 \\
     & 20 & 15.4\% $\pm$ 2.9 & 36.9\% $\pm$ 5.4 & 64.3\% $\pm$ 2.6 & 1.017 $\pm$ 0.176 & 36.99 \\
     & 30 & 28.4\% $\pm$ 3.8 & 51.2\% $\pm$ 4.0 & 71.9\% $\pm$ 2.8 & 0.708 $\pm$ 0.121 & 54.75 \\
    & 40 & 29.4\% $\pm$ 3.8 & 53.6\% $\pm$ 3.1 & 73.3\% $\pm$ 1.6 & 0.845 $\pm$ 0.279 & 72.69 \\
    \midrule
    \multirow{6}{*}{9 layers (16.1M)} & 1 & 0.0\% $\pm$ 0.0 & 1.3\% $\pm$ 0.4 & 2.7\% $\pm$ 0.9 & 1.128 $\pm$ 0.232 & 4.01 \\
     & 5 & 2.8\% $\pm$ 1.3 & 18.5\% $\pm$ 2.2 & 43.4\% $\pm$ 1.3 & 0.953 $\pm$ 0.201 & 13.15 \\
     & 10 & 9.4\% $\pm$ 3.2 & 28.6\% $\pm$ 3.4 & 56.0\% $\pm$ 3.3 & 0.992 $\pm$ 0.322 & 25.85 \\
     & 20 & 22.0\% $\pm$ 5.3 & 40.6\% $\pm$ 4.3 & 65.0\% $\pm$ 3.5 & 0.718 $\pm$ 0.060 & 56.03 \\
     & 30 & 37.8\% $\pm$ 3.3 & 57.4\% $\pm$ 2.6 & 75.0\% $\pm$ 2.3 & 0.714 $\pm$ 0.226 & 87.30 \\
     & 40 & 39.6\% $\pm$ 2.3 & 60.7\% $\pm$ 1.6 & 77.4\% $\pm$ 1.5 & 0.604 $\pm$ 0.090 & 99.92 \\
    \midrule
    \multirow{6}{*}{15 layers (26.7M)} & 1 & 0.0\% $\pm$ 0.0 & 1.2\% $\pm$ 0.6 & 3.0\% $\pm$ 0.7 & 0.977 $\pm$ 0.170 & 5.81 \\
     & 5 & 4.0\% $\pm$ 1.7 & 20.3\% $\pm$ 3.4 & 44.4\% $\pm$ 3.2 & 0.909 $\pm$ 0.076 & 23.30 \\
     & 10 & 11.2\% $\pm$ 2.4 & 31.7\% $\pm$ 2.3 & 60.8\% $\pm$ 0.9 & 1.091 $\pm$ 0.381 & 43.96 \\
     & 20 & 21.8\% $\pm$ 5.0 & 44.6\% $\pm$ 3.7 & 69.7\% $\pm$ 2.2 & 0.764 $\pm$ 0.109 & 83.75 \\
     & 30 & 42.2\% $\pm$ 4.6 & 63.5\% $\pm$ 2.9 & 81.2\% $\pm$ 1.6 & 0.587 $\pm$ 0.079 & 125.12 \\
     & 40 & 45.8\% $\pm$ 3.3 & 64.9\% $\pm$ 2.0 & 82.2\% $\pm$ 2.0 & 0.530 $\pm$ 0.073 & 161.05 \\
    \midrule
    \multirow{6}{*}{30 layers (53.3M)} & 1 & 0.0\% $\pm$ 0.0 & 1.7\% $\pm$ 0.3 & 3.6\% $\pm$ 0.8 & 0.912 $\pm$ 0.047 & 8.57 \\
     & 5 & 5.0\% $\pm$ 1.6 & 21.5\% $\pm$ 1.7 & 47.1\% $\pm$ 3.3 & 0.859 $\pm$ 0.117 & 39.80 \\
     & 10 & 15.0\% $\pm$ 3.9 & 34.8\% $\pm$ 3.3 & 60.3\% $\pm$ 2.5 & 0.927 $\pm$ 0.347 & 79.44 \\
     & 20 & 28.2\% $\pm$ 3.9 & 47.5\% $\pm$ 3.1 & 70.0\% $\pm$ 1.4 & 0.698 $\pm$ 0.165 & 158.83 \\
     & 30 & 45.4\% $\pm$ 5.2 & 65.2\% $\pm$ 4.2 & 79.6\% $\pm$ 2.6 & 0.479 $\pm$ 0.048 & 237.60 \\
     & 40 & 47.8\% $\pm$ 6.0 & 66.6\% $\pm$ 3.4 & 80.6\% $\pm$ 3.2 & 0.494 $\pm$ 0.149 & 304.52 \\
    \bottomrule
    \end{tabular}}
\end{table*}

\clearpage

\subsection{Decoding algorithms comparison}
\label{app:decoding_comparison}

\begin{figure*}[ht]
    \centering
    \begin{subfigure}[t]{0.8\textwidth}
        \centering
        \includegraphics[width=1\textwidth]{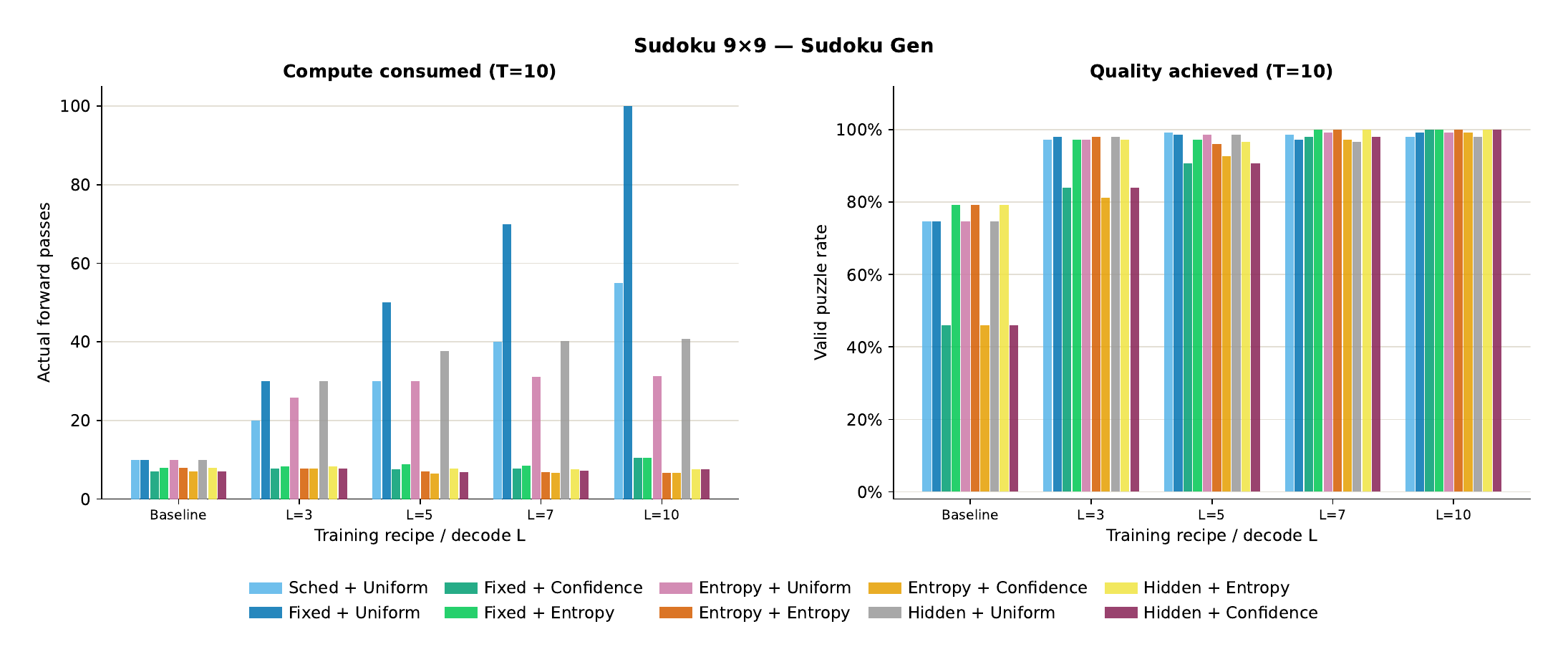}
        \caption{Sudoku $9{\times}9$ (VPR), $T{=}10$.}
        \label{fig:decoding_bars_sudoku}
    \end{subfigure}

    \begin{subfigure}[t]{0.8\textwidth}
        \centering
        \includegraphics[width=1\textwidth]{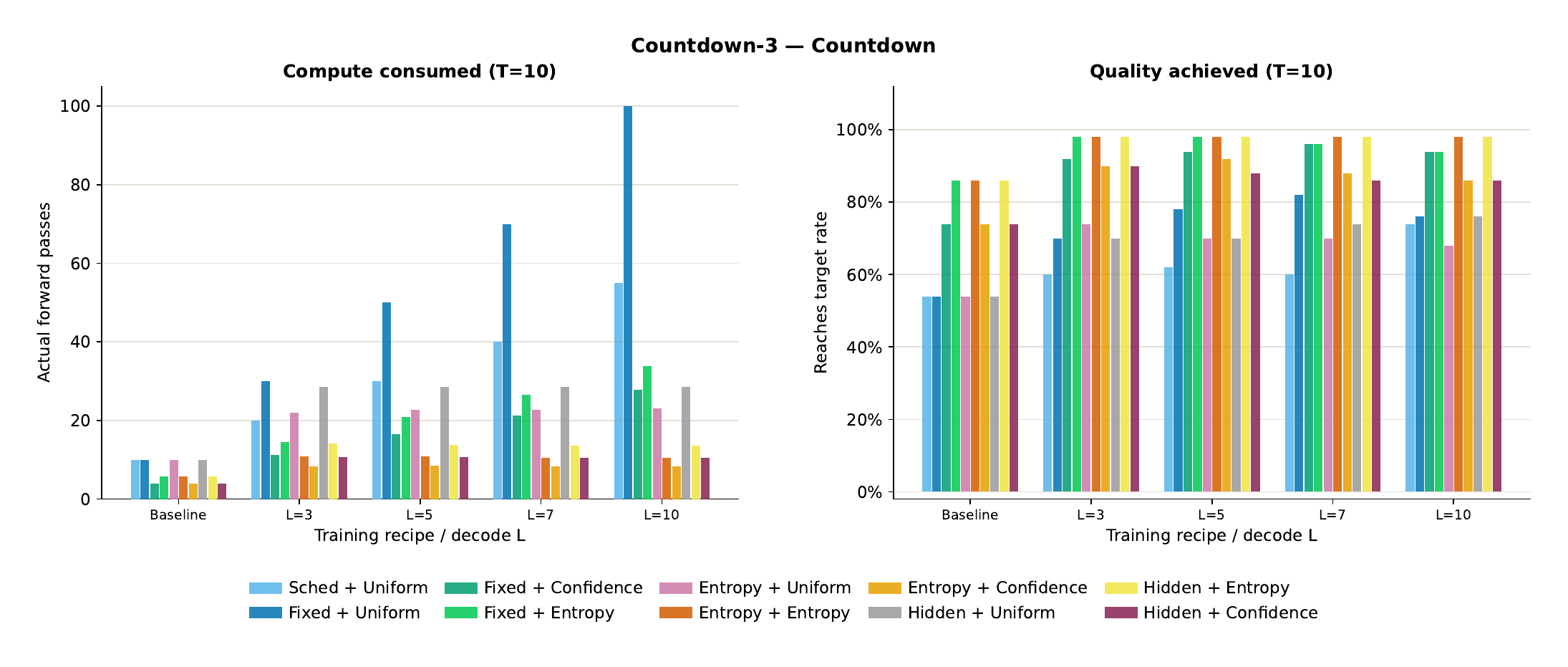}
        \caption{Countdown-3 (RTR), $T{=}10$.}
        \label{fig:decoding_bars_count3}
    \end{subfigure}

    \begin{subfigure}[t]{0.8\textwidth}
        \centering
        \includegraphics[width=1\textwidth]{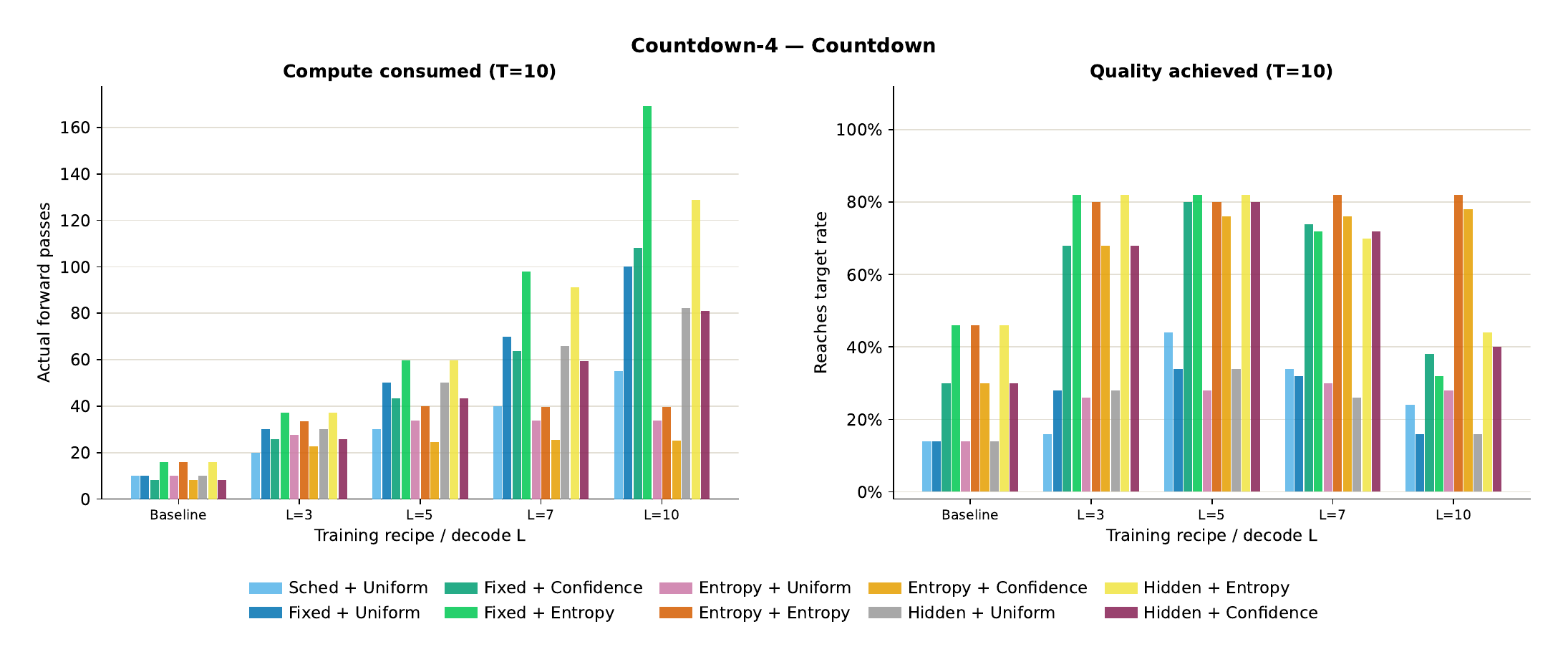}
        \caption{Countdown-4 (RTR), $T{=}10$.}
        \label{fig:decoding_bars_count4}
    \end{subfigure}

    \begin{subfigure}[t]{0.8\textwidth}
        \centering
        \includegraphics[width=1\textwidth]{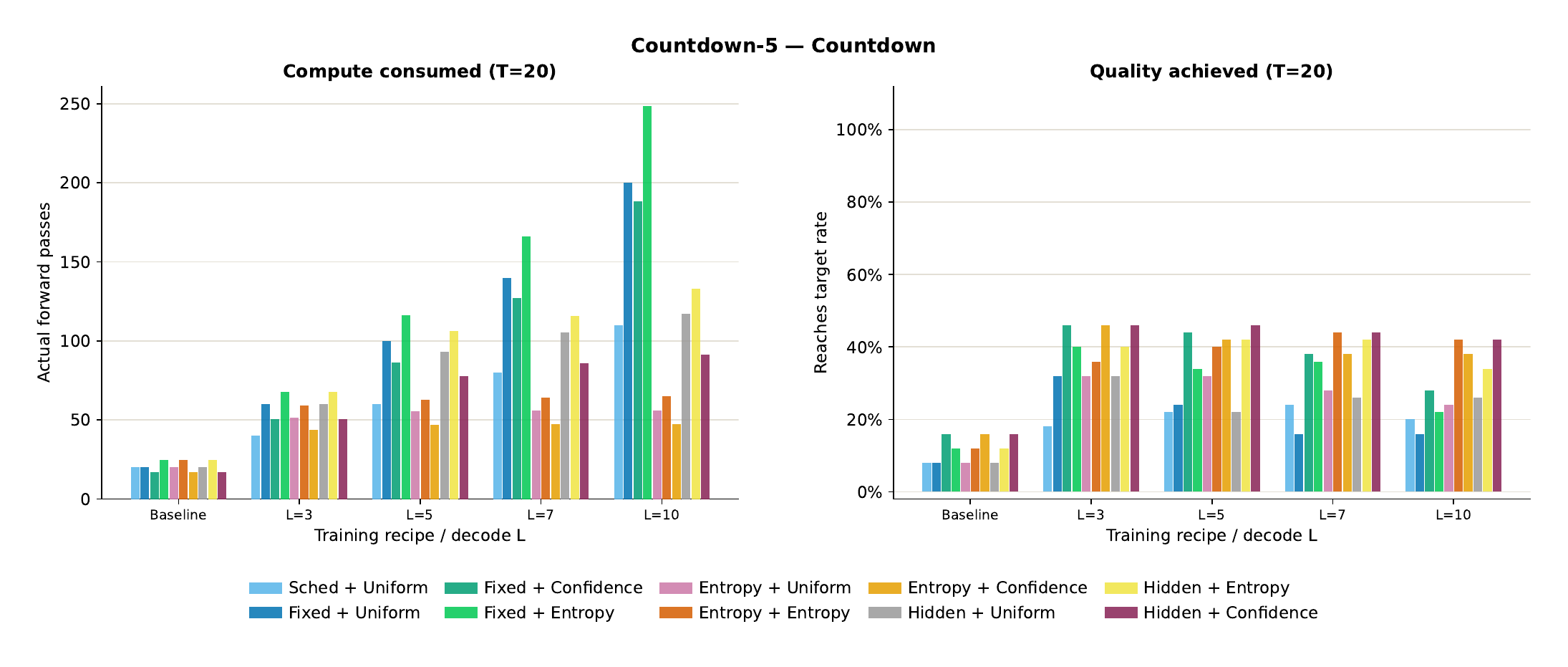}
        \caption{Countdown-5 (RTR), $T{=}20$.}
        \label{fig:decoding_bars_count5}
    \end{subfigure}
    \caption{Decoding comparison on Baseline vs.\ R-MDM ($L_s\in\{3,5,7,10\}$).
    Left: realized NFE (forward passes). Right: task quality at the indicated $T$.
    Entropy-gated policies (particularly \entent{}) match or beat \fixunif{} at substantially lower $\NFE$.}
    \label{fig:decoding_bars}
\end{figure*}

Figure~\ref{fig:decoding_bars} compares decoding policies on the four structured tasks (Sudoku~$9{\times}9$ and Countdown-$\{3,4,5\}$), reporting task quality against \emph{realized} forward passes ($\NFE$) at a fixed denoising budget $T$.
Bars contrast the non-recursive Baseline with R-MDM decoded at $L_s\in\{3,5,7,10\}$.

\paragraph{Entropy as a loop-stopping signal.}
Uniform commits spend a near-constant $L$ loops per denoising step, so $\NFE$ scales
roughly as $L{\times}T$.
Entropy-based controls break this coupling.
\textbf{Entropy-budget} commits (\fixent{}, \entent{}) stop unmasking once a Shannon
budget is exhausted, yielding low, nearly $T$-invariant $\NFE$.
\textbf{Entropy-loop} early stopping (\entunif{}, \entent{}) halts recursion when the mean
token-entropy gain falls below $\lambda$, so extra decode capacity is spent only
where the model is still uncertain.
Together, these mechanisms convert recursion from a fixed compute tax into an
adaptive budget that concentrates loops on hard tokens/steps.
\paragraph{Entropy-gain stopping with entropy commits.}
Across these four tasks, \entent{} is consistently near the quality--compute
frontier: it matches or exceeds \fixunif{} quality while using far fewer
forwards.
On Sudoku it reaches (near-)perfect VPR at single-digit $\NFE$, whereas \fixunif{}
needs $\sim\!L{\times}T$ forwards for the same saturation.
On Countdown-3 it attains the top Reaches-Target rate ($\approx 98\%$) at
$\NFE\!\approx\!10$--$15$, comparable to \fixent{} and well below
uniform looping.
On the harder Countdown-4/5 instances the gap is sharper: \fixunif{} peaks at
moderate $L$ and then \emph{degrades} as $L$ grows (quality drops while NFE keeps
climbing), whereas \entent{} stays competitive at high $L$ with roughly
stable, low $\NFE$---evidence that entropy-gated looping avoids wasteful
over-refinement.
\paragraph{Takeaway.}
Recursion helps when it is spent selectively.
Entropy-based stopping---especially the \entent{} combination of gain-based early
exit and budgeted commits---delivers the best quality per forward pass in these
plots, and is the most robust recipe when increasing $L$ on hard Countdown
instances. It is not uniformly best in absolute terms: on Countdown-5 the highest
reaches-target rate is still obtained by \fixunif{}, at several times the $\NFE$
(Section~\ref{sec:decoding_results}).

\clearpage

\subsection{RNA Infilling: Extended Results}
\label{app:rna_full}

Table~\ref{tab:rna-extended} reports the paired structural gap $\Delta Z$ and the realized \NFE{} for the four decoders of Section~\ref{sec:decoding_rules}, averaged over the three length strata ($n \in \{76, 110, 254\}$, decoded at $T = 6, 9, 20$) described in Appendix~\ref{app:datasets}. Rows with $L{=}1$ are the non-recursive baseline, for which the two loop rules coincide; $L \in \{3,5\}$ use the schedule-trained recursive checkpoint. Lower $\Delta Z$ is better. All mean values but one are positive, so infilled sequences are on average somewhat less structured than the native molecules they replace; the exception, \fixent{} at $L{=}5$ with $60\%$ masking ($\Delta Z=-0.014\pm0.063$), is within one standard error of zero. Since these are averages, individual completions can still fold more tightly than their native counterpart.

\begin{table*}[h]
\centering
\small
\setlength{\tabcolsep}{5pt}
\caption{RNA infilling at 40\%, 50\% and 60\% masked positions. We report the Z-score of the infilled
sequence, the paired structural gap $\Delta Z$, masked-position recovery, exact recovery and realized
\NFE{}. All values are unweighted means over the three length strata ($n\in\{76,110,254\}$, 500
sequences each) at $T=6$, $9$ and $20$ for the three strata; uncertainties are standard errors combined across strata.}
\label{tab:rna-extended}
\resizebox{0.85\linewidth}{!}{
\begin{tabular}{@{}llccccc@{}}
\toprule
\textbf{Policy} & $L$ & \textbf{Z} $\downarrow$ & $\Delta Z \downarrow$ & \textbf{Masked rec. \%} & \textbf{Exact rec. \%} $\downarrow$ & \textbf{NFE} \\
\midrule
\multicolumn{7}{c}{\textbf{40\% masked}} \\ \midrule
\fixunif{}   & 1 & $-$1.381 $\pm$ 0.052 & 0.478 $\pm$ 0.056 & 60.3 $\pm$ 0.7 & \textbf{1.6} $\pm$ 0.3 & 11.7 \\
             & 3 & $-$1.628 $\pm$ 0.056 & 0.229 $\pm$ 0.052 & 62.7 $\pm$ 0.7 & 2.5 $\pm$ 0.4 & 35.0 \\
             & 5 & $-$1.682 $\pm$ 0.054 & 0.175 $\pm$ 0.052 & 62.4 $\pm$ 0.7 & 2.3 $\pm$ 0.4 & 58.3 \\
\midrule
\fixent{}    & 1 & $-$1.408 $\pm$ 0.048 & 0.453 $\pm$ 0.055 & 60.3 $\pm$ 0.7 & 2.0 $\pm$ 0.4 & 13.2 \\
             & 3 & $-$1.688 $\pm$ 0.055 & 0.169 $\pm$ 0.051 & 62.4 $\pm$ 0.7 & 2.2 $\pm$ 0.4 & 33.5 \\
             & 5 & \textbf{$-$1.716} $\pm$ 0.057 & \textbf{0.140} $\pm$ 0.051 & 62.4 $\pm$ 0.7 & 1.8 $\pm$ 0.3 & 50.3 \\
\midrule
\entunif{}   & 1 & $-$1.381 $\pm$ 0.052 & 0.478 $\pm$ 0.056 & 60.3 $\pm$ 0.7 & \textbf{1.6} $\pm$ 0.3 & 11.7 \\
             & 3 & $-$1.554 $\pm$ 0.054 & 0.304 $\pm$ 0.052 & 62.1 $\pm$ 0.7 & 2.3 $\pm$ 0.4 & 26.7 \\
             & 5 & $-$1.588 $\pm$ 0.053 & 0.271 $\pm$ 0.050 & 62.3 $\pm$ 0.7 & 2.5 $\pm$ 0.4 & 27.4 \\
\midrule
\entent{}    & 1 & $-$1.408 $\pm$ 0.048 & 0.453 $\pm$ 0.055 & 60.3 $\pm$ 0.7 & 2.0 $\pm$ 0.4 & 13.2 \\
             & 3 & $-$1.612 $\pm$ 0.054 & 0.247 $\pm$ 0.049 & 62.0 $\pm$ 0.7 & 2.5 $\pm$ 0.4 & 25.9 \\
             & 5 & $-$1.658 $\pm$ 0.056 & 0.200 $\pm$ 0.051 & 62.3 $\pm$ 0.7 & 2.7 $\pm$ 0.4 & 26.2 \\
\midrule
\multicolumn{7}{c}{\textbf{50\% masked}} \\ \midrule
\fixunif{}   & 1 & $-$1.234 $\pm$ 0.047 & 0.625 $\pm$ 0.057 & 58.8 $\pm$ 0.7 & \textbf{1.0} $\pm$ 0.3 & 11.7 \\
             & 3 & $-$1.659 $\pm$ 0.057 & 0.199 $\pm$ 0.055 & 61.1 $\pm$ 0.7 & 1.7 $\pm$ 0.3 & 35.0 \\
             & 5 & $-$1.767 $\pm$ 0.057 & 0.092 $\pm$ 0.057 & 61.2 $\pm$ 0.7 & 1.6 $\pm$ 0.3 & 58.3 \\
\midrule
\fixent{}    & 1 & $-$1.413 $\pm$ 0.048 & 0.447 $\pm$ 0.056 & 58.8 $\pm$ 0.7 & 1.1 $\pm$ 0.3 & 16.5 \\
             & 3 & $-$1.746 $\pm$ 0.054 & 0.114 $\pm$ 0.055 & 61.0 $\pm$ 0.7 & 1.3 $\pm$ 0.3 & 41.5 \\
             & 5 & \textbf{$-$1.800} $\pm$ 0.058 & \textbf{0.056} $\pm$ 0.059 & 61.1 $\pm$ 0.7 & 1.5 $\pm$ 0.3 & 61.3 \\
\midrule
\entunif{}   & 1 & $-$1.234 $\pm$ 0.047 & 0.625 $\pm$ 0.057 & 58.8 $\pm$ 0.7 & \textbf{1.0} $\pm$ 0.3 & 11.7 \\
             & 3 & $-$1.544 $\pm$ 0.053 & 0.312 $\pm$ 0.054 & 60.9 $\pm$ 0.7 & 1.5 $\pm$ 0.3 & 26.7 \\
             & 5 & $-$1.512 $\pm$ 0.053 & 0.345 $\pm$ 0.055 & 60.7 $\pm$ 0.7 & 1.7 $\pm$ 0.3 & 27.6 \\
\midrule
\entent{}    & 1 & $-$1.413 $\pm$ 0.048 & 0.447 $\pm$ 0.056 & 58.8 $\pm$ 0.7 & 1.1 $\pm$ 0.3 & 16.5 \\
             & 3 & $-$1.619 $\pm$ 0.054 & 0.238 $\pm$ 0.055 & 60.9 $\pm$ 0.7 & 1.5 $\pm$ 0.3 & 32.0 \\
             & 5 & $-$1.584 $\pm$ 0.052 & 0.275 $\pm$ 0.054 & 60.9 $\pm$ 0.7 & 1.4 $\pm$ 0.3 & 32.4 \\
\midrule
\multicolumn{7}{c}{\textbf{60\% masked}} \\ \midrule
\fixunif{}   & 1 & $-$1.257 $\pm$ 0.049 & 0.604 $\pm$ 0.061 & 57.4 $\pm$ 0.7 & \textbf{0.7} $\pm$ 0.2 & 11.7 \\
             & 3 & $-$1.588 $\pm$ 0.055 & 0.270 $\pm$ 0.060 & 59.3 $\pm$ 0.7 & 1.3 $\pm$ 0.3 & 35.0 \\
             & 5 & $-$1.765 $\pm$ 0.059 & 0.094 $\pm$ 0.062 & 59.7 $\pm$ 0.7 & 1.0 $\pm$ 0.3 & 58.3 \\
\midrule
\fixent{}    & 1 & $-$1.340 $\pm$ 0.047 & 0.520 $\pm$ 0.060 & 57.3 $\pm$ 0.7 & \textbf{0.7} $\pm$ 0.2 & 19.7 \\
             & 3 & $-$1.769 $\pm$ 0.058 & 0.090 $\pm$ 0.060 & 59.2 $\pm$ 0.7 & 1.0 $\pm$ 0.3 & 48.7 \\
             & 5 & \textbf{$-$1.871} $\pm$ 0.062 & \textbf{$-$0.014} $\pm$ 0.063 & 59.5 $\pm$ 0.7 & 0.8 $\pm$ 0.2 & 70.9 \\
\midrule
\entunif{}   & 1 & $-$1.257 $\pm$ 0.049 & 0.604 $\pm$ 0.061 & 57.4 $\pm$ 0.7 & \textbf{0.7} $\pm$ 0.2 & 11.7 \\
             & 3 & $-$1.501 $\pm$ 0.053 & 0.358 $\pm$ 0.058 & 58.9 $\pm$ 0.7 & 1.0 $\pm$ 0.3 & 26.7 \\
             & 5 & $-$1.464 $\pm$ 0.054 & 0.395 $\pm$ 0.058 & 59.1 $\pm$ 0.7 & 0.9 $\pm$ 0.2 & 27.5 \\
\midrule
\entent{}    & 1 & $-$1.340 $\pm$ 0.047 & 0.520 $\pm$ 0.060 & 57.3 $\pm$ 0.7 & \textbf{0.7} $\pm$ 0.2 & 19.7 \\
             & 3 & $-$1.570 $\pm$ 0.054 & 0.289 $\pm$ 0.058 & 59.2 $\pm$ 0.7 & 0.9 $\pm$ 0.2 & 38.3 \\
             & 5 & $-$1.600 $\pm$ 0.055 & 0.259 $\pm$ 0.061 & 58.9 $\pm$ 0.7 & 0.9 $\pm$ 0.2 & 38.7 \\
\bottomrule
\end{tabular}}
\end{table*}

We evaluate partial infilling on held-out validation sequences: a fraction of
nucleotide positions is blanked uniformly at random and the model fills them
with the remaining positions visible. 

\clearpage
\subsection{TinyGSM: Extended Results}
\label{app:tinygsm_full}

Decoding is infill on a 512-token canvas: the GSM8K question is kept as a prefix and every subsequent position is masked. We evaluate $n{=}512$ GSM8K test problems, shared across every cell so paired comparisons use common random numbers. Tokens are decoded greedily ($\mathrm{top}\text{-}k{=}1$, temperature $0.8$). 

\begin{table}[h]
\centering
\small
\setlength{\tabcolsep}{4.5pt}
\caption{TinyGSM GSM8K accuracy, mean NFE and sampling time for the
schedule-trained recursive checkpoint ($L\in\{3,5\}$) and the non-recursive
baseline ($L{=}1$). Uncertainties are binomial standard errors with $n=512$ shared test problems.
Sampling time is the total wall-clock time to decode all 512 problems.}
\label{tab:tinygsm-extended}
\resizebox{0.7\linewidth}{!}{
\begin{tabular}{@{}llccccc@{}}
\toprule
\textbf{Policy} & $L$ & $T$ & \textbf{Acc.} & \textbf{NFE} & \textbf{Sampling time (s)} \\
\midrule
\fixunif{} & 1 & 32 & 6.4\% $\pm$ 1.1 & 32 & 283 \\
& 1 & 128 & \textbf{15.0\%} $\pm$ 1.6 & 128 & 1118 \\
& 1 & 256 & 11.5\% $\pm$ 1.4 & 256 & 2237 \\
& 1 & 448 & 10.5\% $\pm$ 1.4 & 440 & 3792 \\
\entunif{} & 1 & 32 & 6.4\% $\pm$ 1.1 & 32 & 306 \\
& 1 & 128 & 15.0\% $\pm$ 1.6 & 128 & 1184 \\
& 1 & 256 & 11.5\% $\pm$ 1.4 & 256 & 2286 \\
& 1 & 448 & 10.5\% $\pm$ 1.4 & 440 & 3971 \\
\fixent{} & 1 & --- & 13.7\% $\pm$ 1.5 & 22 & 218 \\
\entent{} & 1 & --- & 13.7\% $\pm$ 1.5 & 22 & 213 \\
\midrule
\fixunif{} & 3 & 32 & 16.0\% $\pm$ 1.6 & 96 & 566 \\
& 3 & 128 & 21.5\% $\pm$ 1.8 & 384 & 2202 \\
& 3 & 256 & \textbf{24.2\%} $\pm$ 1.9 & 768 & 4407 \\
& 3 & 448 & 23.4\% $\pm$ 1.9 & 1319 & 7569 \\
\entunif{} & 3 & 32 & 13.5\% $\pm$ 1.5 & 82 & 519 \\
& 3 & 128 & 21.1\% $\pm$ 1.8 & 326 & 2043 \\
& 3 & 256 & 21.9\% $\pm$ 1.8 & 650 & 4100 \\
& 3 & 448 & 20.3\% $\pm$ 1.8 & 1115 & 6987 \\
\fixent{} & 3 & --- & 24.0\% $\pm$ 1.9 & 69 & 412 \\
\entent{} & 3 & --- & 22.3\% $\pm$ 1.8 & 57 & 374 \\
\midrule
\fixunif{} & 5 & 32 & 7.8\% $\pm$ 1.2 & 160 & 906 \\
& 5 & 128 & 16.0\% $\pm$ 1.6 & 640 & 3623 \\
& 5 & 256 & 15.8\% $\pm$ 1.6 & 1280 & 7230 \\
& 5 & 448 & 15.8\% $\pm$ 1.6 & 2198 & 12382 \\
\entunif{} & 5 & 32 & 11.1\% $\pm$ 1.4 & 83 & 525 \\
& 5 & 128 & 16.8\% $\pm$ 1.7 & 329 & 2066 \\
& 5 & 256 & 17.4\% $\pm$ 1.7 & 653 & 4103 \\
& 5 & 448 & 16.4\% $\pm$ 1.6 & 1113 & 6977 \\
\fixent{} & 5 & --- & 18.4\% $\pm$ 1.7 & 136 & 790 \\
\entent{} & 5 & --- & 17.2\% $\pm$ 1.7 & 63 & 406 \\
\bottomrule
\end{tabular}}
\end{table}

Recursion helps. The schedule-trained model peaks at $24.2\%\pm 1.9$ (\fixunif{}, $L{=}3$, $T{=}256$), against $15.0\%\pm 1.6$ for the non-recursive baseline at its best setting ($T{=}128$). Every recursive cell at $L{=}3$ with $T\ge 128$, and every entropy-budget cell at $L{=}3$, sits above that baseline.

Commit width is the most useful adaptive axis. \fixent{} at $L{=}3$ matches the headline accuracy ($24.0\%\pm 1.9$ vs $24.2\%\pm 1.9$) at $68.8$ NFE instead of $768$ (an $11{\times}$ saving) because $\delta{=}2$ commits several confident tokens per step rather than spreading them across a long uniform schedule. Pushing uniform $T$ from $256$ to $448$ does not help ($24.2\%\to 23.4\%$). Entropy-gain stopping on top of that budget (\entent{}) is slightly worse ($22.3\%\pm 1.8$) at comparable cost ($57$ NFE): the loop controller trims only about half a loop relative to $L{=}3$ and does not improve solves. Entropy-gain stopping with a uniform schedule (\entunif{}) likewise tracks \fixunif{} but never overtakes it.

On the baseline, extra denoising after $T{=}128$ hurts ($15.0\%\to 11.5\%$ at $T{=}256$), and the entropy budget is a cheap alternative ($13.7\%$ at $22$ NFE).

\clearpage

\section{Samples}
\label{app:samples}

\subsection{Sudoku $9 \times 9$}

\begin{figure*}[ht]
    \centering
    \includegraphics[width=\textwidth]{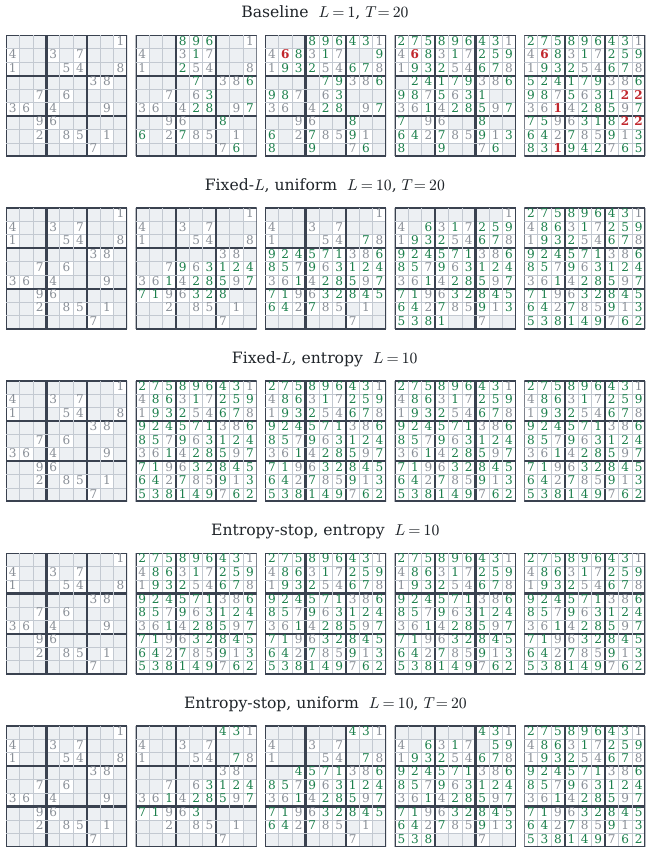}
    \caption{Sudoku trajectories.}
    \vspace{-0.5cm}
    \label{fig:sudokusamples}
\end{figure*}

\clearpage

\subsection{Countdown-5}

\definecolor{cderr}{HTML}{C1272D}
\definecolor{cdok}{HTML}{1B7F4B}

\begin{table*}[ht]
\caption{Countdown-5 samples.}
\label{tab:countdownsamples}
\vspace{-6pt}
\centering
\resizebox{0.95\linewidth}{!}{
\begin{tabular}{@{}lll@{}}
\toprule
  Problem & Model completion & Outcome \\
\midrule
  \multicolumn{3}{@{}l@{}}{\textbf{Baseline ($L{=}1$, $T{=}20$)}} \\
  \addlinespace[1pt]
  \texttt{79,7,7,72,57=108} & \texttt{72+7=79,79+7=86,79-57=22,22+86=108} & \textcolor{cdok}{valid} \\
  \texttt{41,92,85,18,71=165} & \texttt{\textcolor{cderr}{85+85=126},92-71=21,18+126=144,144+21=165} & arithmetic, number reused \\
  \texttt{42,23,16,51,53=-980} & \texttt{16+\textcolor{cderr}{53}=69,\textcolor{cderr}{42-69=-23},-23*\textcolor{cderr}{53}=-929,-929-51=-980} & arithmetic, number reused \\
\midrule
  \multicolumn{3}{@{}l@{}}{\textbf{\fixunif{} ($L{=}5$, $T{=}20$)}} \\
  \addlinespace[1pt]
  \texttt{79,43,65,68,11=244} & \texttt{68+79=147,65+43=108,11-147=-136,108--136=244} & \textcolor{cdok}{valid} \\
  \texttt{79,7,7,72,57=108} & \texttt{72+7=79,7+79=86,79-57=22,86+22=108} & \textcolor{cdok}{valid} \\
  \texttt{41,92,85,18,71=165} & \texttt{\textcolor{cderr}{18+18=103},103-71=32,92+92=124,124+41=165} & arithmetic, number reused \\
  \texttt{42,23,16,51,53=-980} & \texttt{\textcolor{cderr}{16+16=96},23-51=-28,-2\textcolor{cderr}{**}43=-984,-984-96=-980} & arithmetic, number reused, malformed \\
\midrule
  \multicolumn{3}{@{}l@{}}{\textbf{\fixent{} ($L{=}5$)}} \\
  \addlinespace[1pt]
  \texttt{79,7,7,72,57=108} & \texttt{72+7=79,79-57=22,7+22=29,29+79=108} & \textcolor{cdok}{valid} \\
  \texttt{24,87,74,61,90=214} & \texttt{87-61=26,24+26=50,74+90=164,164+50=214} & \textcolor{cdok}{valid} \\
  \texttt{41,92,85,18,71=165} & \texttt{85+92=177,18+41=59,\textcolor{cderr}{59-177=-94},71--94=165} & arithmetic \\
  \texttt{42,23,16,51,53=-980} & \texttt{23-53=-30,\textcolor{cderr}{-30-16=-76},51+-76=-25,42*-25=-980} & arithmetic \\
\midrule
  \multicolumn{3}{@{}l@{}}{\textbf{\entent{} ($L{=}5$)}} \\
  \addlinespace[1pt]
  \texttt{41,92,85,18,71=165} & \texttt{41+85=126,126+18=144,92-71=21,21+144=165} & \textcolor{cdok}{valid} \\
  \texttt{79,43,65,68,11=244} & \texttt{65+79=144,43+144=187,11-187=-176,68--176=244} & \textcolor{cdok}{valid} \\
  \texttt{42,23,16,51,53=-980} & \texttt{\textcolor{cderr}{53}-42=11,\textcolor{cderr}{16*53=968},11-968=-957,-957-23=-980} & arithmetic, number reused \\
  \texttt{45,66,65,6,34=1485} & \texttt{65-6=59,66-34=32,59-32=27,\textcolor{cderr}{45*27=1485}} & arithmetic \\
\midrule
  \multicolumn{3}{@{}l@{}}{\textbf{\entunif{} ($L{=}5$, $T{=}20$)}} \\
  \addlinespace[1pt]
  \texttt{79,7,7,72,57=108} & \texttt{7+72=79,7+79=86,79-57=22,86+22=108} & \textcolor{cdok}{valid} \\
  \texttt{4,35,33,62,63=7} & \texttt{33-4=29,62-35=27,63-29=34,34-27=7} & \textcolor{cdok}{valid} \\
  \texttt{41,92,85,18,71=165} & \texttt{\textcolor{cderr}{18+18=103},103-71=32,32+92=124,124+41=165} & arithmetic, number reused \\
  \texttt{42,23,16,51,53=-980} & \texttt{42-23=19,\textcolor{cderr}{51*51=946},19-946=-927,-927-53=-980} & arithmetic, number reused \\
\bottomrule
\end{tabular}}
\end{table*}

\subsection{TinyGSM}

\definecolor{cderr}{HTML}{C1272D}
\definecolor{cdok}{HTML}{1B7F4B}

{%
\small
\setlength{\tabcolsep}{5pt}%
\centering
\resizebox{0.9\linewidth}{!}{
\begin{tabular}{@{}>{\raggedright\arraybackslash}p{0.32\linewidth}
                 >{\raggedright\arraybackslash\ttfamily}p{0.45\linewidth}
                 >{\raggedright\arraybackslash}p{0.15\linewidth}@{}}
\toprule
  Problem & Generated program & Outcome \\
\midrule
  \multicolumn{3}{@{}l@{}}{\textbf{Baseline ($L{=}1$, $T{=}128$)}} \\
  \addlinespace[4pt]
  Argo has 200 toys. He gives 40 to Alyssa, 80 to Bonnie, and 30 to Nicky. How many remain?
    & toys=200; toys-=40; toys-=80; toys-=30
    & \textcolor{cdok}{valid} \\
  \addlinespace[6pt]
  Laptop \$1000, 20\% off. How much to pay?
    & result=\textcolor{cderr}{20}
    & constant guess \\
  \addlinespace[6pt]
  3 rose bushes, 25 roses each, 8 thorns per rose. Total thorns?
    & result=\textcolor{cderr}{18}
    & constant guess \\
  \addlinespace[6pt]
  Ann is 9; her brother is twice her age. Brother's age in 3 years?
    & result=\textcolor{cderr}{\ldots}
    & truncated \\
\midrule
  \multicolumn{3}{@{}l@{}}{\textbf{\fixunif{} ($L{=}3$, $T{=}256$)}} \\
  \addlinespace[4pt]
  Laptop \$1000, 20\% off. How much to pay?
    & price=1000; discount=0.2; discounted\_price=\allowbreak{}price-\allowbreak{}(\allowbreak{}price*\allowbreak{}discount)\allowbreak{}
    & \textcolor{cdok}{valid} \\
  \addlinespace[6pt]
  3 rose bushes, 25 roses each, 8 thorns per rose. Total thorns?
    & roses=3*25; thorns=roses*8
    & \textcolor{cdok}{valid} \\
  \addlinespace[6pt]
  22 games, won 8 more than lost. How many wins?
    & games\_played=22; games\_won=\allowbreak{}\textcolor{cderr}{games\_played+\allowbreak{}8}
    & wrong equation \\
  \addlinespace[6pt]
  Eraser \$2, pencil \$3; cost of 6 erasers and 8 pencils?
    & eraser=2; pencil=3; total=\allowbreak{}6*\allowbreak{}eraser+\allowbreak{}8*\allowbreak{}pencil \textcolor{cderr}{\normalfont\itshape(no return)}
    & no return \\
\midrule
  \multicolumn{3}{@{}l@{}}{\textbf{\fixent{} ($L{=}3$)}} \\
  \addlinespace[4pt]
  Laptop \$1000, 20\% off. How much to pay?
    & price=1000; discount=0.2; discounted\_price=\allowbreak{}price-\allowbreak{}(\allowbreak{}price*\allowbreak{}discount)\allowbreak{}
    & \textcolor{cdok}{valid} \\
  \addlinespace[6pt]
  3 rose bushes, 25 roses each, 8 thorns per rose. Total thorns?
    & roses=3*25; thorns=roses*8
    & \textcolor{cdok}{valid} \\
  \addlinespace[6pt]
  22 games, won 8 more than lost. How many wins?
    & total\_games=22; games\_won=\allowbreak{}(\allowbreak{}total\_games+\allowbreak{}8)\allowbreak{}/\allowbreak{}2
    & \textcolor{cdok}{valid} \\
  \addlinespace[6pt]
  Eraser \$2, pencil \$3; cost of 6 erasers and 8 pencils?
    & erasers=6; pencils=8; cost=\allowbreak{}erasers*\allowbreak{}2+\allowbreak{}pencils*\allowbreak{}3 \textcolor{cderr}{\normalfont\itshape(no return)}
    & no return \\
\midrule
  \multicolumn{3}{@{}l@{}}{\textbf{\entent{} ($L{=}3$)}} \\
  \addlinespace[4pt]
  Laptop \$1000, 20\% off. How much to pay?
    & price=1000; discount=0.2; discounted\_price=\allowbreak{}price-\allowbreak{}(\allowbreak{}price*\allowbreak{}discount)\allowbreak{}
    & \textcolor{cdok}{valid} \\
  \addlinespace[6pt]
  22 games, won 8 more than lost. How many wins?
    & total\_games=22; wins=\allowbreak{}(\allowbreak{}total\_games+\allowbreak{}8)\allowbreak{}/\allowbreak{}2
    & \textcolor{cdok}{valid} \\
  \addlinespace[6pt]
  Eraser \$2, pencil \$3; cost of 6 erasers and 8 pencils?
    & cost=2*6+3*8
    & \textcolor{cdok}{valid} \\
  \addlinespace[6pt]
  John collects peaches for 3 hours at 2 per minute. How many peaches?
    & hours=3; minutes=2; minutes=hours*60; result=\textcolor{cderr}{hours*minutes}
    & wrong equation \\
\midrule
  \multicolumn{3}{@{}l@{}}{\textbf{\entunif{} ($L{=}3$, $T{=}256$)}} \\
  \addlinespace[4pt]
  Laptop \$1000, 20\% off. How much to pay?
    & price=1000; discount=0.2; discounted\_price=\allowbreak{}price-\allowbreak{}(\allowbreak{}price*\allowbreak{}discount)\allowbreak{}
    & \textcolor{cdok}{valid} \\
  \addlinespace[6pt]
  3 rose bushes, 25 roses each, 8 thorns per rose. Total thorns?
    & roses=3*25; thorns=roses*8
    & \textcolor{cdok}{valid} \\
  \addlinespace[6pt]
  22 games, won 8 more than lost. How many wins?
    & games\_played=22; games\_won=\allowbreak{}\textcolor{cderr}{games\_played+\allowbreak{}8}
    & wrong equation \\
  \addlinespace[6pt]
  Eraser \$2, pencil \$3; cost of 6 erasers and 8 pencils?
    & eraser=2; pencil=3; total=\allowbreak{}6*\allowbreak{}eraser+\allowbreak{}8*\allowbreak{}pencil \textcolor{cderr}{\normalfont\itshape(no return)}
    & no return \\
\bottomrule
\end{tabular}}
\par
}%

\clearpage

\subsection{RNA}

\definecolor{cderr}{HTML}{C1272D}
\definecolor{cdgrey}{HTML}{8A8A8A}

{%
\small\ttfamily
\setlength{\tabcolsep}{3pt}%
\begin{tabular}{@{}l l@{}}
  \multicolumn{2}{@{}l@{}}{\normalfont\bfseries Held-out validation sequence \#285 ($n{=}76$\,nt)} \\
  \addlinespace[5pt]
  \multicolumn{2}{@{}l@{}}{\normalfont\itshape 40\% blanks (30/76)} \\
  WT & AGGGCUAUAGCUCAGUUAGGUAGAGCGCCUCGUUUACACACCGAGAAGGUCUACGGUUCGAAUCCGUAUAGCCCUA \\
  IN & AG\textcolor{cdgrey}{.}GC\textcolor{cdgrey}{.}\textcolor{cdgrey}{.}UAGCU\textcolor{cdgrey}{.}\textcolor{cdgrey}{.}G\textcolor{cdgrey}{.}\textcolor{cdgrey}{.}AG\textcolor{cdgrey}{.}UAGA\textcolor{cdgrey}{.}\textcolor{cdgrey}{.}\textcolor{cdgrey}{.}C\textcolor{cdgrey}{.}UC\textcolor{cdgrey}{.}UUUACACAC\textcolor{cdgrey}{.}\textcolor{cdgrey}{.}AGA\textcolor{cdgrey}{.}\textcolor{cdgrey}{.}\textcolor{cdgrey}{.}U\textcolor{cdgrey}{.}U\textcolor{cdgrey}{.}CGG\textcolor{cdgrey}{.}\textcolor{cdgrey}{.}\textcolor{cdgrey}{.}\textcolor{cdgrey}{.}AA\textcolor{cdgrey}{.}\textcolor{cdgrey}{.}CGUAU\textcolor{cdgrey}{.}\textcolor{cdgrey}{.}\textcolor{cdgrey}{.}CCU\textcolor{cdgrey}{.} \\
  B1 & AGGGCUAUAGCUCAGU\textcolor{cderr}{C}AGGUAGAGC\textcolor{cderr}{A}CCUCGUUUACACAC\textcolor{cderr}{A}GAGAAGGUCU\textcolor{cderr}{C}CGGUUC\textcolor{cderr}{A}AAUCCGUAUAGCCCUA \\
     & \normalfont\scriptsize Baseline $L{=}1$: recovery 0.83, $Z=-3.58$ (WT $Z=-2.27$) \\
  R5 & AGGGCUAUAGCUCAGU\textcolor{cderr}{C}AGGUAGAGC\textcolor{cderr}{A}C\textcolor{cderr}{U}UCGUUUACACAC\textcolor{cderr}{G}\textcolor{cderr}{A}AGAAGGUCUACGGUUC\textcolor{cderr}{A}AAUCCGUAUAGCCCUA \\
     & \normalfont\scriptsize Recursive $L{=}5$: recovery 0.80, $Z=-3.84$ (WT $Z=-2.27$) \\
  \addlinespace[5pt]
  \multicolumn{2}{@{}l@{}}{\normalfont\itshape 50\% blanks (38/76)} \\
  WT & AGGGCUAUAGCUCAGUUAGGUAGAGCGCCUCGUUUACACACCGAGAAGGUCUACGGUUCGAAUCCGUAUAGCCCUA \\
  IN & AG\textcolor{cdgrey}{.}G\textcolor{cdgrey}{.}\textcolor{cdgrey}{.}\textcolor{cdgrey}{.}UAGCU\textcolor{cdgrey}{.}\textcolor{cdgrey}{.}\textcolor{cdgrey}{.}\textcolor{cdgrey}{.}\textcolor{cdgrey}{.}AG\textcolor{cdgrey}{.}UA\textcolor{cdgrey}{.}A\textcolor{cdgrey}{.}\textcolor{cdgrey}{.}\textcolor{cdgrey}{.}C\textcolor{cdgrey}{.}UC\textcolor{cdgrey}{.}UU\textcolor{cdgrey}{.}ACA\textcolor{cdgrey}{.}AC\textcolor{cdgrey}{.}\textcolor{cdgrey}{.}AGA\textcolor{cdgrey}{.}\textcolor{cdgrey}{.}\textcolor{cdgrey}{.}U\textcolor{cdgrey}{.}U\textcolor{cdgrey}{.}\textcolor{cdgrey}{.}GG\textcolor{cdgrey}{.}\textcolor{cdgrey}{.}\textcolor{cdgrey}{.}\textcolor{cdgrey}{.}AA\textcolor{cdgrey}{.}\textcolor{cdgrey}{.}CGUAU\textcolor{cdgrey}{.}\textcolor{cdgrey}{.}\textcolor{cdgrey}{.}C\textcolor{cdgrey}{.}\textcolor{cdgrey}{.}\textcolor{cdgrey}{.} \\
  B1 & AGGGCUAUAGCUCAGU\textcolor{cderr}{C}AG\textcolor{cderr}{U}UAGAGCGCCUCGUU\textcolor{cderr}{C}ACACAC\textcolor{cderr}{A}GAGAAGGUCUACGGUUC\textcolor{cderr}{A}AAUCCGUAUA\textcolor{cderr}{C}CCCUA \\
     & \normalfont\scriptsize Baseline $L{=}1$: recovery 0.84, $Z=-2.42$ (WT $Z=-2.27$) \\
  R5 & AGGGCUAUAGCUCAG\textcolor{cderr}{C}\textcolor{cderr}{C}AGGUAGAGC\textcolor{cderr}{A}C\textcolor{cderr}{U}UCGUUUACACAC\textcolor{cderr}{G}\textcolor{cderr}{A}AGAA\textcolor{cderr}{U}GUCUACGGUUCGAAUCCGUAUA\textcolor{cderr}{C}CCCUA \\
     & \normalfont\scriptsize Recursive $L{=}5$: recovery 0.79, $Z=-2.40$ (WT $Z=-2.27$) \\
  \addlinespace[5pt]
  \multicolumn{2}{@{}l@{}}{\normalfont\itshape 60\% blanks (46/76)} \\
  WT & AGGGCUAUAGCUCAGUUAGGUAGAGCGCCUCGUUUACACACCGAGAAGGUCUACGGUUCGAAUCCGUAUAGCCCUA \\
  IN & AG\textcolor{cdgrey}{.}G\textcolor{cdgrey}{.}\textcolor{cdgrey}{.}\textcolor{cdgrey}{.}UA\textcolor{cdgrey}{.}CU\textcolor{cdgrey}{.}\textcolor{cdgrey}{.}\textcolor{cdgrey}{.}\textcolor{cdgrey}{.}\textcolor{cdgrey}{.}\textcolor{cdgrey}{.}G\textcolor{cdgrey}{.}\textcolor{cdgrey}{.}A\textcolor{cdgrey}{.}A\textcolor{cdgrey}{.}\textcolor{cdgrey}{.}\textcolor{cdgrey}{.}C\textcolor{cdgrey}{.}U\textcolor{cdgrey}{.}\textcolor{cdgrey}{.}U\textcolor{cdgrey}{.}\textcolor{cdgrey}{.}ACA\textcolor{cdgrey}{.}\textcolor{cdgrey}{.}C\textcolor{cdgrey}{.}\textcolor{cdgrey}{.}AGA\textcolor{cdgrey}{.}\textcolor{cdgrey}{.}\textcolor{cdgrey}{.}U\textcolor{cdgrey}{.}U\textcolor{cdgrey}{.}\textcolor{cdgrey}{.}GG\textcolor{cdgrey}{.}\textcolor{cdgrey}{.}\textcolor{cdgrey}{.}\textcolor{cdgrey}{.}A\textcolor{cdgrey}{.}\textcolor{cdgrey}{.}\textcolor{cdgrey}{.}CGU\textcolor{cdgrey}{.}U\textcolor{cdgrey}{.}\textcolor{cdgrey}{.}\textcolor{cdgrey}{.}C\textcolor{cdgrey}{.}\textcolor{cdgrey}{.}\textcolor{cdgrey}{.} \\
  B1 & AGGG\textcolor{cderr}{G}UAUAGCUCAGUU\textcolor{cderr}{G}G\textcolor{cderr}{U}UAGAGC\textcolor{cderr}{A}CCU\textcolor{cderr}{G}\textcolor{cderr}{U}UU\textcolor{cderr}{C}ACAC\textcolor{cderr}{U}C\textcolor{cderr}{A}GAGAAGGUCUACGGUUC\textcolor{cderr}{A}A\textcolor{cderr}{U}UCCGUAU\textcolor{cderr}{G}\textcolor{cderr}{U}CCCUA \\
     & \normalfont\scriptsize Baseline $L{=}1$: recovery 0.72, $Z=-0.93$ (WT $Z=-2.27$) \\
  R5 & AGGG\textcolor{cderr}{U}UAUAGCUCAGUU\textcolor{cderr}{G}G\textcolor{cderr}{U}UAGAGC\textcolor{cderr}{A}C\textcolor{cderr}{U}U\textcolor{cderr}{G}\textcolor{cderr}{U}UUUACA\textcolor{cderr}{U}\textcolor{cderr}{U}C\textcolor{cderr}{U}\textcolor{cderr}{A}AGAA\textcolor{cderr}{A}GUCUACGGUUC\textcolor{cderr}{A}AAUCCGUAU\textcolor{cderr}{G}GCCCUA \\
     & \normalfont\scriptsize Recursive $L{=}5$: recovery 0.70, $Z=-1.47$ (WT $Z=-2.27$) \\
  \addlinespace[8pt]
  \multicolumn{2}{@{}l@{}}{\normalfont\bfseries Held-out validation sequence \#463 ($n{=}76$\,nt)} \\
  \addlinespace[5pt]
  \multicolumn{2}{@{}l@{}}{\normalfont\itshape 40\% blanks (30/76)} \\
  WT & CUCCUGUUAGUUCAAACGGAAGAACAUCGACCUUCUAAGUCGCAGGUUACUGGUUCGAGUCCAGUACGGGAGACAA \\
  IN & C\textcolor{cdgrey}{.}C\textcolor{cdgrey}{.}UGUUA\textcolor{cdgrey}{.}U\textcolor{cdgrey}{.}CAA\textcolor{cdgrey}{.}C\textcolor{cdgrey}{.}G\textcolor{cdgrey}{.}AG\textcolor{cdgrey}{.}A\textcolor{cdgrey}{.}AU\textcolor{cdgrey}{.}\textcolor{cdgrey}{.}ACCUU\textcolor{cdgrey}{.}U\textcolor{cdgrey}{.}\textcolor{cdgrey}{.}G\textcolor{cdgrey}{.}C\textcolor{cdgrey}{.}C\textcolor{cdgrey}{.}GGU\textcolor{cdgrey}{.}\textcolor{cdgrey}{.}C\textcolor{cdgrey}{.}GG\textcolor{cdgrey}{.}\textcolor{cdgrey}{.}\textcolor{cdgrey}{.}GA\textcolor{cdgrey}{.}UC\textcolor{cdgrey}{.}AGUAC\textcolor{cdgrey}{.}GGA\textcolor{cdgrey}{.}\textcolor{cdgrey}{.}\textcolor{cdgrey}{.}A\textcolor{cdgrey}{.} \\
  B1 & CUCCUGUUAGUUCAA\textcolor{cderr}{U}CGG\textcolor{cderr}{C}AGAACAU\textcolor{cderr}{U}GACCUU\textcolor{cderr}{U}UAAGUC\textcolor{cderr}{A}CAGGUUACUGGUUCGAGUCCAGUACGGGAG\textcolor{cderr}{G}CAA \\
     & \normalfont\scriptsize Baseline $L{=}1$: recovery 0.80, $Z=-1.68$ (WT $Z=-2.24$) \\
  R5 & CUC\textcolor{cderr}{U}UGUUAGUUCAA\textcolor{cderr}{C}CGG\textcolor{cderr}{C}AGAACAU\textcolor{cderr}{U}GACCUU\textcolor{cderr}{U}UAAGUC\textcolor{cderr}{A}CAGGUUACUGGUUCGA\textcolor{cderr}{A}UCCAGUAC\textcolor{cderr}{A}GGAGACAA \\
     & \normalfont\scriptsize Recursive $L{=}5$: recovery 0.73, $Z=-1.99$ (WT $Z=-2.24$) \\
  \addlinespace[5pt]
  \multicolumn{2}{@{}l@{}}{\normalfont\itshape 50\% blanks (38/76)} \\
  WT & CUCCUGUUAGUUCAAACGGAAGAACAUCGACCUUCUAAGUCGCAGGUUACUGGUUCGAGUCCAGUACGGGAGACAA \\
  IN & C\textcolor{cdgrey}{.}C\textcolor{cdgrey}{.}U\textcolor{cdgrey}{.}UUA\textcolor{cdgrey}{.}U\textcolor{cdgrey}{.}CAA\textcolor{cdgrey}{.}C\textcolor{cdgrey}{.}\textcolor{cdgrey}{.}\textcolor{cdgrey}{.}A\textcolor{cdgrey}{.}\textcolor{cdgrey}{.}A\textcolor{cdgrey}{.}AU\textcolor{cdgrey}{.}\textcolor{cdgrey}{.}ACCUU\textcolor{cdgrey}{.}\textcolor{cdgrey}{.}\textcolor{cdgrey}{.}\textcolor{cdgrey}{.}G\textcolor{cdgrey}{.}C\textcolor{cdgrey}{.}C\textcolor{cdgrey}{.}G\textcolor{cdgrey}{.}U\textcolor{cdgrey}{.}\textcolor{cdgrey}{.}C\textcolor{cdgrey}{.}GG\textcolor{cdgrey}{.}\textcolor{cdgrey}{.}\textcolor{cdgrey}{.}G\textcolor{cdgrey}{.}\textcolor{cdgrey}{.}UC\textcolor{cdgrey}{.}\textcolor{cdgrey}{.}\textcolor{cdgrey}{.}UAC\textcolor{cdgrey}{.}GGA\textcolor{cdgrey}{.}\textcolor{cdgrey}{.}\textcolor{cdgrey}{.}A\textcolor{cdgrey}{.} \\
  B1 & CUCCUGUUAGUUCAA\textcolor{cderr}{U}CGG\textcolor{cderr}{U}AGAACAU\textcolor{cderr}{U}GACCUU\textcolor{cderr}{U}\textcolor{cderr}{A}AAGUCGCAGGUUAC\textcolor{cderr}{A}GGUUCGAGUCC\textcolor{cderr}{U}GUACGGGAG\textcolor{cderr}{G}CAA \\
     & \normalfont\scriptsize Baseline $L{=}1$: recovery 0.79, $Z=-1.52$ (WT $Z=-2.24$) \\
  R5 & CUCCUGUUAGUUCAA\textcolor{cderr}{U}CGG\textcolor{cderr}{C}AGAACAU\textcolor{cderr}{U}GACCUU\textcolor{cderr}{U}UAAGUCGCAGGUUAC\textcolor{cderr}{A}GGUUCGA\textcolor{cderr}{A}UCC\textcolor{cderr}{U}GUACGGGAGACAA \\
     & \normalfont\scriptsize Recursive $L{=}5$: recovery 0.82, $Z=-1.95$ (WT $Z=-2.24$) \\
  \addlinespace[5pt]
  \multicolumn{2}{@{}l@{}}{\normalfont\itshape 60\% blanks (46/76)} \\
  WT & CUCCUGUUAGUUCAAACGGAAGAACAUCGACCUUCUAAGUCGCAGGUUACUGGUUCGAGUCCAGUACGGGAGACAA \\
  IN & C\textcolor{cdgrey}{.}\textcolor{cdgrey}{.}\textcolor{cdgrey}{.}\textcolor{cdgrey}{.}\textcolor{cdgrey}{.}UUA\textcolor{cdgrey}{.}\textcolor{cdgrey}{.}\textcolor{cdgrey}{.}CAA\textcolor{cdgrey}{.}C\textcolor{cdgrey}{.}\textcolor{cdgrey}{.}\textcolor{cdgrey}{.}A\textcolor{cdgrey}{.}\textcolor{cdgrey}{.}A\textcolor{cdgrey}{.}AU\textcolor{cdgrey}{.}\textcolor{cdgrey}{.}AC\textcolor{cdgrey}{.}\textcolor{cdgrey}{.}U\textcolor{cdgrey}{.}\textcolor{cdgrey}{.}\textcolor{cdgrey}{.}\textcolor{cdgrey}{.}G\textcolor{cdgrey}{.}C\textcolor{cdgrey}{.}C\textcolor{cdgrey}{.}G\textcolor{cdgrey}{.}U\textcolor{cdgrey}{.}\textcolor{cdgrey}{.}C\textcolor{cdgrey}{.}GG\textcolor{cdgrey}{.}\textcolor{cdgrey}{.}\textcolor{cdgrey}{.}\textcolor{cdgrey}{.}\textcolor{cdgrey}{.}\textcolor{cdgrey}{.}\textcolor{cdgrey}{.}C\textcolor{cdgrey}{.}\textcolor{cdgrey}{.}\textcolor{cdgrey}{.}UAC\textcolor{cdgrey}{.}GG\textcolor{cdgrey}{.}\textcolor{cdgrey}{.}\textcolor{cdgrey}{.}\textcolor{cdgrey}{.}A\textcolor{cdgrey}{.} \\
  B1 & C\textcolor{cderr}{C}CC\textcolor{cderr}{C}GUUAGUUCAA\textcolor{cderr}{C}CGG\textcolor{cderr}{C}AGAACAUCGACCUU\textcolor{cderr}{U}\textcolor{cderr}{A}AAGUCGCAGGUUAC\textcolor{cderr}{A}GGUUC\textcolor{cderr}{A}AGUCC\textcolor{cderr}{C}GUACGGG\textcolor{cderr}{G}G\textcolor{cderr}{G}CAA \\
     & \normalfont\scriptsize Baseline $L{=}1$: recovery 0.76, $Z=-0.90$ (WT $Z=-2.24$) \\
  R5 & C\textcolor{cderr}{C}C\textcolor{cderr}{U}\textcolor{cderr}{C}GUUAGUUCAA\textcolor{cderr}{C}CGG\textcolor{cderr}{U}AGAACAUCGACCUU\textcolor{cderr}{U}UAAGUCGCAGGUUAC\textcolor{cderr}{A}GGUUC\textcolor{cderr}{A}A\textcolor{cderr}{A}UCC\textcolor{cderr}{U}GUACGGG\textcolor{cderr}{G}GACAA \\
     & \normalfont\scriptsize Recursive $L{=}5$: recovery 0.76, $Z=-1.31$ (WT $Z=-2.24$) \\
\end{tabular}
}

\end{document}